\documentclass[acmtog,authorversion,nonacm]{acmart} %
\acmSubmissionID{576}

\usepackage{booktabs} %

\usepackage[ruled]{algorithm2e} %

\SetAlFnt{\small}
\SetAlCapFnt{\small}
\SetAlCapNameFnt{\small}
\SetAlCapHSkip{0pt}

\copyrightyear{2022}
\acmYear{2022}
\setcopyright{acmcopyright}
\acmConference{SIGGRAPH 2022}{8-11 Aug.}{Vancouver}
\acmDOI{10.1145/8888888.7777777}
\acmISBN{978-1-4503-1234-5/22/07x}

\setcopyright{acmlicensed}\acmJournal{TOG}
\acmYear{2022}\acmVolume{41}\acmNumber{4}\acmArticle{122}\acmMonth{7} \acmDOI{10.1145/3528223.3530156}

\usepackage[normalem]{ulem}
\usepackage{stackengine}
\usepackage{bm}
\usepackage[ruled]{algorithm2e} %
\usepackage{wrapfig}
\usepackage[export]{adjustbox}
\usepackage{tabularx}
\usepackage{tikz}
\usepackage{booktabs}

\renewcommand{\vec}[1]{\mathbf{#1}}

\newcommand{\todo}[1]{}

\newcommand\hide[1]{}

\newcommand\review[1]{{#1}}

\definecolor{bkcolor}{RGB}{210,10,210}
\definecolor{vcacolor}{RGB}{123,50,210}
\definecolor{barbaracolor}{RGB}{100,100,200}
\definecolor{gccolor}{RGB}{60,145,110}
\definecolor{hxcolor}{RGB}{123,104,238}
\definecolor{lwcolor}{RGB}{0,0,200}
\definecolor{awesome}{rgb}{1.0, 0.13, 0.32}
\definecolor{gzcolor}{rgb}{0.2, 0.7, 0.2}

\newcommand\ie{\textit{i.e.},~}
\newcommand\eg{\textit{e.g.},~}
\newcommand\Fig[1]{Fig.~\ref{fig:#1}}
\newcommand\Sec[1]{Sec.~\ref{sec:#1}}
\newcommand\Eq[1]{Eqn.~(\ref{eq:#1})}

\newcommand\Tab[1]{Table~(\ref{tab:#1})}

\newcommand{\SO}[0]{\mathrm{SO}}

\newcommand{\0}[0]{\mathbf{0}}

\renewcommand{\a}[0]{\check{\mathbf{a}}}
\renewcommand{\b}[0]{\mathbf{b}}
\newcommand{\h}[0]{\mathbf{h}}
\newcommand{\z}[0]{\mathbf{z}}
\newcommand{\s}[0]{\mathbf{s}}      %
\newcommand{\f}[0]{\check{\mathbf{f}}}      %
\renewcommand{\r}[0]{\check{\mathbf{r}}}    %
\renewcommand{\u}[0]{\mathbf{u}}    %
\newcommand{\x}[0]{\mathbf{x}}      %
\newcommand{\A}[0]{\mathbf{A}}      %
\newcommand{\F}[0]{\mathbf{F}}      %
\newcommand{\R}[0]{\mathbf{R}}      %
\newcommand{\G}[0]{\mathbf{G}}      %
\newcommand{\W}[0]{\mathbf{W}}      %
\renewcommand{\H}[0]{\mathbf{H}}    %
\newcommand{\N}[0]{\mathcal{N}}     %
\renewcommand{\vec}[0]{\operatorname{vec}}

\usepackage{amsmath}

\DeclareMathOperator*{\argmin}{arg\,min}

\newcolumntype{P}[1]{>{\centering\arraybackslash}p{#1}}

\begin{document}

\title[Implicit Neural Representation for Physics-driven Actuated Soft Bodies]
{%
Implicit Neural Representation for Physics-driven Actuated Soft Bodies}

\author{Lingchen Yang}
\affiliation{%
 \institution{ETH Zurich}
 \country{Switzerland}}
\email{lingchen.yang@inf.ethz.ch}
\author{Byungsoo Kim}
\affiliation{%
 \institution{ETH Zurich}
 \country{Switzerland}}
\email{kimby@inf.ethz.ch}
\author{Gaspard Zoss}
\affiliation{%
 \institution{DisneyResearch|Studios}
 \country{Switzerland}}
\email{gaspard.zoss@disneyresearch.com}
\author{Baran Gözcü}
\affiliation{%
 \institution{ETH Zurich}
 \country{Switzerland}}
\email{baran.goezcue@inf.ethz.ch}
\author{Markus Gross}
\affiliation{%
 \institution{ETH Zurich}
 \country{Switzerland}}
\email{grossm@inf.ethz.ch}
\author{Barbara Solenthaler}
\affiliation{%
 \institution{ETH Zurich}
 \country{Switzerland}}
\email{solenthaler@inf.ethz.ch}

\renewcommand\shortauthors{L. Yang, B. Kim, G. Zoss, B. Gözcü, M. Gross, B. Solenthaler}

\begin{abstract}

Active soft bodies can affect their shape through an internal actuation mechanism that induces a deformation. Similar to recent work, this paper utilizes a differentiable, quasi-static, and physics-based simulation layer to optimize for actuation signals parameterized by neural networks.
Our key contribution is a general and implicit formulation to control active soft bodies by defining a function that enables a continuous mapping from a spatial point in the material space to the actuation value. This property allows us to capture the signal's dominant frequencies, making the method discretization agnostic and widely applicable.
We extend our implicit model to mandible kinematics for the particular case of facial animation and show that we can reliably reproduce facial expressions captured with high-quality capture systems. We apply the method to volumetric soft bodies, human poses, and facial expressions, demonstrating artist-friendly properties, such as simple control over the latent space and resolution invariance at test time.  
Please refer to our \href{https://studios.disneyresearch.com/2022/07/24/implicit-neural-representation-for-physics-driven-actuated-soft-bodies/}{{\textcolor{orange}{\emph{project page}}}} for more details.

\end{abstract}



\begin{teaserfigure}
\centering
\includegraphics[width=0.27\textwidth]{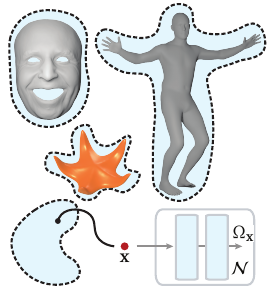}\hspace{10pt}
\includegraphics[width=0.6\textwidth]{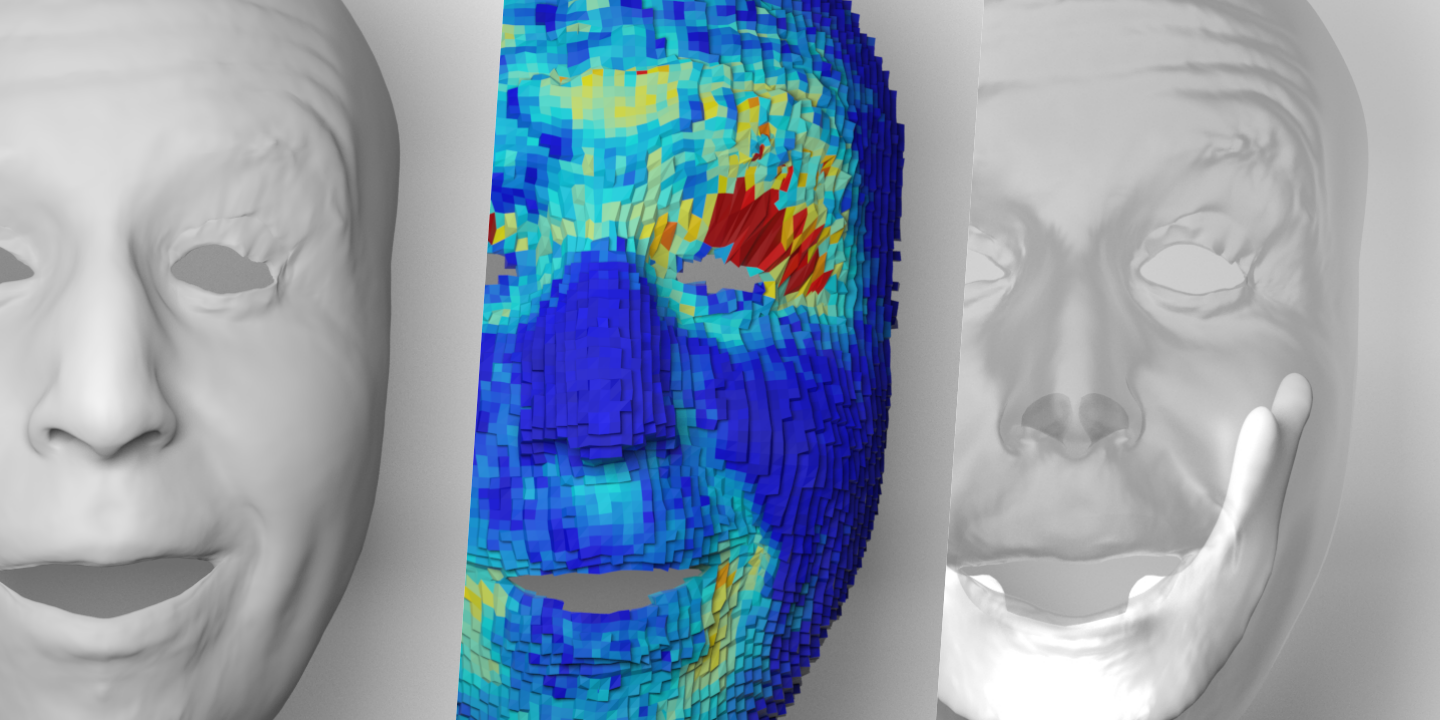}
\caption{
We present a method to control active soft bodies using an implicit neural representation. 
A continuous mapping from a spatial point $\x$ in the material space to the control parameters $\Omega_{\x}$ is established, rendering the method discretization agnostic and applicable to various soft body types (left). 
For faces, we consider both actuation and jaw kinematics for articulating high-fidelity expressions (right).}
\label{fig:teaser}
\end{teaserfigure}

\maketitle

\section{Introduction}

Active soft bodies can deform their shape through an internal actuation mechanism and play a key role in physics-based animation. The actuation mechanism induces an internal force that drives the motion of the discrete mesh elements of a body, and can be driven by artists for shape targeting problems~\cite{klar2020shape}, induced by the underlying musculature in the case of human bodies and faces~\cite{Sifakis05}, or controlled by embedded sensors in robotics applications~\cite{baecher21}.  

In order to guide a simulation towards a target shape, an inverse problem arises when the optimal actuation signals, which deform the simulation mesh to match a given target surface, must be found for each element.
This problem can be solved with a differentiable simulator, either through a neural network approach~\cite{Fulton19} or by directly differentiating the governing equations and thus inducing a strong physics prior~\cite{Du2022DiffPD:Dynamics}. For the particular application of faces, \citeN{srinivasan2021learning}~ postulates that the control parameters of the soft tissue actuation mechanism are low-dimensional and can be learned by a neural network that is coupled with a differentiable quasi-static physics solver. Their actuation-generative network is tied to a specific discretization, which requires the re-definition of the network each time the discretization of the input shape changes. It is thus not generally applicable to soft bodies of arbitrary shapes and resolution inputs.

In this work, we present a general implicit physics- and data-driven formulation to control active soft bodies. We build upon the critical assumption that we can learn a continuous function which maps the spatial point in the material space (undeformed space) to the actuation signal. We hypothesize that the trained model can effectively capture the dominant frequencies of the actuation signal and reproduce them at test time. 
Our approach offers the core advantage that the method is agnostic to the underlying shape representation. In other words, with our method, there is no need to manually re-define the network architecture or retrain the model if the underlying representation or resolution changes. These fundamental properties render the method generally applicable to arbitrary soft body inputs and reduce the required expert knowledge, allowing artists to generate new poses and animations efficiently.
We underline our claims by showing results for volumetric soft bodies, human body motion, and facial expressions. For the particular case of faces, the motion of the soft tissue is highly dominated by the movement of an internal bone structure that defines Dirichlet boundary conditions in the simulator. Therefore, our implicit model optimizes not only actuation signals but also mandible kinematics.   

The main contributions of our work are summarized as follows: 
\begin{itemize}
    \item An implicit neural method for computing actuation signals of active soft bodies, enabling applicability to arbitrary shapes. 
    \item Extension of the implicit method to facial animations, where mandible kinematics are optimized and coupled with the physics solver via Dirichlet boundary conditions. 
    \item Conditioning of the actuation network on a continuous resolution input to enable resolution invariance at test time.
    \item Improved performance due to a closed-form Hessian of the energy density function in the implicit model and a weight matrix modulation in the network design.
    \item Demonstration of the effectiveness and generality of our method on active objects of different types and resolutions. 
\end{itemize}
\section{Related Work}

We briefly review prior work on soft body and face simulation, focusing on inverse problems, data-driven methods, and implicit neural representations.

\paragraph{Differentiable simulation}
Differentiable physics solvers enable the use of gradient-based methods for solving complex control problems. ChainQueen~\cite{hu2019chainqueen} and DiffTaichi~\cite{Hu2019difftaichi} apply differentiability to the material point method. \citeN{Hahn19}~and~\citeN{Geilinger20} use differentiable implicit FEM simulators that systematically apply a combination of the adjoint method and sensitivity analysis to derive gradients.
\review{\citeN{Qiao2021Differentiable} introduces a differentiable framework to dynamically simulate a soft body with rigid skeleton actuation. Similar to them, we consider skeleton kinematics but restricted to the quasi-static state. Hence, we adopt the quasi-static skeleton position for the target shape as the Dirichlet boundary condition in the simulation.}
Differentiable Projective Dynamics (DiffPD) with implicit time stepping is used in~\cite{Du2022DiffPD:Dynamics}. A focus is put on optimizing the backpropagation to achieve fast gradient computations that outperform the standard Newton's method. In our work, we adopt DiffPD as the backward solver for the optimization.
Neural physics solvers are naturally differentiable. An autoencoder and latent-space dynamics are used in~\cite{Fulton19} to simulate deformable bodies, and a graph neural network is applied to particle-based soft body dynamics in~\cite{Sanchez20}.

\paragraph{Inverse Modeling}
Differentiable physics solvers are used to find minimizers for inverse problems in robotics control tasks~\cite{Degrave19,Macklin20}, \review{cloth simulation~\cite{Liang19,Qiao20}}, and fluid dynamics~\cite{Holl20,Tang21}. In the context of digital humans, inverse problems are solved to identify mechanical properties of the face, in particular heterogeneous stiffness and prestrain, using 3D scans and imaging data~\cite{Kadlecek2019BuildingData}. A differentiable physics-based muscle model is applied to 3D facial poses and 2D images, alike~\cite{Bao19}. Instead of explicitly modeling muscles, a simulation mesh is subdivided into tetrahedral elements where each can be activated \cite{ichim2017phace,klar2020shape}. \citeN{ichim2017phace}~additionally includes jaw kinematics, similar to our work. However, different from them, we explicitly build the differential connection between the network and the kinematics. \citeN{srinivasan2021learning}~uses a neural network in combination with a differentiable soft body solver to learn an actuation mechanism such that a facial mesh is deformed into a desired expression after physics-based simulation. The network depends on the input representation, and hence needs to be manually changed if the input changes. We overcome this limitation in our work by using an implicit representation that enables a continuous mapping from a spatial point to a control value.

\paragraph{Blendshapes and data-driven models} 
Blendshapes are widely used for facial and human body animation~\cite{lewis2014egstarblendshape}. They approximate the space of valid poses, and new poses can be computed as a linear, weighted combination of the basis shapes.
These early methods have been improved by data-driven models, which leverage high-quality surface scan datasets~\cite{klehm15egstar,Egger20}. More recently, neural networks have been used to disentangle the facial identity and expression, and jointly model the geometry and appearance~\cite{chandran2020semanticface,li2020faceattribute}. Deep learning methods have also been applied to human bodies, either to model soft tissue deformation as a function of body motion~\cite{casas2018softtissue}, in combination with body shapes~\cite{loper2015smpl,PonsMollDyna15}, or by using nonlinear subspaces to encode tissue deformation~\cite{Santesteban2020softsmpl}. 

\paragraph{Coordinate-based MLPs}
Coordinate-based multilayer perceptrons (MLPs)~\cite{tancik2020fourierfeature} have recently been very successful at generating high-fidelity representations of different visual problems.
A series of works have adopted coordinate-based MLPs for 3D scene reconstruction~\cite{nerf2020} or 3D objects~\cite{park2019deepsdf,mescheder2019occ}.
\review{\citeN{nerf2020}} demonstrates the effectiveness of architectures using a positional encoding scheme for representing high-frequency signals and presents photo-realistic novel view synthesis.
SIREN~\cite{siren2020} adopts periodic activation functions for implicit neural representations of intricate objects and their derivatives.
Our network design is motivated by SIREN. However, SIREN's limited performance on high-dimensional inputs makes their design not directly applicable to our problem setting.
A couple of works~\cite{zhou2021cips3d,chan2021pigan,orel2021stylesdf} extend SIREN to condition on the input by using adaptive instance normalization (AdaIN)~\cite{huang2017adain} in the feature space.
In contrast to AdaIN, we condition the weight matrix on the latent code, which reduces memory and enables an efficient backpropagation.
\begin{figure*}[ht!] %
\centering
\includegraphics[trim=0px 0px 0px 0px, clip, width=0.925\textwidth]{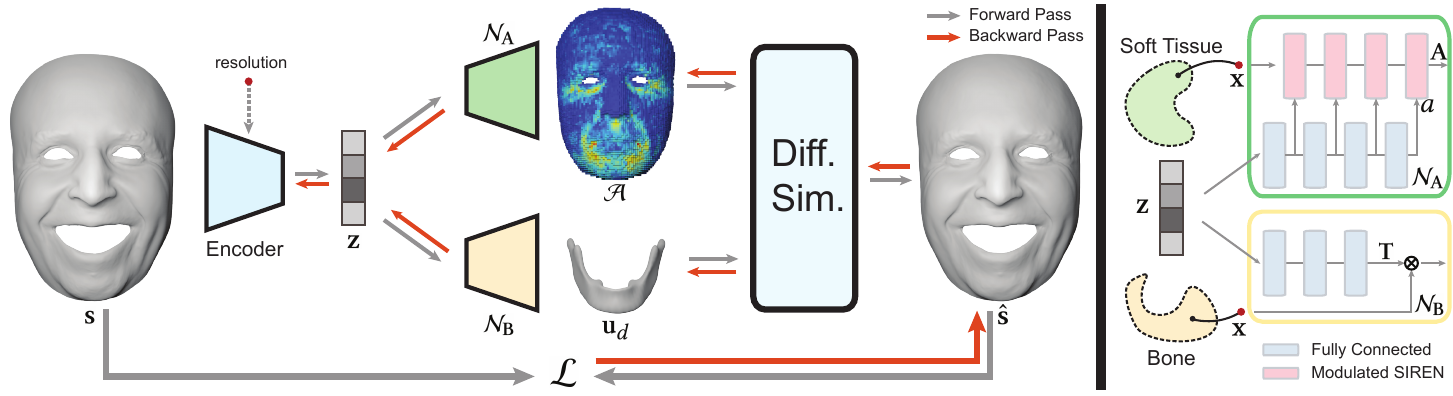}
\caption{Overview of our method.
Using a set of observations $\s$ (target poses) we learn actuation signals $\mathcal{A}$ and mandible kinematics $\u_d$ (for faces only), such that when using these parameters in a forward pass the simulation output $\hat\s$ matches ground truth. We implicitly represent the two mechanical properties using the networks $\N_{\mathbf{A}}$ and $\N_{\mathbf{B}}$, and couple them with a differentiable \review{quasi-static} soft body simulator to allow gradient information to flow from the solver to the networks. The encoder is a global shape descriptor and outputs a latent code $\z$.  
}
\label{fig:overviewFace}
\end{figure*}

\section{Inverse Control}

Active objects can affect their shape through an internal actuation mechanism. 
We take an example pose of an object as input, 
and solve the following inverse control problem: 
For each element of a soft body mesh we want to find an activation value, 
such that when used in a \review{forward quasi-static} physics solver, 
the rest pose of the body is deformed into the target shape. 
To solve this, we follow~\citeN{srinivasan2021learning} and 
learn a low-dimensional control space, 
where a latent code $\z$ is sampled to get the fine-grained actuation mechanism 
corresponding to a shape. 
We train the model via differentiable physics, 
which allows us to integrate the numerical solver into the training of neural networks where gradient information flows from the quasi-static simulator into the network. This general idea can be extended to complex bodies as well, such as in facial soft tissues, where deformations are heavily dominated by the underlying bone movement.
An overview of our complete method is shown in \Fig{overviewFace}.

We can generally define an inverse problem as follows. 
We want to learn simulation parameters $\Omega$ from a set of 
observations ${\s}$ such that after optimization, 
a simulator using $\Omega$ can reconstruct ${\s}$. 
To achieve this, we first need to define the energy function $E(\u, \Omega)$, which describes the relation between the simulation mesh vertices $\u$ and the parameters of interest $\Omega$.
The quasi-static simulation requires us to find the minimizer of $E$ over $\u$ given $\Omega$, thereby finding the optimal positions ${\u}^*$ that make the net force become zero everywhere:
\begin{equation}
    \label{eq:dE}
     \nabla_{\u}E({\u}^*, \Omega) = \0.
\end{equation}
With this implicit constraint between ${\u}^*$ and $\Omega$, we can define the inverse problem as
\begin{equation}
    \label{eq:goal}
    \Omega^* = \underset{\Omega}{\argmin} \; \mathcal{L}(\u^*(\Omega), {\s}),
\end{equation}
where the loss function $\mathcal{L}$ measures the difference between the simulation and the observation.
The sensitivity matrix $\partial \u^* / \partial \Omega$, which plays a key role in the optimization, is derived by differentiating on both sides of \Eq{dE} with respect to $\Omega$:
\begin{equation}
    \left. \frac{\partial \nabla_{\u}E}{\partial \u} \right|_{(\u^*, \Omega)} \frac{\partial \u^*}{\partial \Omega} + 
    \left. \frac{\partial \nabla_{\u}E}{\partial \Omega} \right|_{(\u^*, \Omega)} = \0.
\end{equation}
By omitting the explicit evaluation at $(\u^*, \Omega)$, we get a concise form:
\begin{equation}
    \label{eq:sensitivity}
    \frac{\partial \u^*}{\partial \Omega} = - 
    {\underbrace{\Big(\frac{\partial \nabla_{\u}E}{\partial \u}\Big)}_{\H_{\u}}}^{-1} 
    \underbrace{\frac{\partial \nabla_{\u}E}{\partial \Omega}}_{\H_{\Omega}}.
\end{equation}
This matrix is used to propagate the gradient accumulated on $\u^*$ to $\Omega$ by the chain rule 
when optimizing $\mathcal{L}$:
\begin{equation}
    \frac{\partial \mathcal{L}}{\partial \Omega} =  
    \frac{\partial \mathcal{L}}{\partial \u} \frac{\partial \u^*}{\partial \Omega}
    = -\frac{\partial \mathcal{L}}{\partial \u} {\H_{\u}}^{-1} {\H_{\Omega}}.
\end{equation}

\section{Implicit Formulation}
\label{sec:actuation}

Generally, we can implicitly represent any continuous mechanical property using a neural network $\N$ that takes as input a spatial point $\x$ in the material space (undeformed space) and outputs the corresponding property value, \eg $\Omega_\x= \N(\x)$, as illustrated in \Fig{teaser}.
Here, we derive an implicit formulation of an internal actuation mechanism. To animate an active object, we control this mechanism via a network $\N_\A$, trained on target poses.

We adopt the energy density function $\Psi$ from shape targeting \cite{klar2020shape} for the internal actuation model, given as
\begin{equation}\label{eq:shapetarget}
    \Psi(\F, \A) = \underset{\R \in \SO(3)}{\operatorname{min}}||\F - \R\A||^2_F,
\end{equation}
where $\F$ is the local deformation gradient and 
$\A$ is a symmetric $3 \times 3$ actuation matrix, 
whose eigenvectors indicate the muscle contraction directions and eigenvalues the magnitudes of the local deformation.
The rotation matrix $\R$ factors away the rotational difference between $\F$ and $\A$ and makes $\Psi$ rotationally-invariant.
We assign an actuation matrix to every spatial point $\x$ in the object's material space, 
\ie $\A(\x) = \N_{\mathbf{A}}(\x)$.

For the following equations we first define $\vec(\cdot)$ as row-wise flattening of a matrix into a vector,
 and the expanded matrix $\hat{\A}$ as:
\begin{equation}\label{eq:actmat}
    \hat{\mathbf{A}} = \left[
                \begin{array}{ccc} 
                    \mathbf{A}&0&0 \\
                    0&\mathbf{A}&0 \\
                    0&0&\mathbf{A} \\
                \end{array} 
            \right].
\end{equation}
Using $\f=\vec(\F)$, $\r=\vec(\R)$, and $\hat{\A}\r=\vec(\R\A)$, 
the continuous energy function $\tilde{E}$ in the simulation is defined as
\begin{equation}
    \tilde{E} = \int_{\mathcal{D}^{0}} \frac{1}{2}\left\|{{\f}(\x)}-{\hat{\A}(\x){\r}^{*}(\x)}\right\|_{2}^{2} dV,
\end{equation}
where $\mathcal{D}^{0}$ denotes the material space of the object, 
and ${\r}^*$ is the vectorized rotation matrix for point $\x$,
precomputed from the polar decomposition of $\F\A$~\cite{klar2020shape}. 
Following the standard practices of the Finite Element method, 
we discretize $\mathcal{D}^{0}$ using tiny elements connected by nodal vertices $\u$:
\begin{align}
    \tilde{E}    &\approx \sum_{e}\int_{\mathcal{D}_{e}^{0}} \frac{1}{2}\left\|{\f}(\x)-\hat{\A}(\x) {\r}^{*}(\x)\right\|_{2}^{2} dV \\
    \label{eq:energy2}     
    &\approx \sum_{e}\frac{V_e}{N}\sum_{i}^{N} \frac{1}{2}\left\|{\f}(\x_{e, i})-\hat{\A}(\x_{e, i})  {\r}^{*}(\x_{e, i})\right\|_{2}^{2} \\
    \label{eq:energy3} 
    E(\u, \mathcal{A})    &= \sum_{e}\frac{V_e}{N}\sum_{i}^{N} \underbrace{\review{\frac{1}{2}}\left\|\G(\x_{e, i})\u_e-\hat{\A}(\x_{e, i}){\r}^{*}(\x_{e, i}))\right\|_{2}^{2}}_{\Psi_{e,i}},
\end{align}
where $\mathcal{A}$ denotes all the sampled actuation matrices $\A$ 
from the network $\N_{\A}$, 
\ie $\A(\x)=\N_{\A}(\x)$,
$e$ denotes an element, and
$\mathcal{D}_{e}^{0}$ indicates the continuous region inside $e$ while $V_e$ is its volume.
We sample $N$ points inside each $\mathcal{D}_{e}^{0}$ to approximate the integral (\Eq{energy2}).
The deformation gradient 
$\f$ at each point can be approximated by the concatenated nodal vertices $\u_e$, 
associated with the element $e$ where $\x_{e, i}$ resides, 
via the corresponding mapping matrix $\G$ (\Eq{energy3}).
We adopt hexahedral elements and use projective dynamics \cite{Bouaziz2014ProjectiveSimulation} for minimizing $E$.

$\Psi_{e, i}$ in \Eq{energy3} is the key for deriving $\H_\u$ in \Eq{sensitivity}, 
since $\H_\u$ is the accumulation of all Hessians $\nabla^2 \Psi_{e, i}$.
We omit $(\x, e, i, *)$ except $\u_e$ for simplicity. 
As a result of projective dynamics, $\nabla \Psi = \G^{\top}(\G\u_e - \hat{\A}\r)$.
Taking the derivative of $\nabla \Psi$, we get
\begin{align}
    \frac{\partial \nabla \Psi}{\partial \u_e} 
    &= \G^{\top}\G - \G^{\top}\hat{\A} \frac{\partial \r}{\partial \hat{\A}\f}\frac{\partial \hat{\A}\f}{\partial \u_e}\\
    &= \G^{\top}\G - \G^{\top}\hat{\A} \frac{\partial \r}{\partial \hat{\A}\f}\hat{\A}\G \\
    \label{eq:hessian1} &= \G^{\top}\G - \G^{\top}\hat{\A} \H_\R \hat{\A}\G.
\end{align}
Note that $\r$ comes from the polar decomposition of $\hat{\A}\f$, 
$\H_\R$ is \review{the rotation gradient}, namely the derivative with respect to $\hat{\A}\f$.
Luckily, the closed form for $\H_\R$ has already been derived in \cite{kim2020dynamic}, 
which can be constructed from its off-the-shelf three eigenvectors $\mathbf{q}$ and eigenvalues $\lambda$, 
\ie $\H_\R = \sum_{i}\lambda_i \mathbf{q}_{i}\mathbf{q}^{\top}_{i}$.
Similarly, we have $\partial \nabla\Psi / \partial \a$ as
\begin{equation}\label{eq:hessian2}
    \frac{\partial \nabla \Psi}{\partial \a} 
    = - \G^{\top}\hat{\A}\H_\R\hat{\F} - \G^{\top}\hat{\R},
\end{equation}
which can be used for constructing $\H_{\Omega}$  
in \Eq{sensitivity} ($\Omega$ denotes actuation $\mathcal{A}$ here). 
The definition for the expanded matrices $\hat{\F}$ and $\hat{\R}$, and more details on the derivation
can be found in Supplemental Material.
The sensitivity matrix will be used to calculate the
gradient accumulated on the sampled actuation matrices, 
building the connection between the network $\N_{\mathbf{A}}$ and the simulator. 
\review{Compared to \cite{srinivasan2021learning}
where $\H_\u$ and $\H_\Omega$ are evaluated with 9 iterations of sensitivity analysis, 
our closed form requires a single iteration only.} 
In general, the key advantage of our implicit formulation is that it is agnostic to the shape representation, which renders the technique widely applicable.

\section{Network architecture}

For the actuation network $\N_{\mathbf{A}}$ we adopt SIREN \cite{siren2020} due to its excellent representational power.
The most effective part of SIREN is its sine activation function $\h \mapsto \sin (\W\h + \b)$,
where $\h$ is the output hidden vector from a previous layer, and $\W$ and $\b$ are trainable parameters in the current layer.
Since we found that SIREN does not handle high-dimensional inputs well, \ie directly concatenating the latent code $\z$ and spatial point $\x$ as input, 
we instead condition $\W$ on $\z$ as
\begin{equation}
    \W_{i, j} = a_{i} \cdot \hat{\W}_{i,j},
\end{equation} 
where $a_i$ is the modulating coefficient decoded from $\z$ with a tiny MLP, 
$\hat{\W}$ is the shared weight matrix in this layer,
and $i$ and $j$ enumerate the dimensions of input and output feature vectors, respectively.
The motivation for this conditioning mechanism is that a signal (actuation) can be approximated 
as the linear combination ($\W$) of a set of basis functions (the output $\h$ of the previous SIREN layer).
One advantage of modulating the weight matrix $\W$ instead of $\h$ is to save memory.
Since we aim at sampling a large number of points ($\sim268$k in our case) all at once for the simulation,
modulating $\h$ will cache a hidden vector for each sampled points used in backpropagation; 
modulating $\W$ on the other hand entails only one such operation, which will be shared for each sampled point.

\subsection{Training Strategy}
Similar to \cite{srinivasan2021learning}, we use a two-stage training strategy. 
In the first stage, we calculate a plausible approximation of the actuation from the target poses,
which will be used to pretrain $\N_{\mathbf{A}}$ without the differentiable simulator.
Specifically, we approximate the actuation by dragging our volumetric simulation mesh into the given target shape 
with virtual zero-rest-length springs. 
The resulting local deformations are used to initialize the actuation tensors $\mathcal{A}$, 
as in \cite{srinivasan2021learning}.
This pretraining serves as a warm-up and can be used to identify the appropriate dimension of the latent code $\z$.

In the second stage, we fine-tune the parameters of $\N_{\mathbf{A}}$ using our simulator-integrated
pipeline. The loss function $\mathcal{L}$ for the inverse problem is defined as
\begin{equation}
    \mathcal{L} = \sum_{i}|\hat{\s}_i - \s_i| + \alpha |1 - \hat{\mathbf{n}}_{i}^{\top}\mathbf{n}_{i}|,
\label{eq:loss}    
\end{equation}
where $i$ refers to the vertices in the observation space, 
$\s$ and $\mathbf{n}$ indicate the observed vertex positions and normals, $\hat{\s}$ and $\hat{\mathbf{n}}$ are the vertex positions and normals mapped from the simulation space $\u$, and $\alpha\in\mathbb{R}$ is a control weight.

\subsection{Continuous Resolution} 
The main motivation for our presented model is its generalization capability across shapes and resolutions. Resolution invariance is a prerequisite for practical use, as it enables a decision on the resolution at test time depending on the given requirements on speed or geometric fidelity. Our network captures the dominant frequencies of the signal, and thus transfer learning on a different resolution (1 epoch) would be sufficient to make the network consistent with the new discretization. Alternatively, we propose to condition the actuation-generative network $\N_{\A}$ on a continuous resolution input, such that we only need one forward pass to get the optimal actuation.
To do so, we add another small MLP on top of $\N_{\A}$, that takes as input the resolution value, and adds up the output to the
shape latent code $\z$. A detailed overview of the final architecture can be found in the Supplemental Material.

\section{Facial Animation}
For the specific application of our method to faces, we have extended the underlying geometry with bone structures. 
We can articulate diverse expressions by only considering the relative motion 
between the skull and the mandible. 
Thus, the skull is fixed all the time, 
and we include the mandible kinematics in the optimization.

\subsection{Bone Kinematics}
Different from the actuation mechanism acting on the inside of the soft tissue,
the bone structure is located at the boundary and defines the Dirichlet condition in the simulation.
We introduce a second network $\N_{\mathbf{B}}$, which takes a spatial point on the bone region of the object as input and outputs the transformed position. 

For learning the parameters of $\N_{\mathbf{B}}$, 
we divide the simulation vertices $\u$ into two parts $\u = [\u_c, \u_{d}]$,
where $\u_{c}$ denotes the nodal vertices located inside the soft tissue, and $\u_d$ those on the bone region which defines the Dirichlet condition. 
Given that $\u_d$ is calculated with $\mathcal{N}_{\mathbf{B}}$, a quasi-static solution for $\u_{c}^*$ satisfies
\begin{equation}
    \label{eq:dirichlet}
    \nabla_{{\u_c}}E([\u_c^*, \u_{d}], \mathcal{A}) = \0.
\end{equation}
With such an implicit constraint between $\u_c^*$ and $\u_d$, 
we can derive the sensitivity matrix $\partial \u_c / \partial \u_d$ 
by taking the derivative of \Eq{dirichlet} on both sides with respect to $\u_d$:
\begin{equation}
    \label{eq:sensitivity_dirichlet}
    \frac{\partial \u_c^*}{\partial \u_d} = - 
    {\underbrace{\Big(\frac{\partial \nabla_{\u_c}E}{\partial \u_c}\Big)}_{\H_{cc}}}^{-1} 
    \underbrace{\frac{\partial \nabla_{\u_c}E}{\partial \u_d}}_{\H_{cd}},
\end{equation}
where $\H_{cc}$ and $\H_{cd}$ are 
the submatrices of the Hessian $\H_\u$ (which we derive in the \Sec{actuation}).
They can be constructed by only keeping the relevant rows and columns of $\H_\u$.
We use the sensitivity matrix to calculate the
gradient accumulated on the evaluated Dirichlet condition, 
building the connection between the network $\mathcal{N}_{\mathbf{B}}$ and the simulator.

\subsection{Extended Architecture}

The architecture for facial animation consists of three parts as illustrated in \Fig{overviewFace}; an encoder, 
an actuation-generative network $\N_{\mathbf{A}}$,
and a jaw-generative network $\N_{\mathbf{B}}$.
The encoder outputs a latent code $\z$ representing an input face shape $\s$.
$\N_{\mathbf{A}}$ is conditioned on $\z$ to generate the actuation matrix $\A$ for each input point in the soft tissue domain.
$\N_{\mathbf{B}}$ is also conditioned on $\z$ to produce the transformed bone position 
for each input point in the mandible domain.

For $\N_{\mathbf{B}}$, we assume rigid motion, 
and use an MLP to regress the transformation matrix $\mathbf{T}$, from $\z$ such that 
$\N_{\mathbf{B}}(\x, \z)$ $= \mathbf{T}(\z)\x$. Note that $\mathbf{T}$ is shared for each input point but differs from expression to expression.
During the first training stage, we approximate the mandible position 
using the tracking method presented in Zoss et al.~\shortcite{Zoss2019}. 
The second stage then includes $\N_{\mathbf{B}}$ in the optimization. More details can be found in the Supplemental Material.

\section{Implementation}
The two main designs of our pipeline are the differentiable quasistatic physics solver 
and the conditional coordinate-based network.
For the simulator, we use a hexahedral mesh, leveraging the advantage that the discretization is straightforward and can be robustly applied.
We explicitly handle non-manifold topological features by duplicating vertices~\cite{mitchell2015gridiron} around a cut, \eg around the lips to enable mouth opening.
We use Projective Dynamics~\cite{Bouaziz2014ProjectiveSimulation} as our forward solver, 
and differentiable Projective Dynamics~\cite{Du2022DiffPD:Dynamics} as backward solver.
Specifically, we \review{uniformly} sample 8 points inside each hexahedron to estimate the energy function $E$.
We employ a highly optimized singular value decomposition method~\cite{mcadams2011fastsvd} for polar decomposition, 
and use cuSPARSE from CUDA Toolkit to solve large sparse linear systems on the GPU.
As in~\cite{Du2022DiffPD:Dynamics} we utilize the adjoint method to avoid the explicit calculation of ${\H_\u}^{-1}\H_\Omega$,
and adopt the scalability optimization from \cite{srinivasan2021learning} to implicitly evaluate $\H_\Omega$.
\review{The optimization is terminated when either the relative progress is less than 1e-6 or if the maximum of 300 iterations is reached.}

For the network architecture, 
we use our proposed modulated SIREN layer as the backbone
for the actuation-generative network $\N_{\mathbf{A}}$. 
For the other layers of $\N_\A$ and $\N_{\mathbf{B}}$ we use a fully connected layer with a LeakyReLU nonlinearity.
We have chosen $\alpha=0$ for the normal weight in the loss function of \Eq{loss} for both the starfish and human body examples, 
and $\alpha=1$ for the face examples, as we found that the inclusion of the normal constraint positively affects 
the fidelity of the resulting wrinkles and facial details. 
We use ADAM optimizer~\cite{kingma2015adam} to jointly train our networks.
For resolution invariance, we use the number of sampled points on the simulation mesh as the resolution indicator.
We adopt the network pretrained on one resolution ($268$K), and use our simulator integrated pipeline to enable continuous resolution conditioning via transfer learning. We uniformly sample 20 resolutions from $42$K to $268$K for training and 25 different resolutions for testing. 
We run all the experiments using PyTorch~\cite{PyTorch2019} on an NVIDIA RTX 2080Ti with the simulator integrated as a layer.
For additional information on the architecture and training we refer to the Supplemental Material.

\begin{figure}[t] %
\centering
\renewcommand{\arraystretch}{0}
\begin{tabular}{
P{0.25\linewidth}@{\hspace{0pt}}
P{0.25\linewidth}@{\hspace{0pt}}
P{0.25\linewidth}@{\hspace{0pt}}
P{0.25\linewidth}@{\hspace{0pt}}
}
\includegraphics[trim=480 0 480 0, clip, width=1\linewidth]{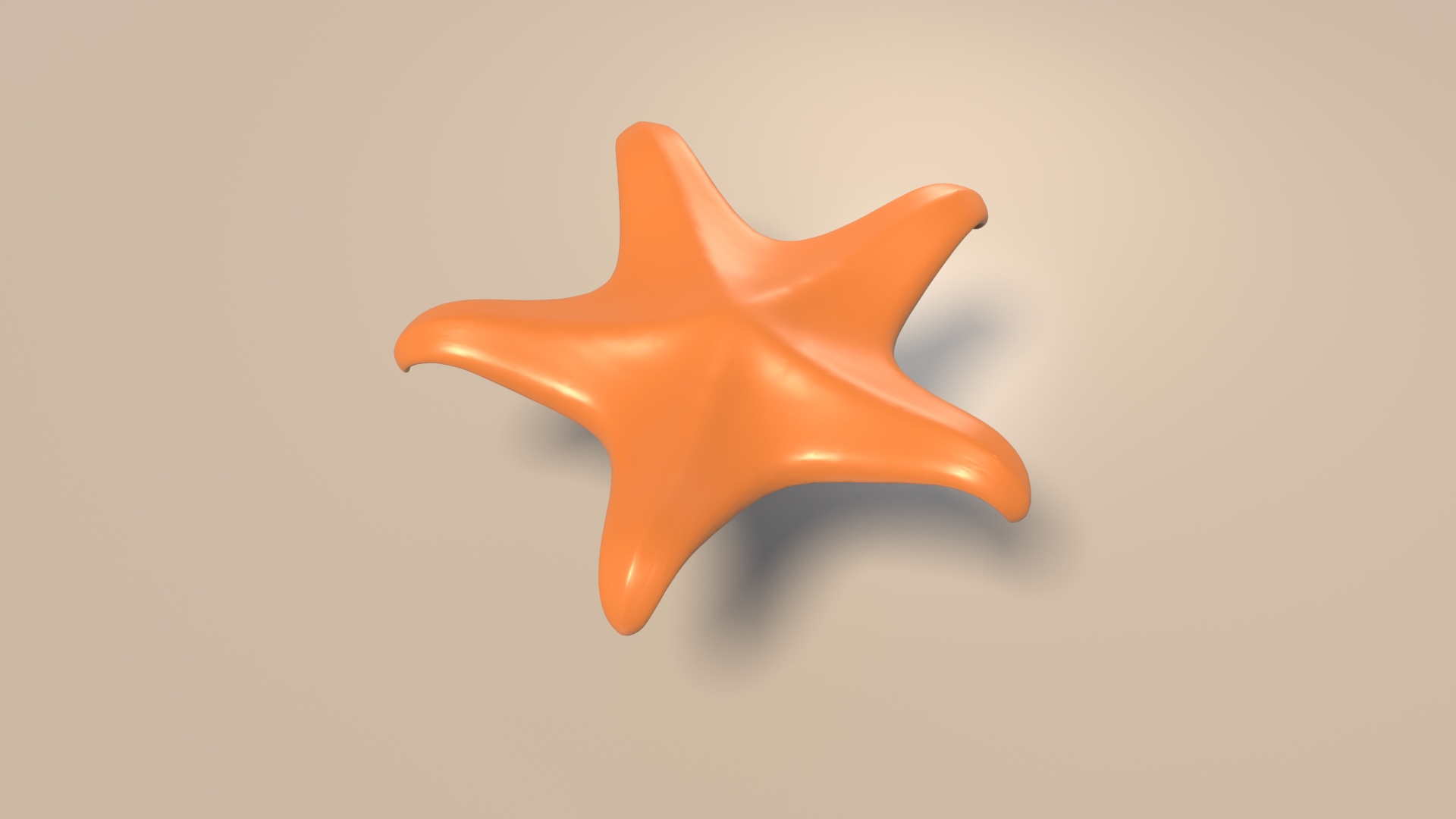} &
\includegraphics[trim=480 0 480 0, clip, width=1\linewidth]{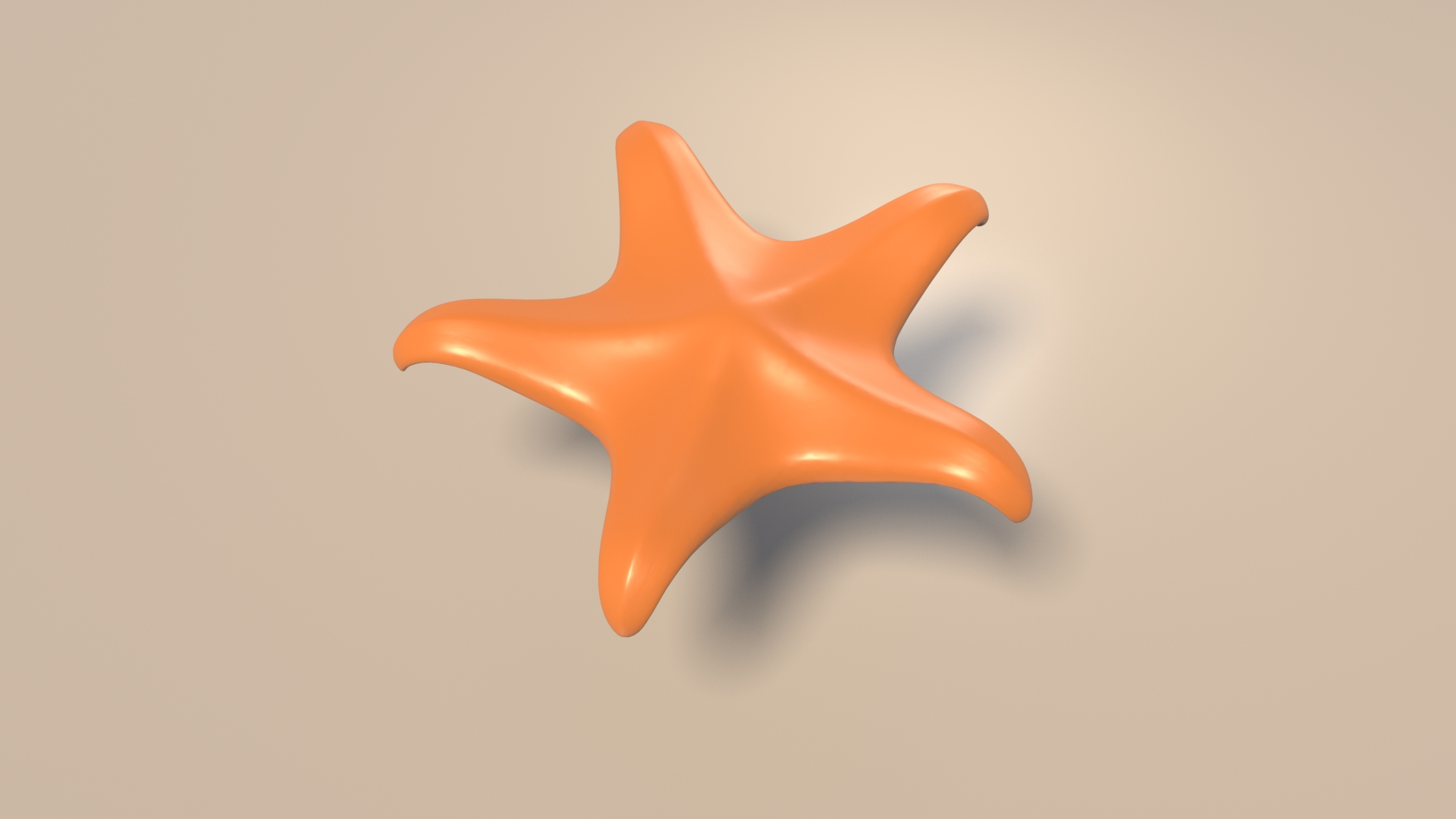} &
\begin{tikzpicture}
\node[inner sep=0pt,anchor=south east] at (0,0)
{ \includegraphics[trim=480 0 480 0, clip, width=1\linewidth]{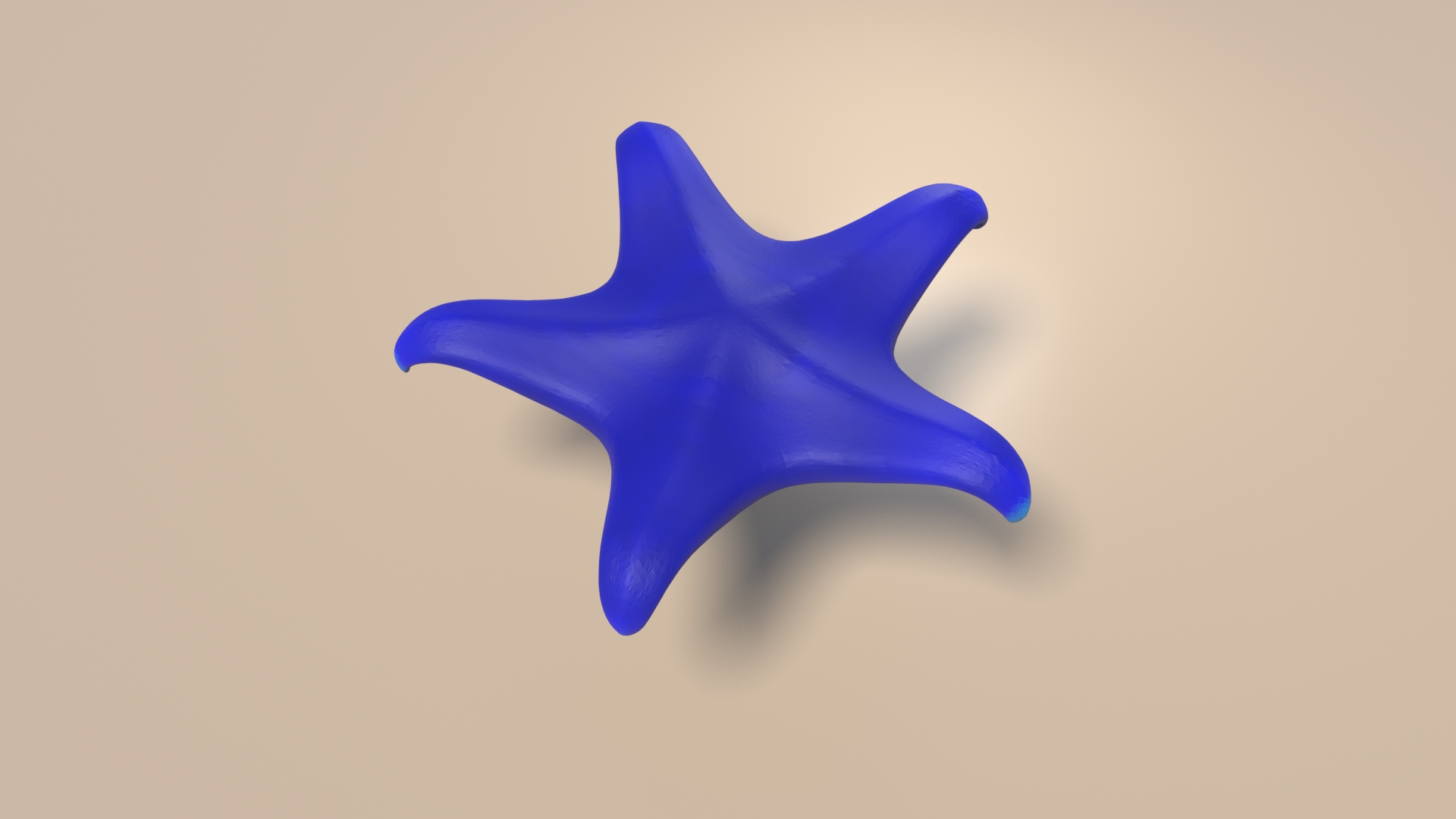} };
\node[inner sep=0pt,anchor=south east] at (-0.2,0.22)
{ \includegraphics[trim=0 0 0 0, clip, width=0.3\linewidth]{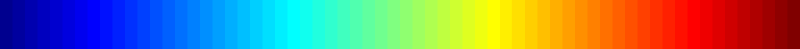} };
\node[inner sep=0pt,anchor=south east] at (0,0.05)
{ \tiny{> 5mm} };
\end{tikzpicture} &
\includegraphics[trim=480 0 480 0, clip, width=1\linewidth]{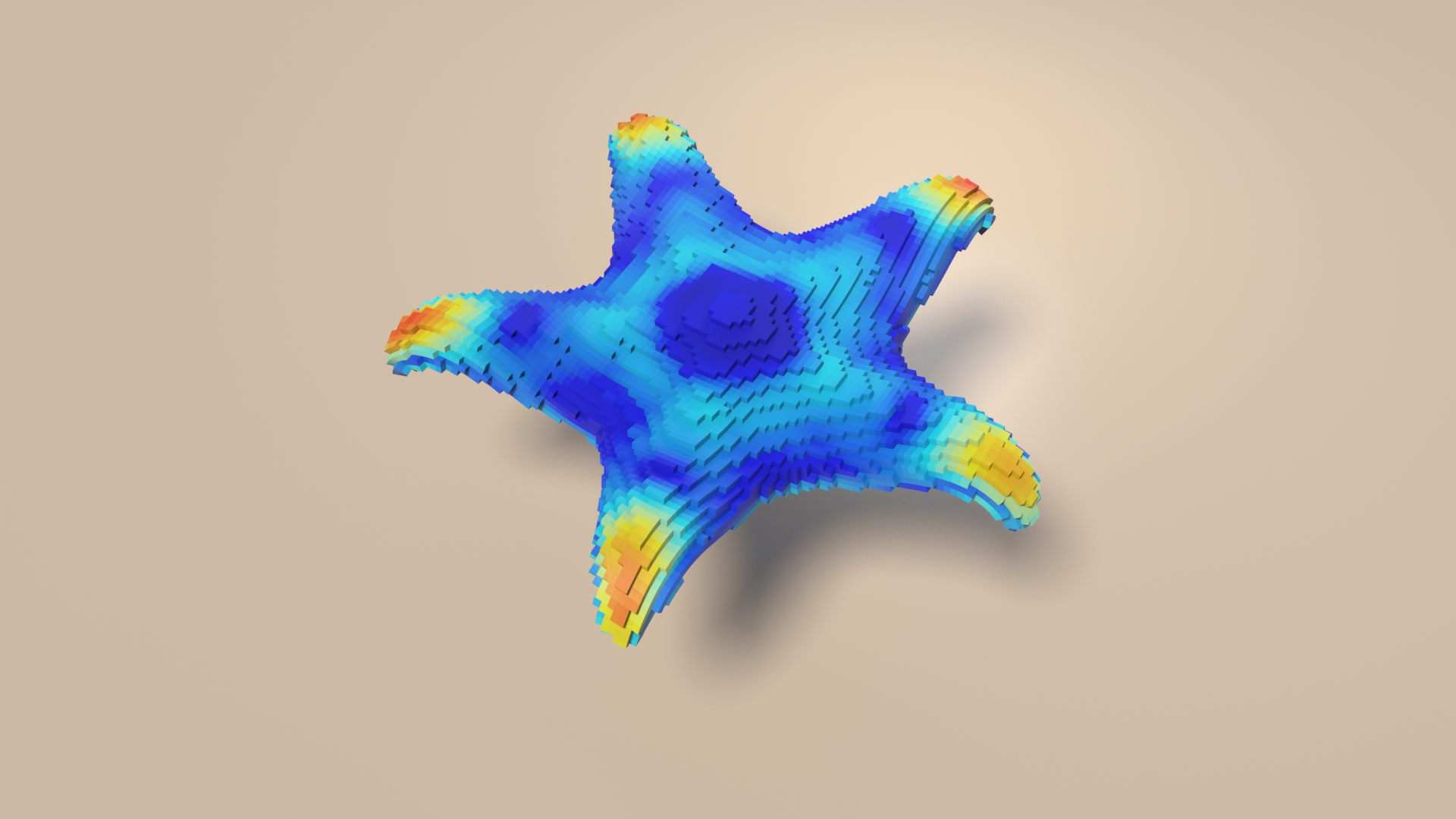}
\\
\includegraphics[trim=480 0 480 0, clip, width=1\linewidth]{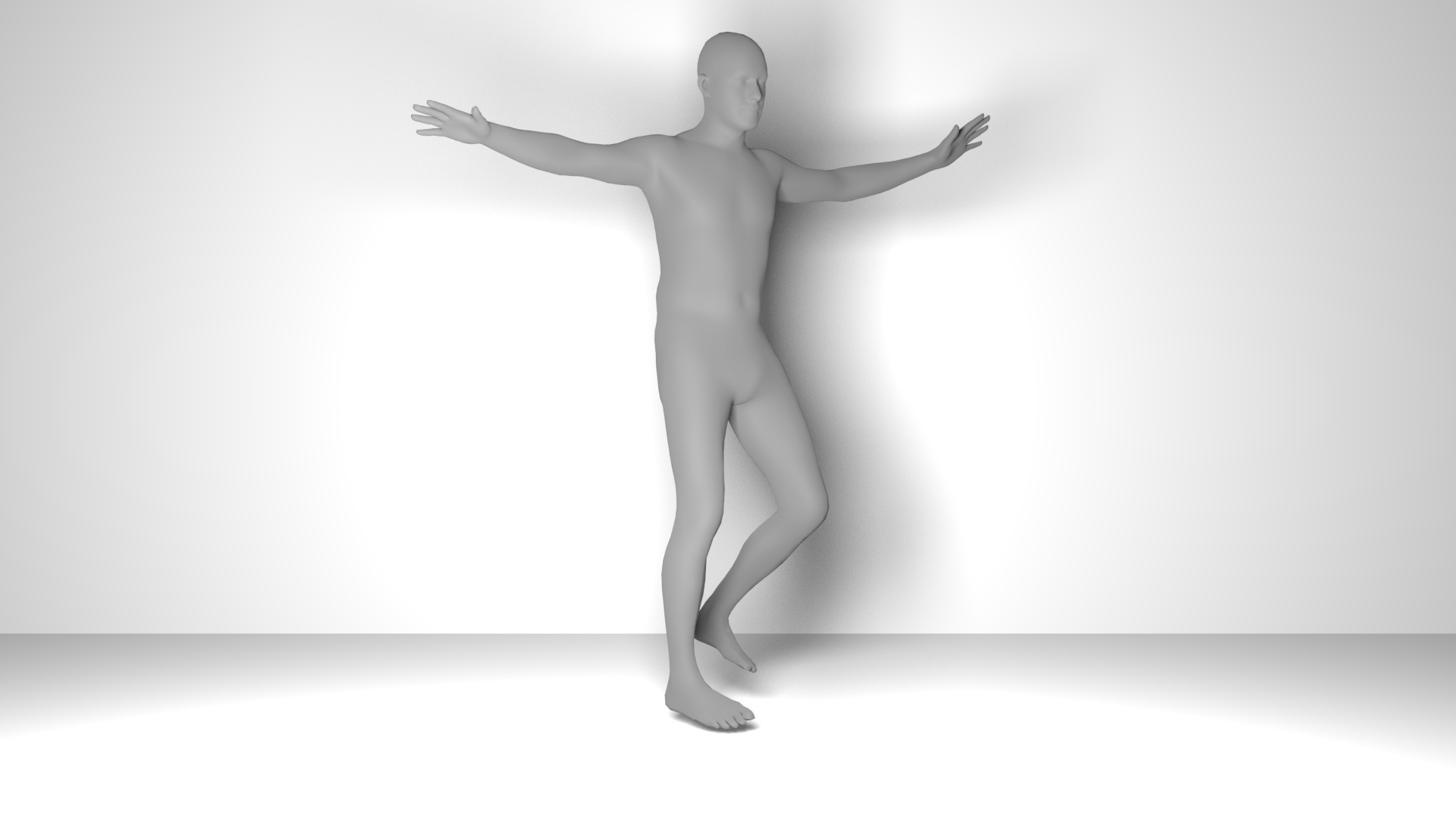} &
\includegraphics[trim=480 0 480 0, clip, width=1\linewidth]{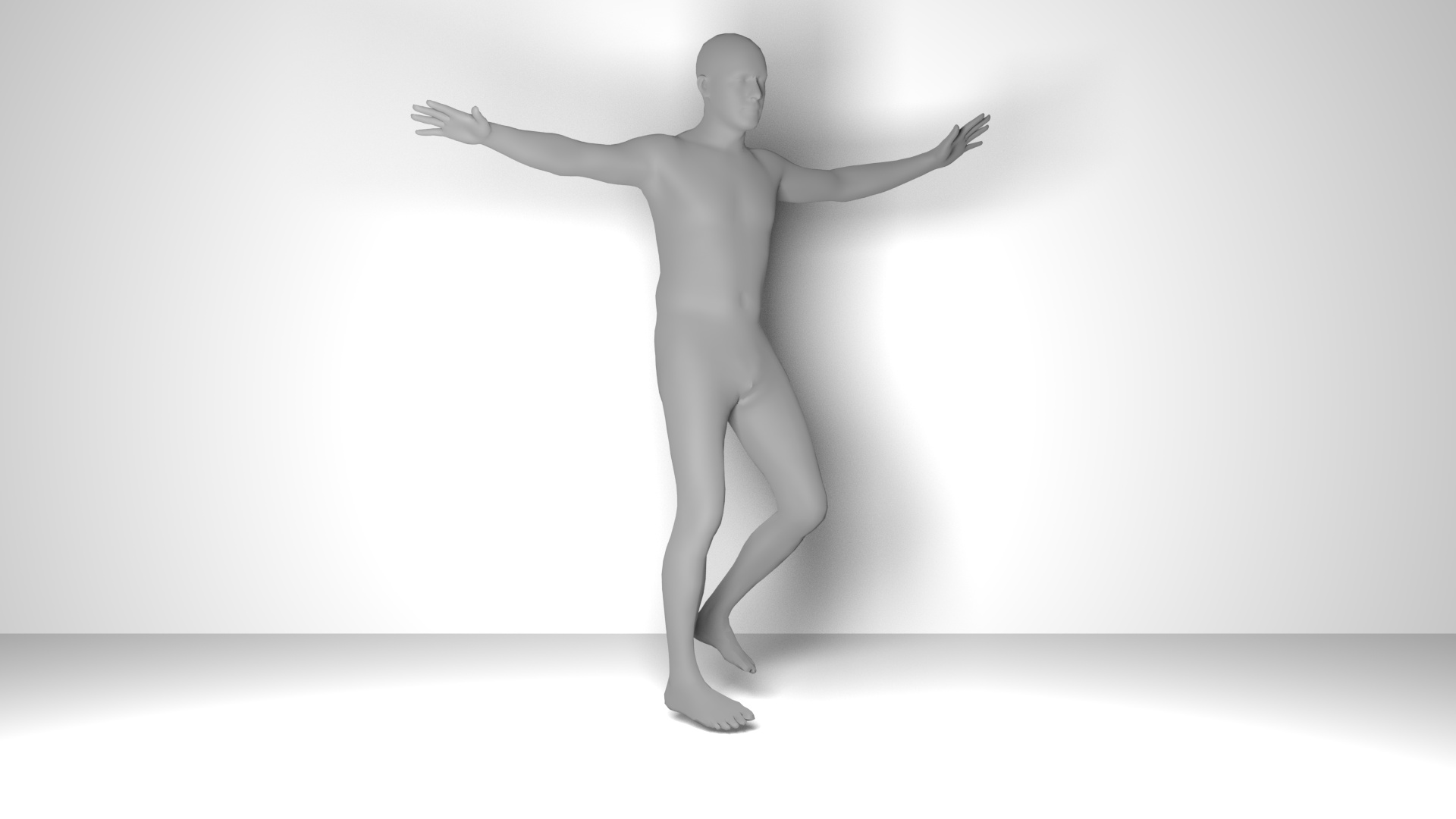} &
\begin{tikzpicture}
\node[inner sep=0pt,anchor=south east] at (0,0)
{ \includegraphics[trim=480 0 480 0, clip, width=1\linewidth]{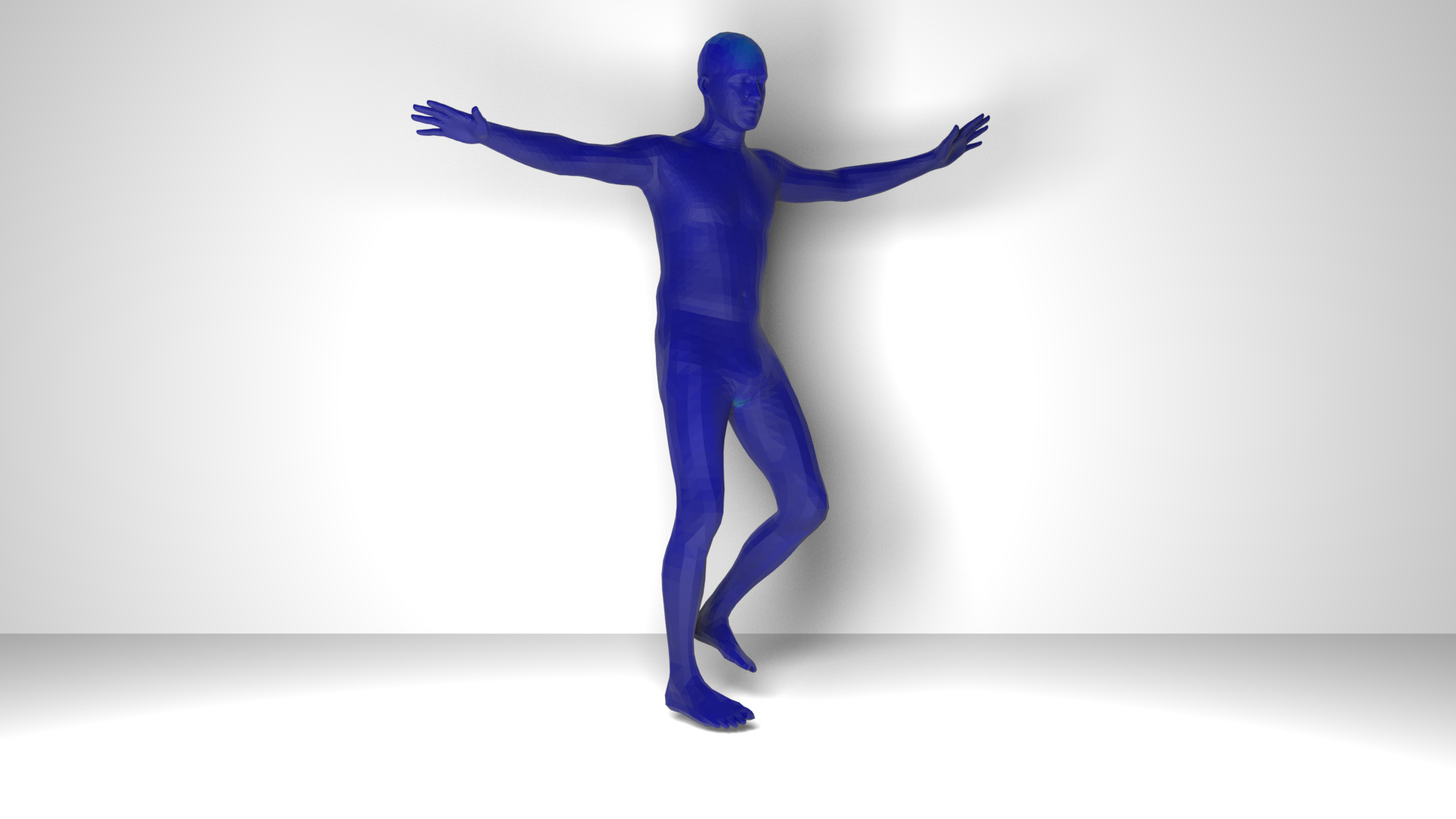} };
\node[inner sep=0pt,anchor=south east] at (-0.2,0.22)
{ \includegraphics[trim=0 0 0 0, clip, width=0.3\linewidth]{fig/cbar} };
\node[inner sep=0pt,anchor=south east] at (0,0.05)
{ \tiny{> 50mm} };
\end{tikzpicture} &
\includegraphics[trim=480 0 480 0, clip, width=1\linewidth]{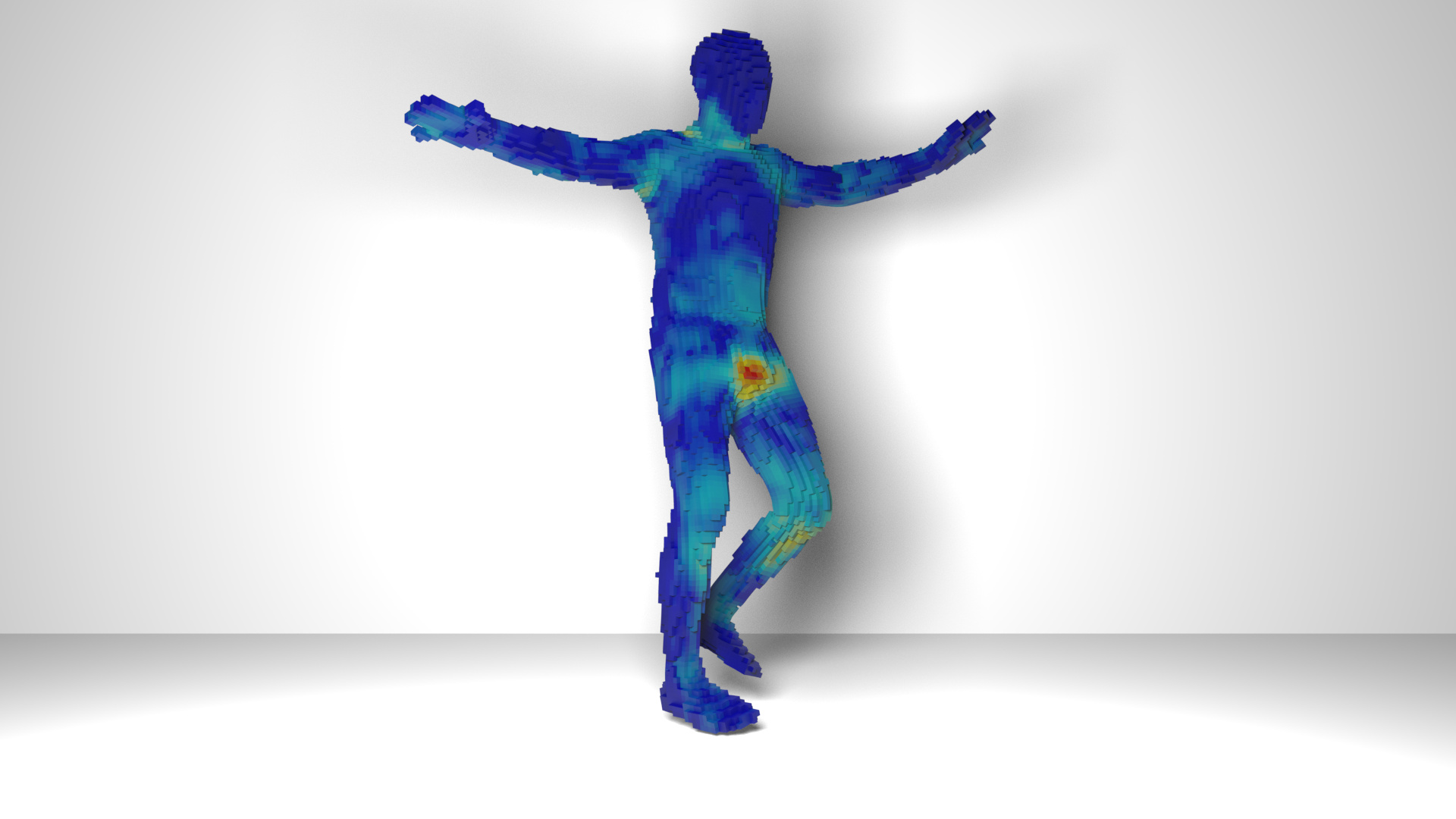}
\\
\includegraphics[trim=480 0 480 0, clip, width=1\linewidth]{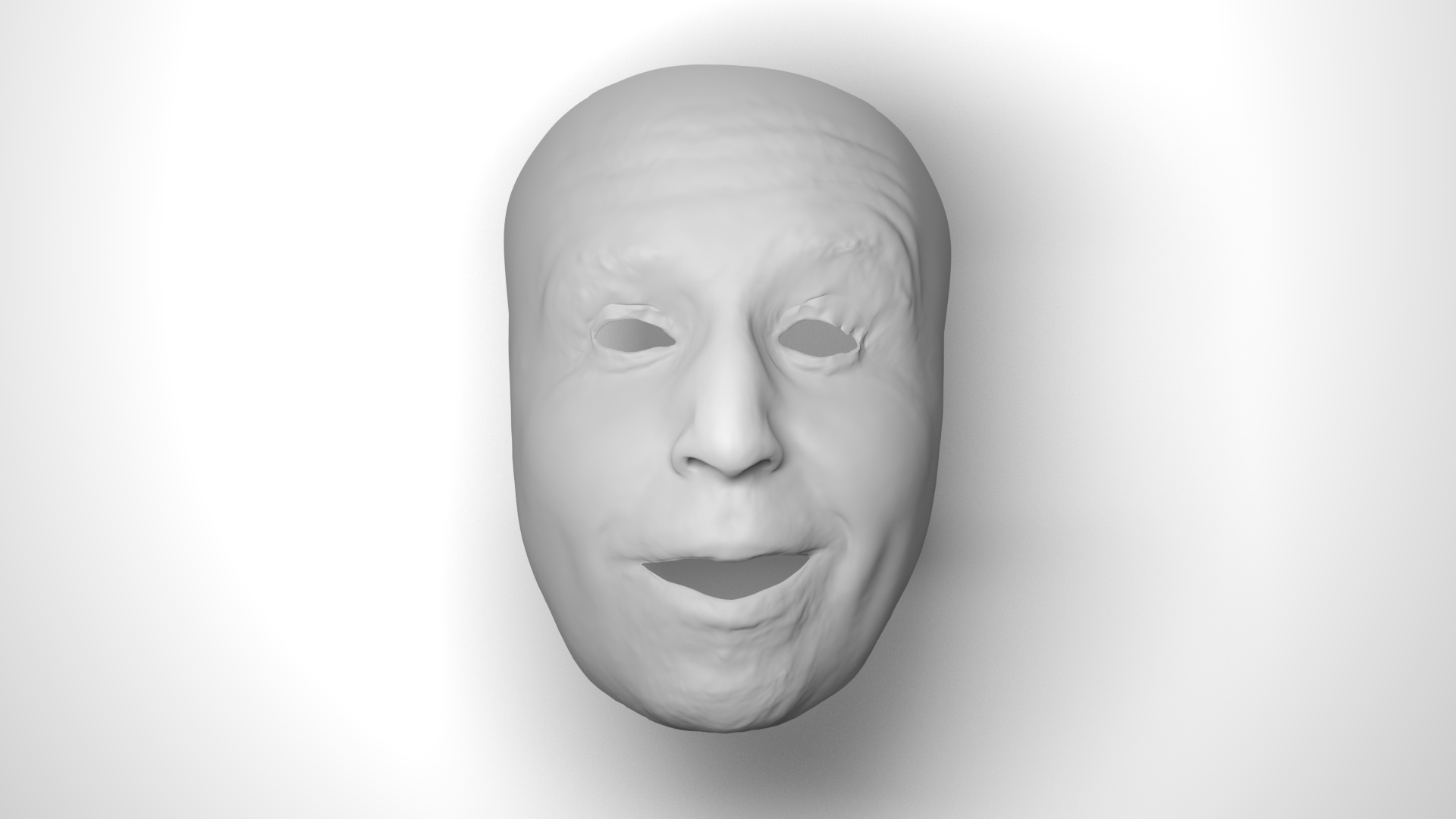} &
\includegraphics[trim=480 0 480 0, clip, width=1\linewidth]{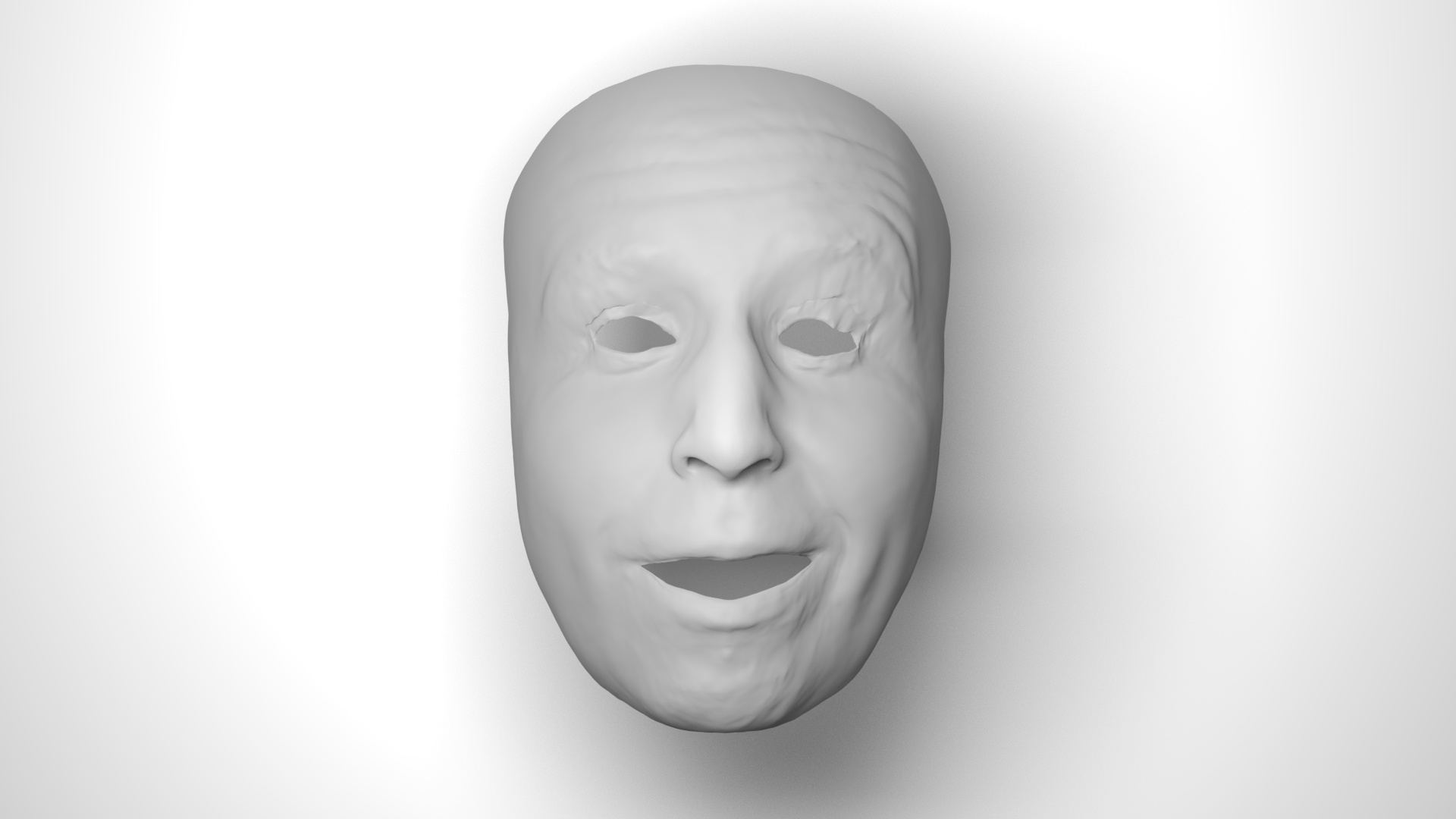} &
\begin{tikzpicture}
\node[inner sep=0pt,anchor=south east] at (0,0)
{ \includegraphics[trim=480 0 480 0, clip, width=1\linewidth]{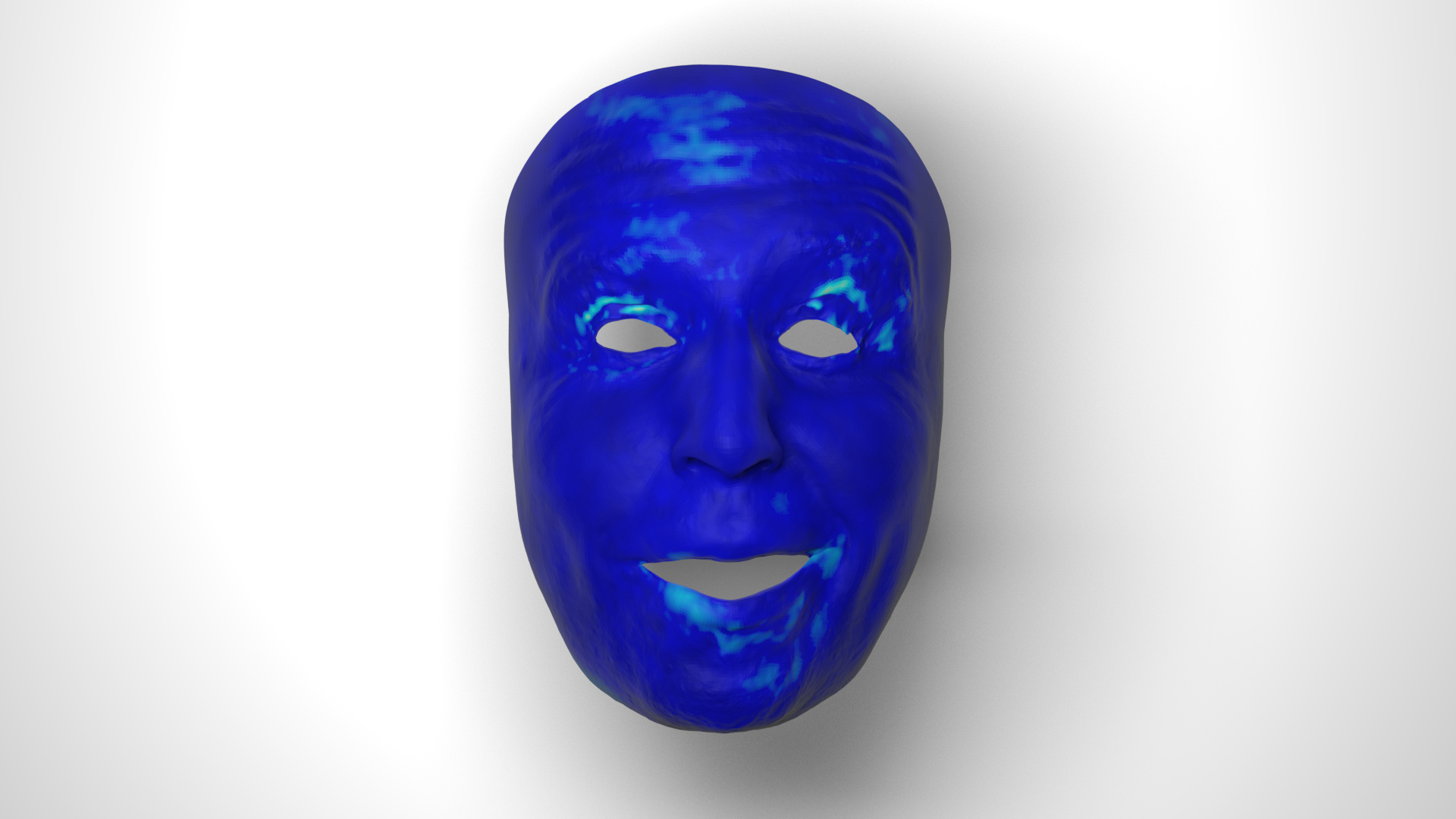} };
\node[inner sep=0pt,anchor=south east] at (-0.2,0.22)
{ \includegraphics[trim=0 0 0 0, clip, width=0.3\linewidth]{fig/cbar} };
\node[inner sep=0pt,anchor=south east] at (0,0.05)
{ \tiny{> 5mm} };
\end{tikzpicture} &
\includegraphics[trim=480 0 480 0, clip, width=1\linewidth]{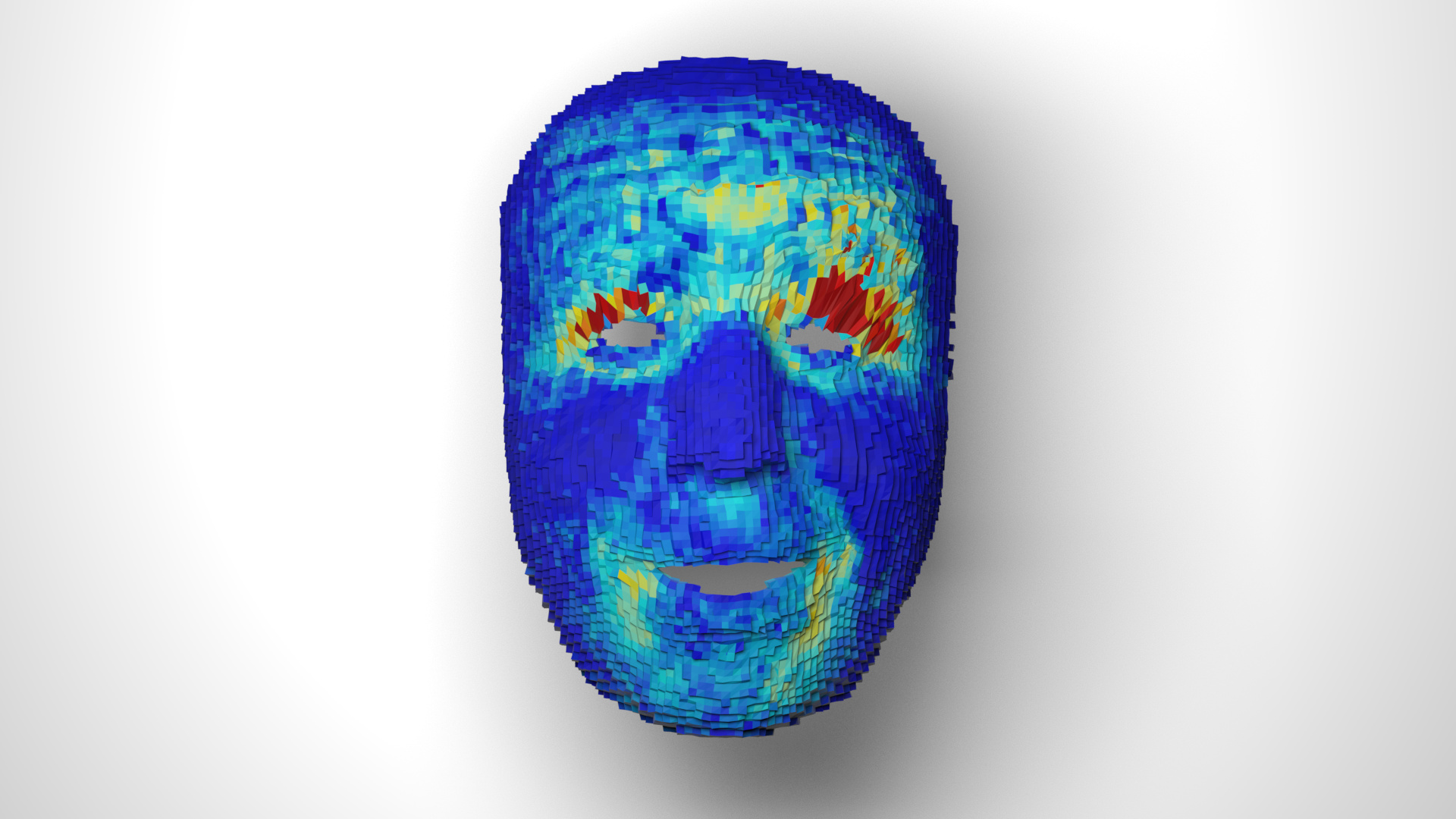}
\\
\vspace{3pt}\small{Target} & \vspace{3pt}\small{Ours} & \vspace{3pt}\small{Error} & \vspace{3pt}\small{Actuation}
\end{tabular}
\caption{Training results using the starfish, human body and face datasets. From left to right: target pose, simulation result, reconstruction error, optimized actuation magnitudes.}
\label{fig:train}
\end{figure}
\input{fig/train_jaw.tex} 

\section{Experiments}

We demonstrate our results using different soft body types. We first describe the datasets, followed by a discussion of the results.

\subsection{Datasets}

\paragraph{Volumetric soft body} 
We generate a synthetic dataset using a starfish surface mesh consisting of 25K vertices and 50K faces. The voxelized representation contains 27K hexahedra and 32K vertices, which we use as our simulation mesh. The dynamic sequence used for learning is generated following the method of DiffPD \cite{Du2022DiffPD:Dynamics}, where the constitutive model consists of the co-rotational elasticity and the volume-preserving term.
We generate 200 frames in total, from which 180 are used for training and 20 for testing.
We sample 216K points at once to evaluate the actuation values for the simulation. 
We use PCA dimensionality reduction to express each shape of the sequence as a 16 dimensional vector, which is then used as input to the encoder. Training takes about 1 day, and testing takes about 10s for 1 forward pass.

\paragraph{Human body motion}
We use a dancing sequence of the AMASS dataset~\cite{AMASS:ICCV:2019}. We sample 1600 frames for training and 100 frames for testing.
We create a voxelized version of the template body mesh, containing 42K hexahedra and 51K vertices.
We sample 336K points at once to evaluate the actuation values for the simulation. 
We use the provided skeleton positions to get the Dirichlet boundary conditions in the simulation, but do not optimize the kinematic chain for simplicity. The skeleton vertices on the simulation mesh can be obtained by forward kinematics.
We directly use the 63 dimensional SMPL body pose parameters~\cite{loper2015smpl} as the global pose descriptor, which is followed by 3 fully connected layers.
Training takes about 5 days.
Testing takes about 15s for 1 forward pass.
\paragraph{Facial expressions}
We use a subset of the facial performance dataset introduced in Zoss et al.~\shortcite{Zoss2020}.  This subset contains 23 expression blendshapes and 641 frames of diverse dialog and facial workout performances for one subject.  We extract 100 frames as testing set and use the remainder for training.  As this dataset contains only surface meshes, we create a voxelized version of the reference (neutral) mesh, containing 33K hexahedra and 46K vertices.  In our experiments with this dataset we sample 268K points at once to evaluate the actuation values for the simulation.  We use the 23 expression blendshapes to fit a per-frame blendweight vector which we use as input to our encoder. \review{We fit a template mandible mesh to the neutral face following the method proposed by \cite{Zoss2018} and approximate the mandible position for each expression using
the tracking method presented in \cite{Zoss2019}.}
Training takes about 3 days. Testing takes about 11s for 1 forward pass.

\begin{figure}[t] %
\centering
\renewcommand{\arraystretch}{0}
\begin{tabular}{
P{0.25\linewidth}@{\hspace{0pt}}
P{0.25\linewidth}@{\hspace{0pt}}
P{0.25\linewidth}@{\hspace{0pt}}
P{0.25\linewidth}@{\hspace{0pt}}
}
\includegraphics[trim=480 0 480 0, clip, width=1\linewidth]{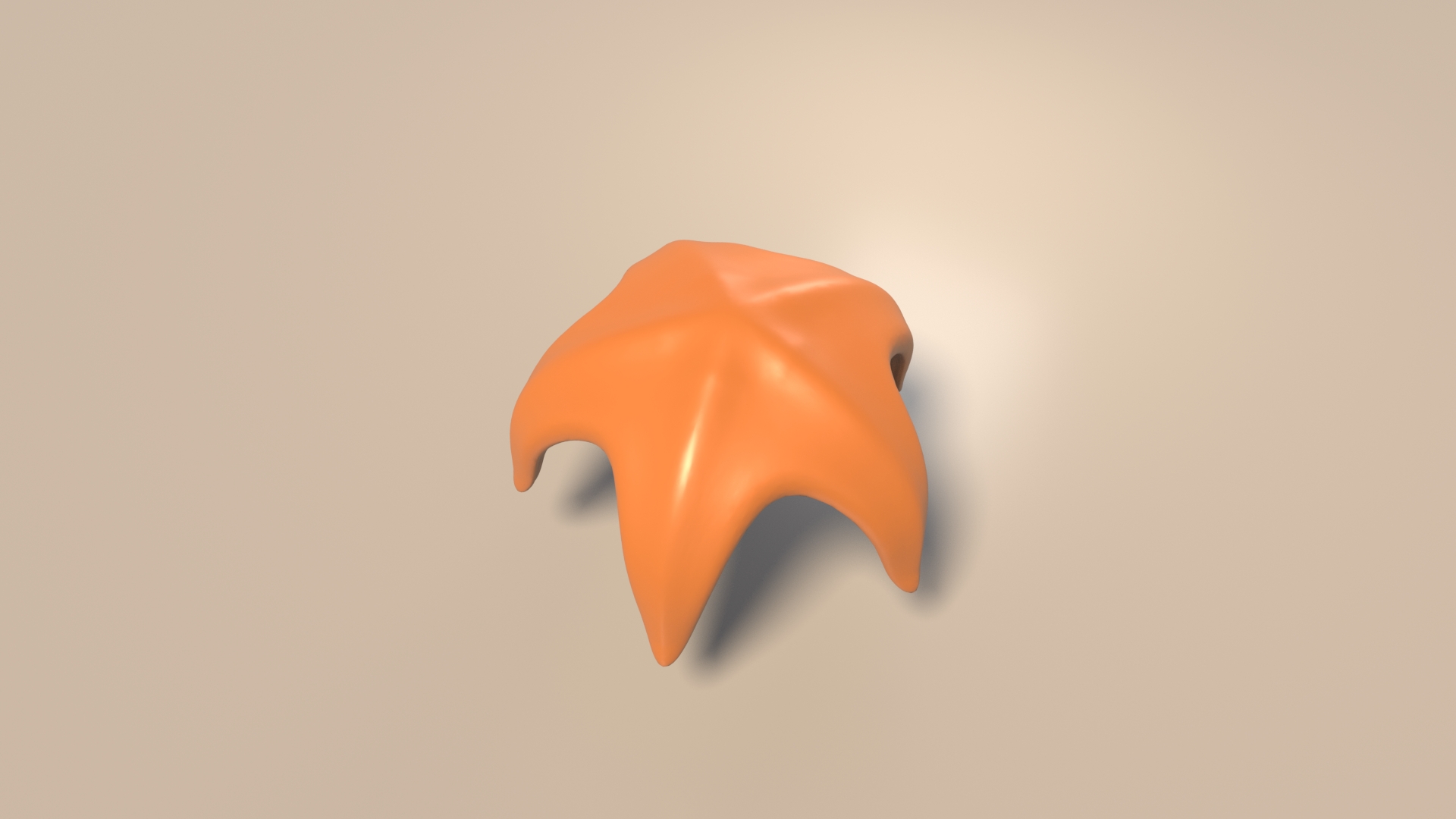} &
\includegraphics[trim=480 0 480 0, clip, width=1\linewidth]{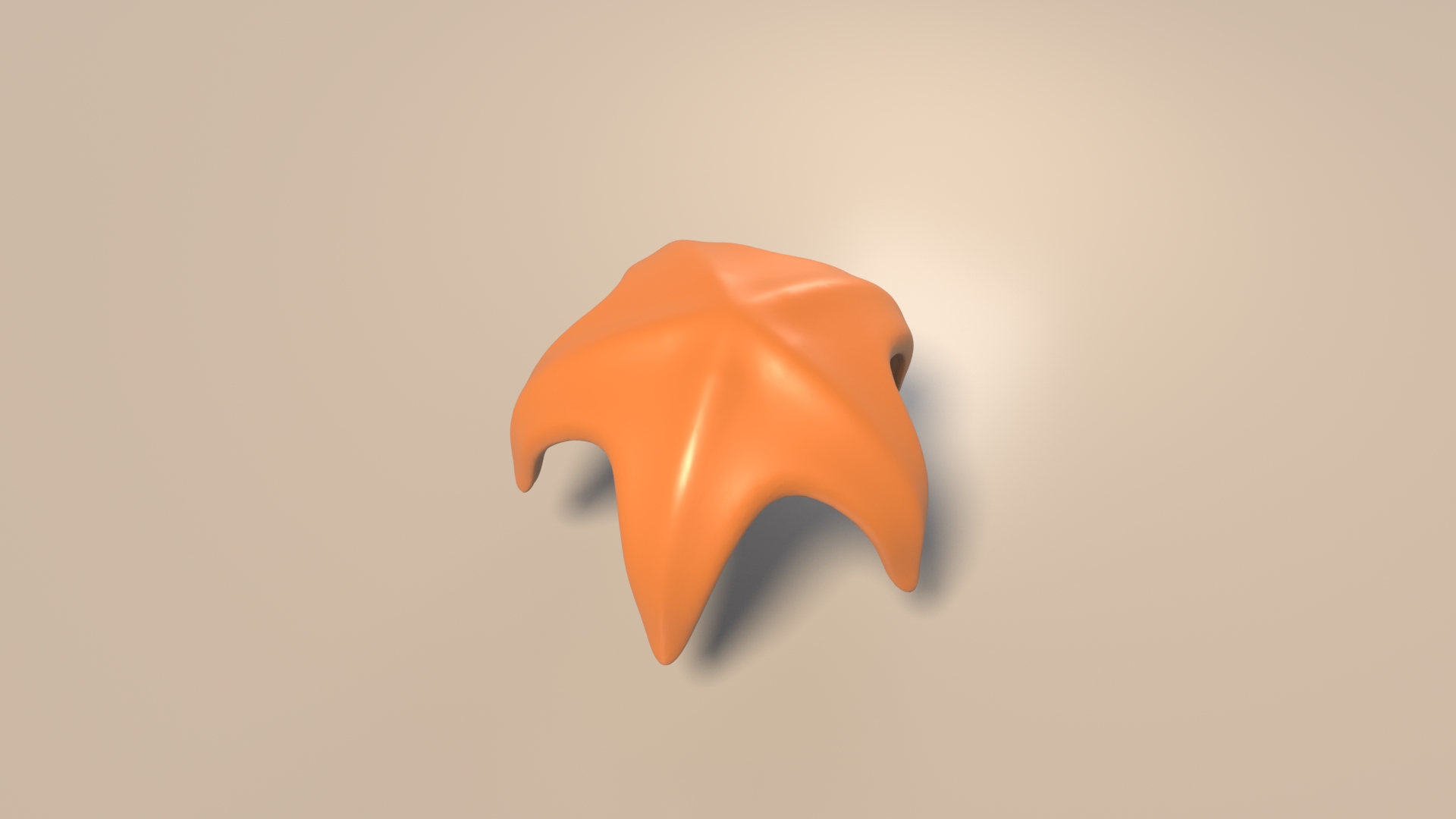} &
\begin{tikzpicture}
\node[inner sep=0pt,anchor=south east] at (0,0)
{ \includegraphics[trim=480 0 480 0, clip, width=1\linewidth]{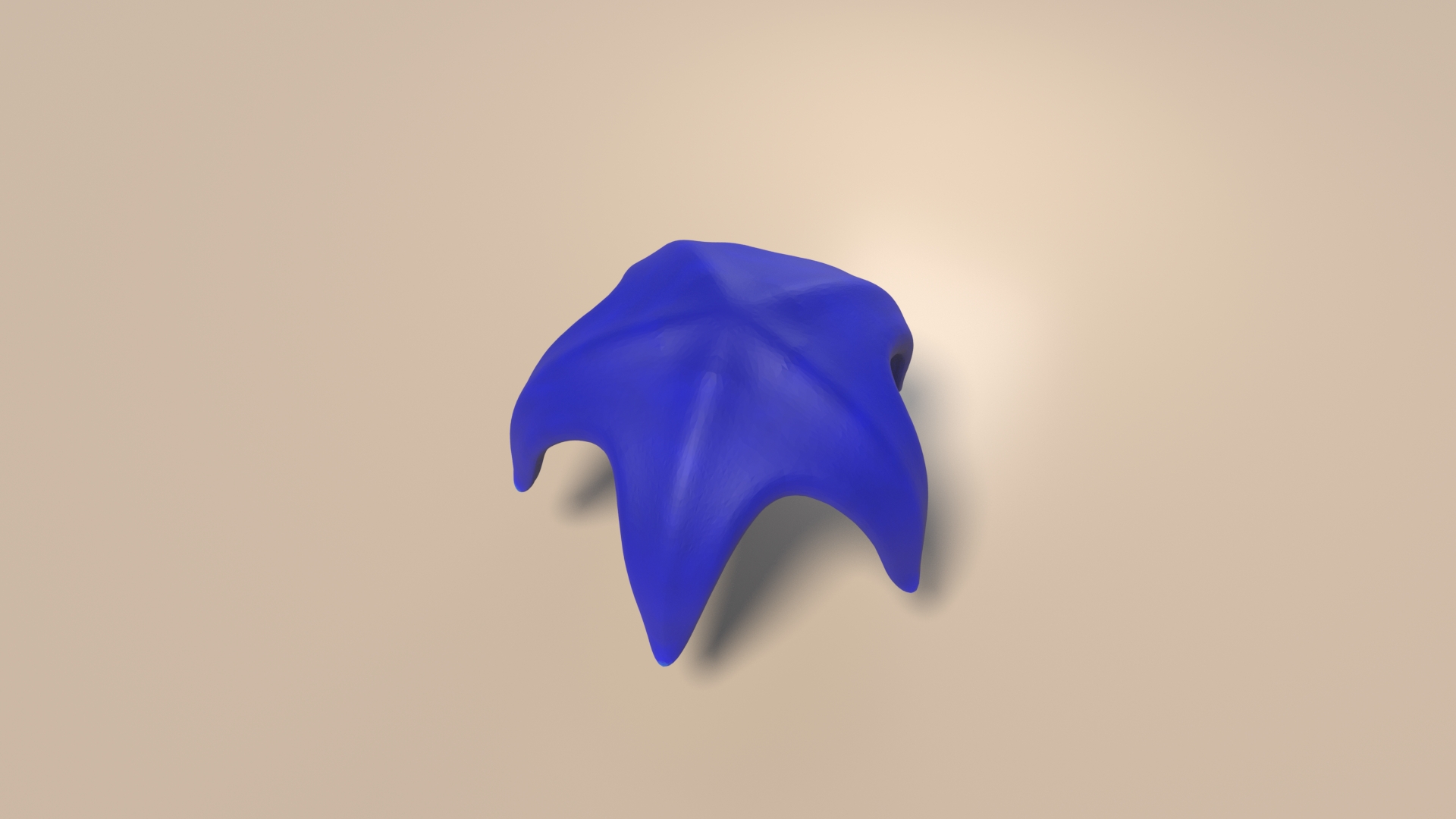} };
\node[inner sep=0pt,anchor=south east] at (-0.2,0.22)
{ \includegraphics[trim=0 0 0 0, clip, width=0.3\linewidth]{fig/cbar} };
\node[inner sep=0pt,anchor=south east] at (0,0.05)
{ \tiny{> 5mm} };
\end{tikzpicture} &
\includegraphics[trim=480 0 480 0, clip, width=1\linewidth]{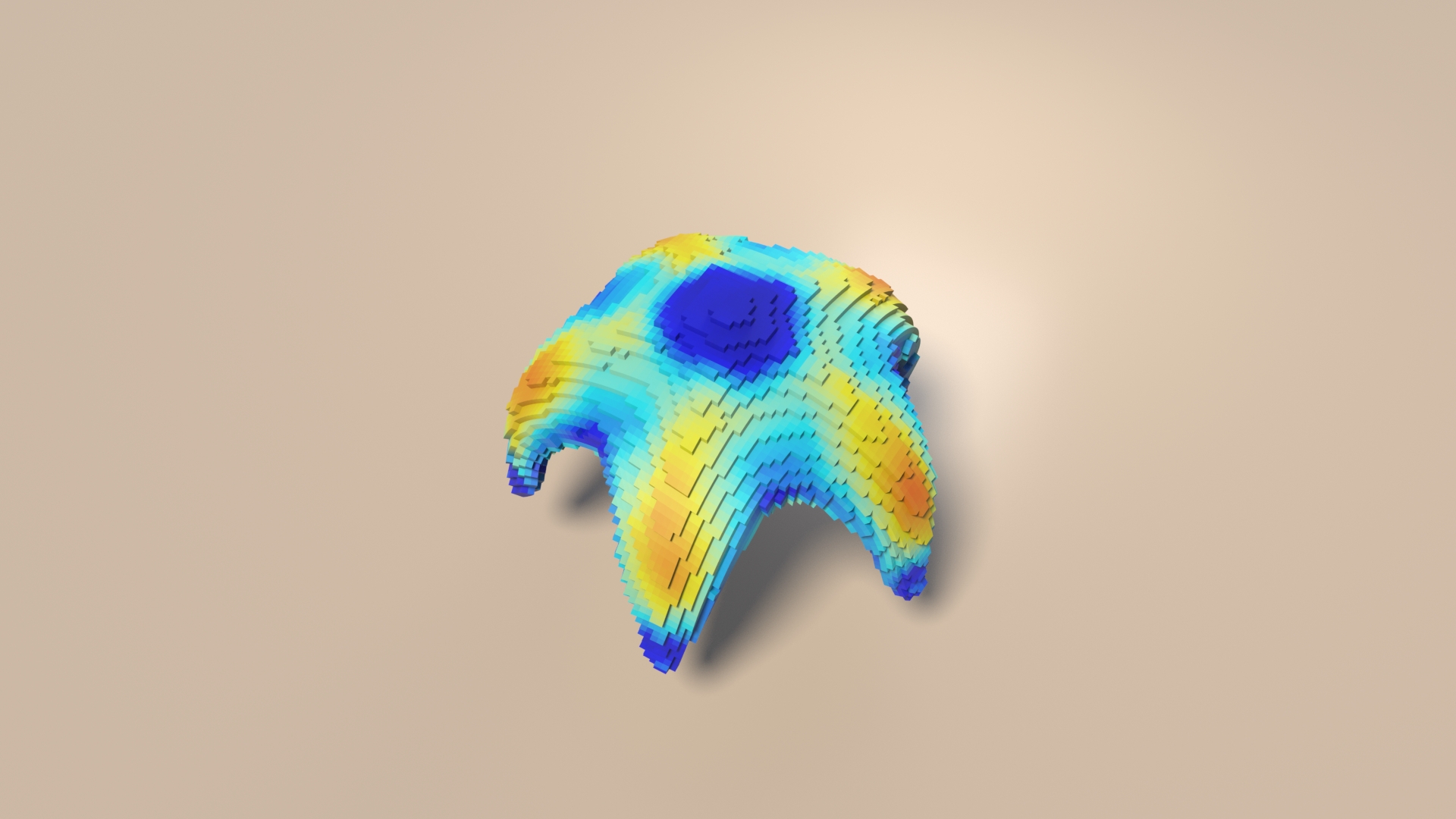}
\\
\includegraphics[trim=480 0 480 0, clip, width=1\linewidth]{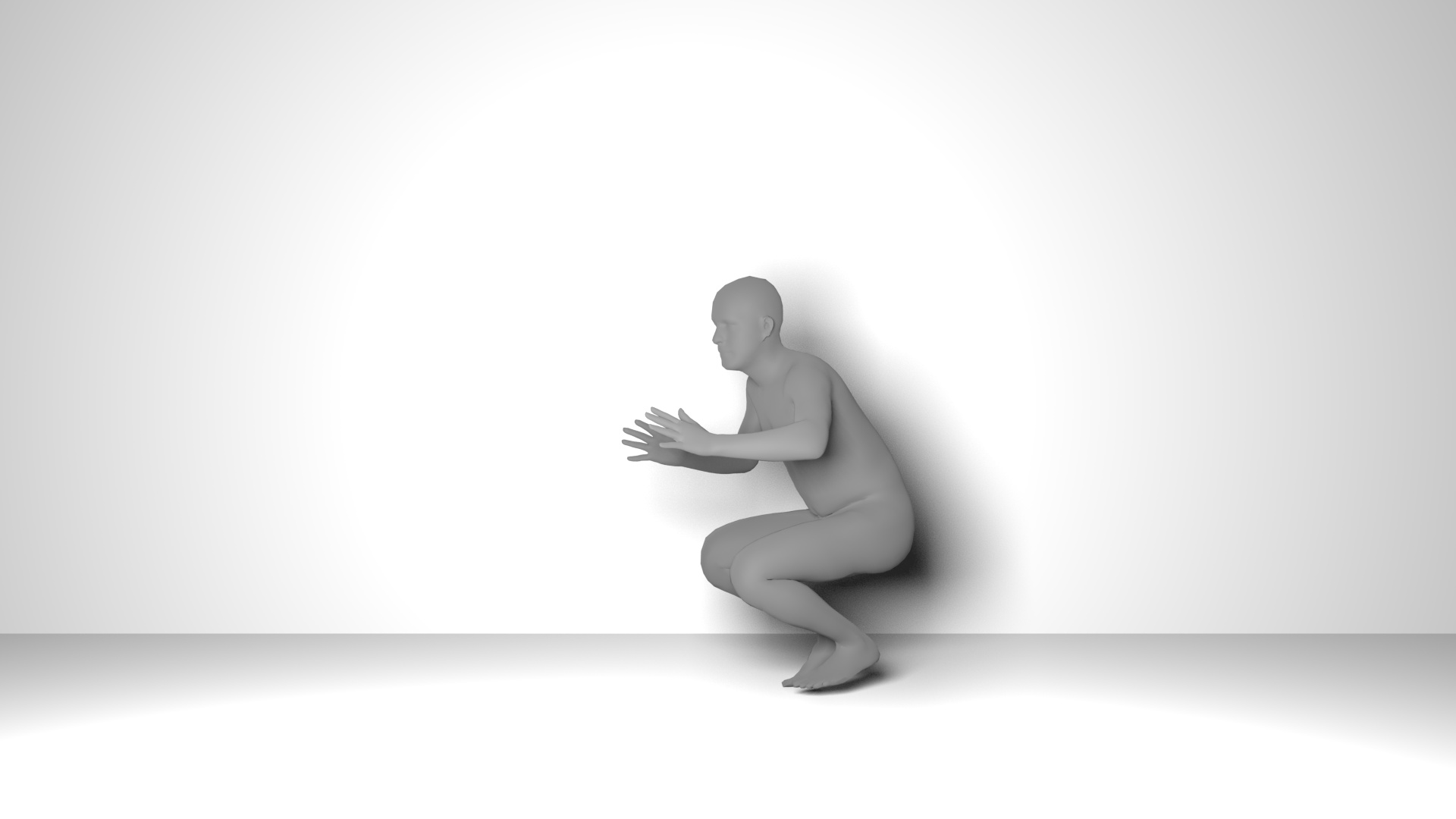} &
\includegraphics[trim=480 0 480 0, clip, width=1\linewidth]{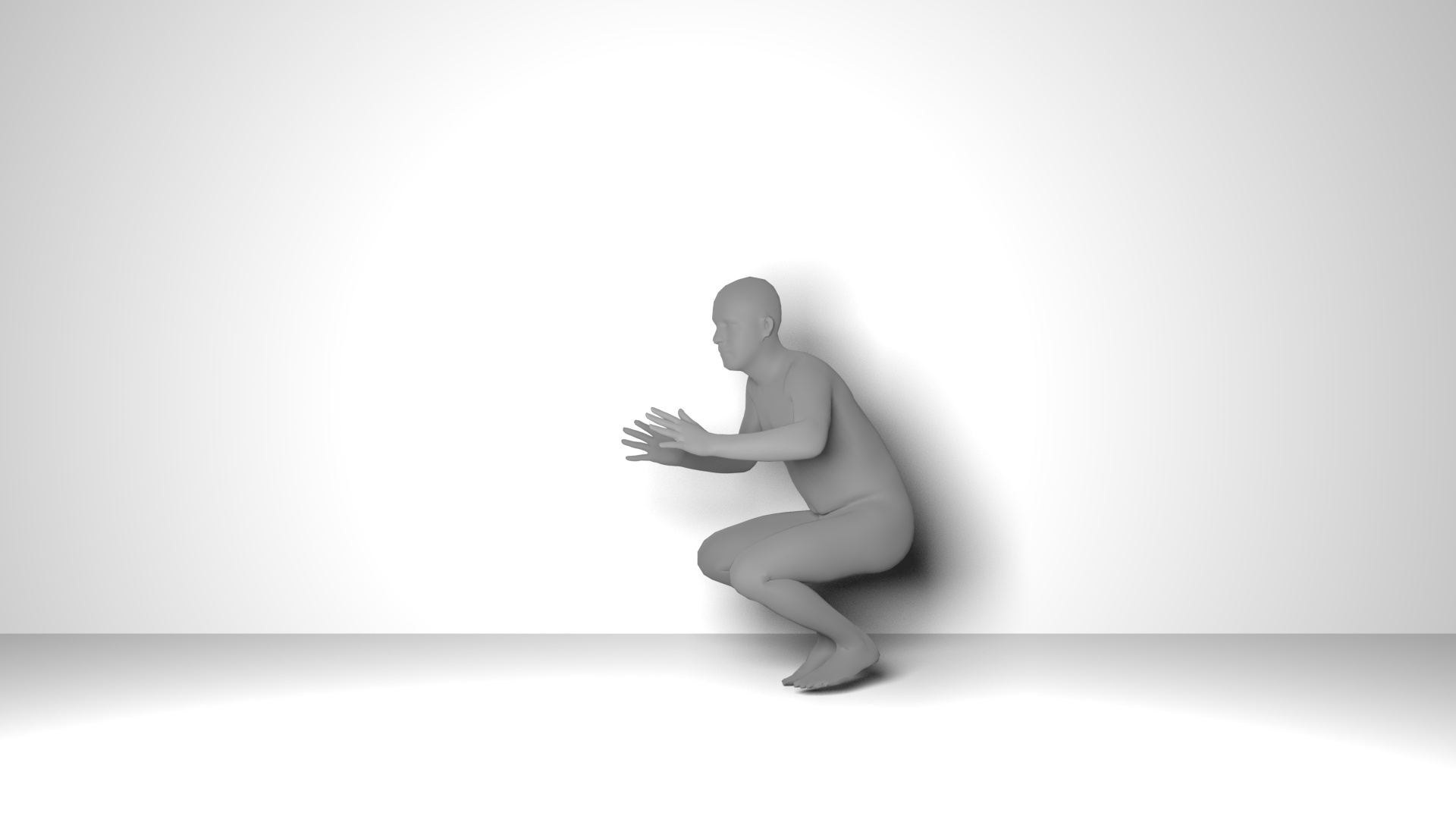} &
\begin{tikzpicture}
\node[inner sep=0pt,anchor=south east] at (0,0)
{ \includegraphics[trim=480 0 480 0, clip, width=1\linewidth]{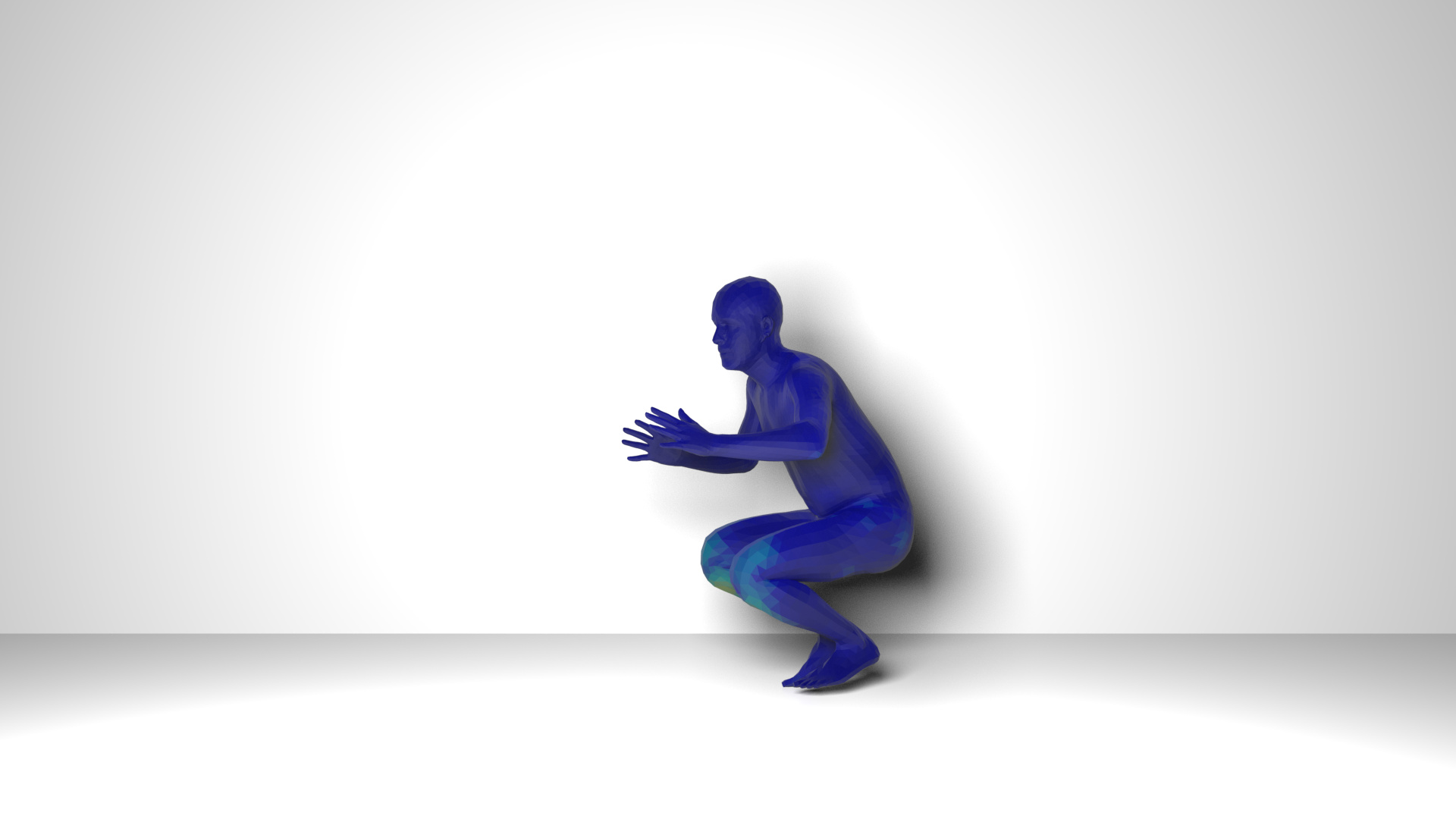} };
\node[inner sep=0pt,anchor=south east] at (-0.2,0.22)
{ \includegraphics[trim=0 0 0 0, clip, width=0.3\linewidth]{fig/cbar} };
\node[inner sep=0pt,anchor=south east] at (0,0.05)
{ \tiny{> 50mm} };
\end{tikzpicture} &
\includegraphics[trim=480 0 480 0, clip, width=1\linewidth]{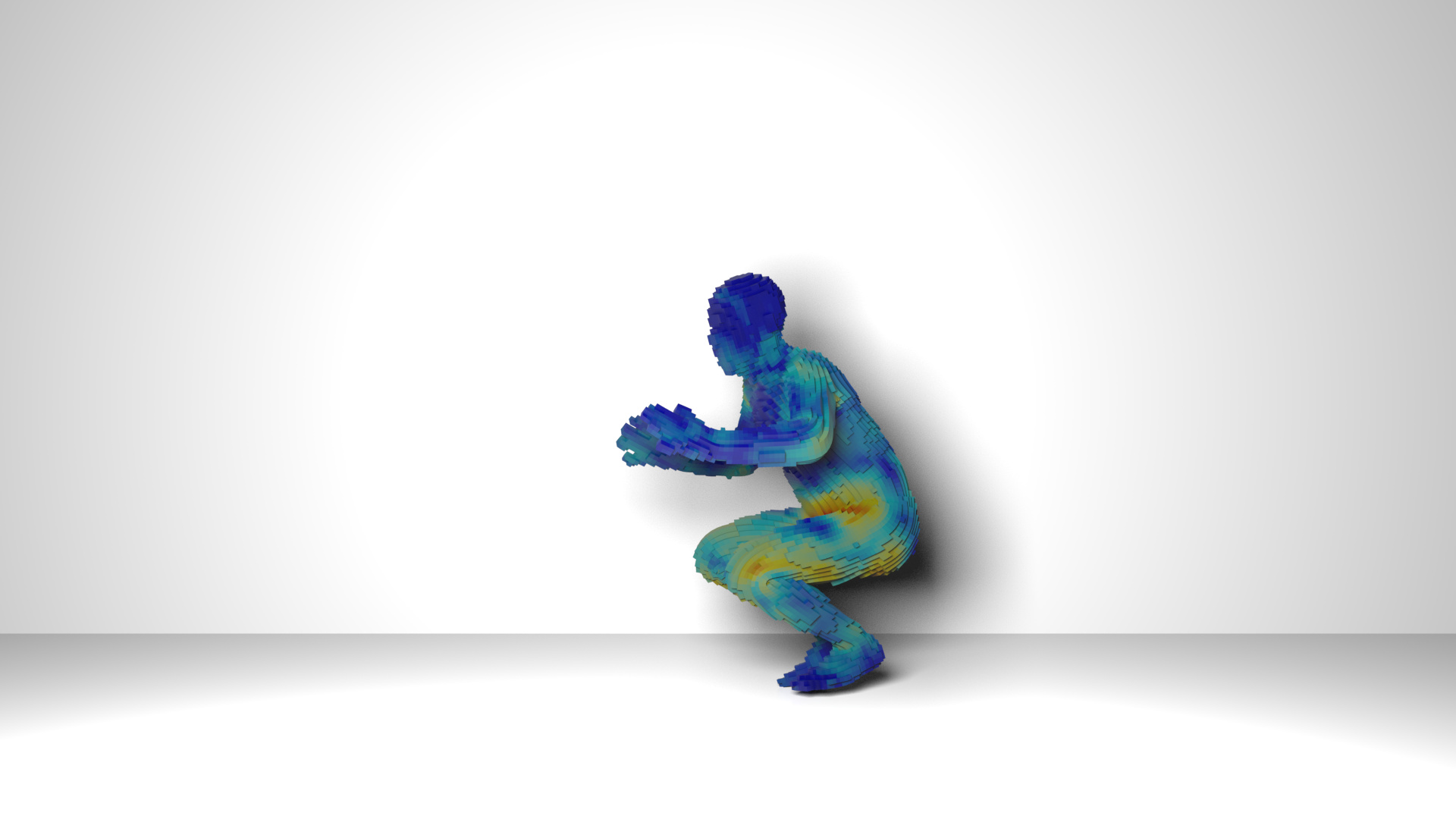}
\\
\includegraphics[trim=480 0 480 0, clip, width=1\linewidth]{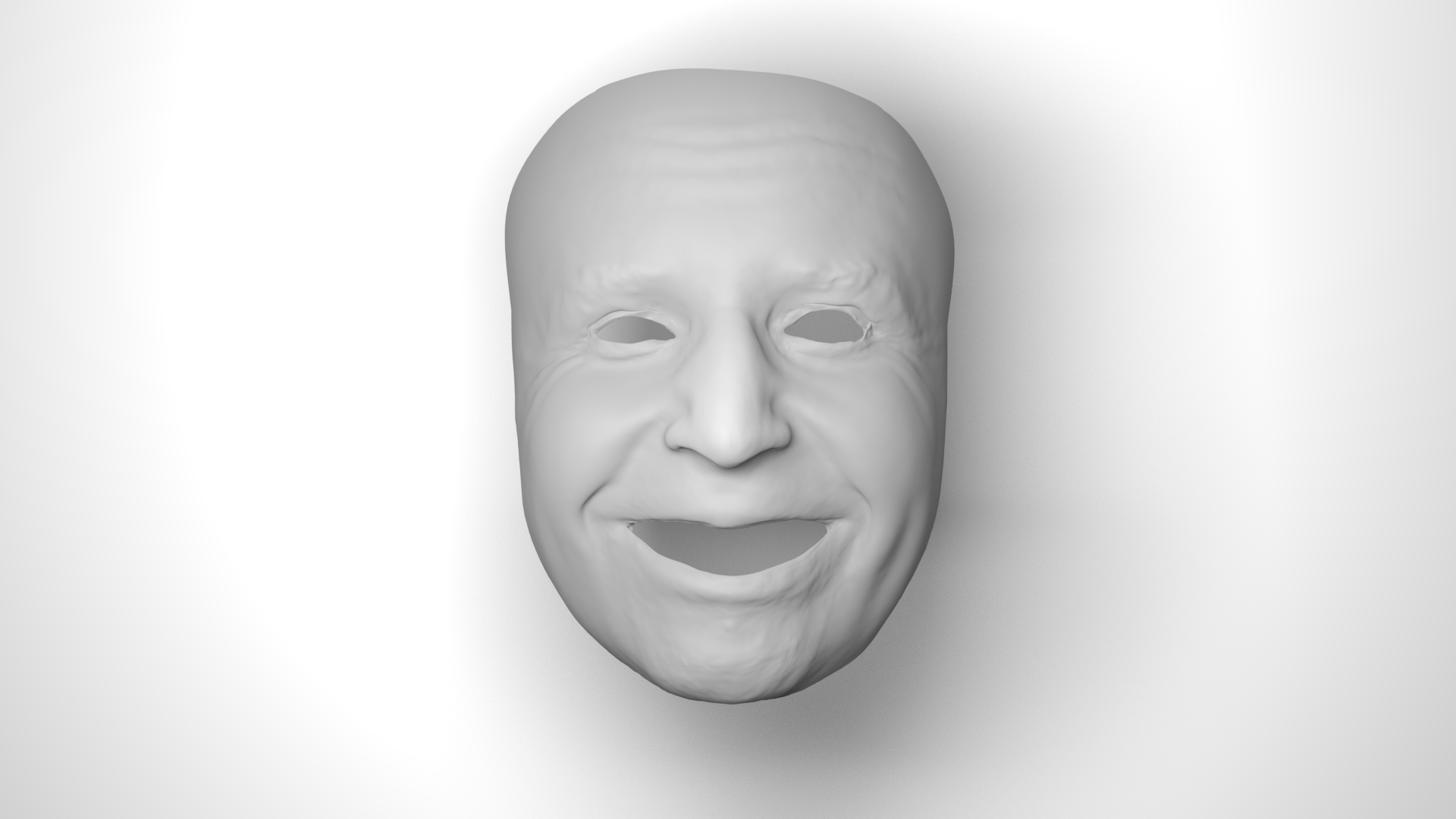} &
\includegraphics[trim=480 0 480 0, clip, width=1\linewidth]{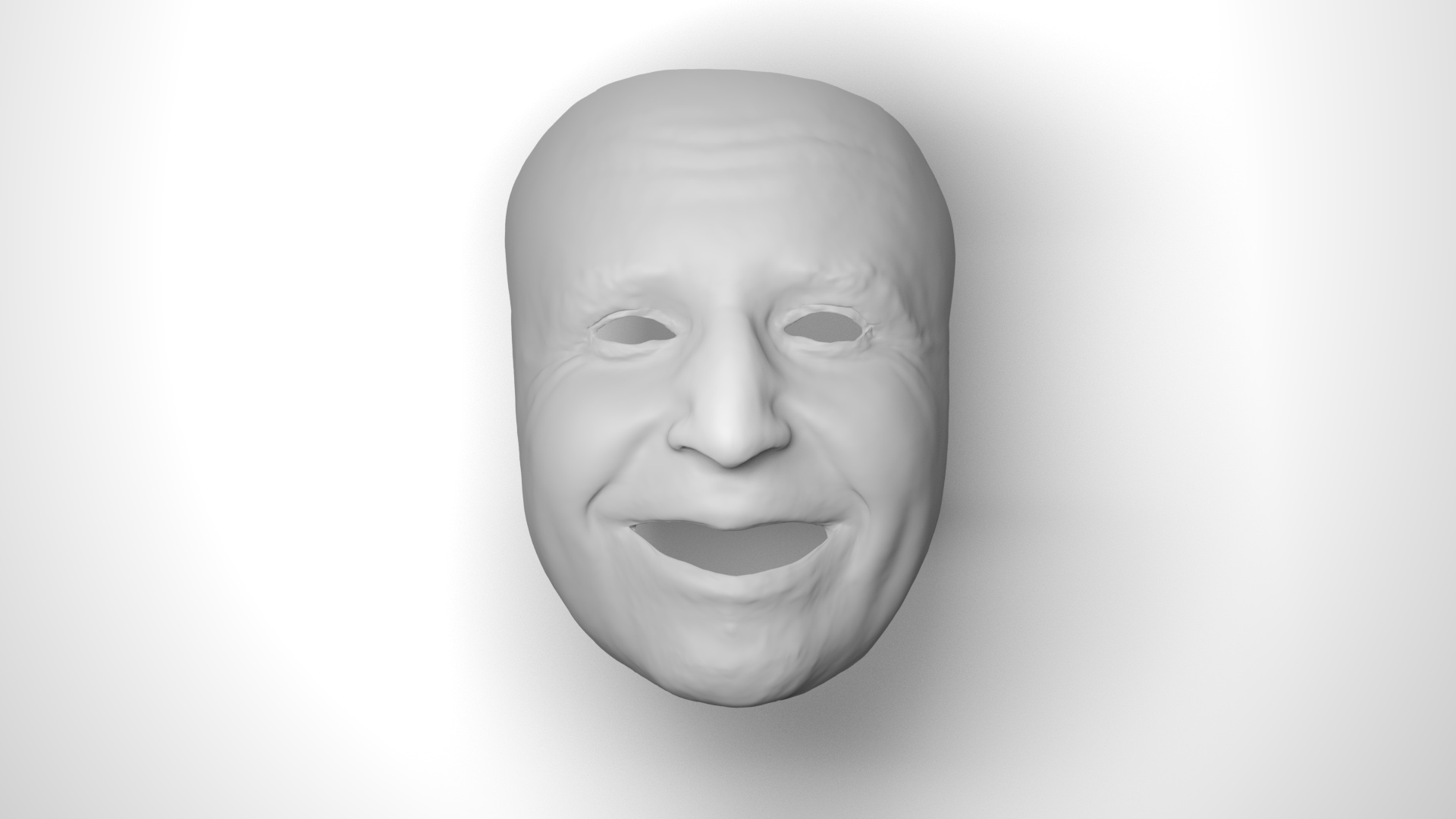} &
\begin{tikzpicture}
\node[inner sep=0pt,anchor=south east] at (0,0)
{ \includegraphics[trim=480 0 480 0, clip, width=1\linewidth]{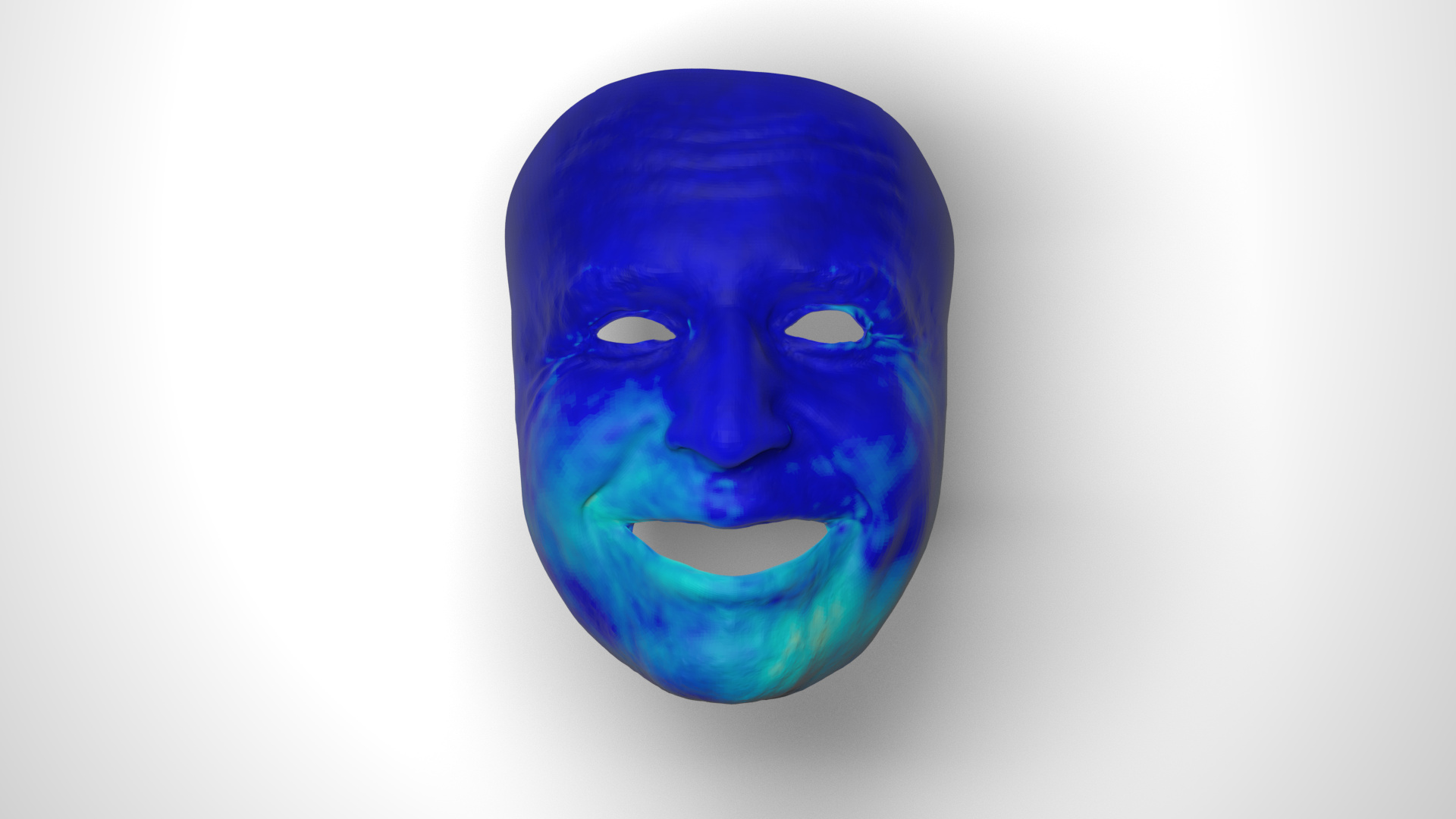} };
\node[inner sep=0pt,anchor=south east] at (-0.2,0.22)
{ \includegraphics[trim=0 0 0 0, clip, width=0.3\linewidth]{fig/cbar} };
\node[inner sep=0pt,anchor=south east] at (0,0.05)
{ \tiny{> 5mm} };
\end{tikzpicture} &
\includegraphics[trim=480 0 480 0, clip, width=1\linewidth]{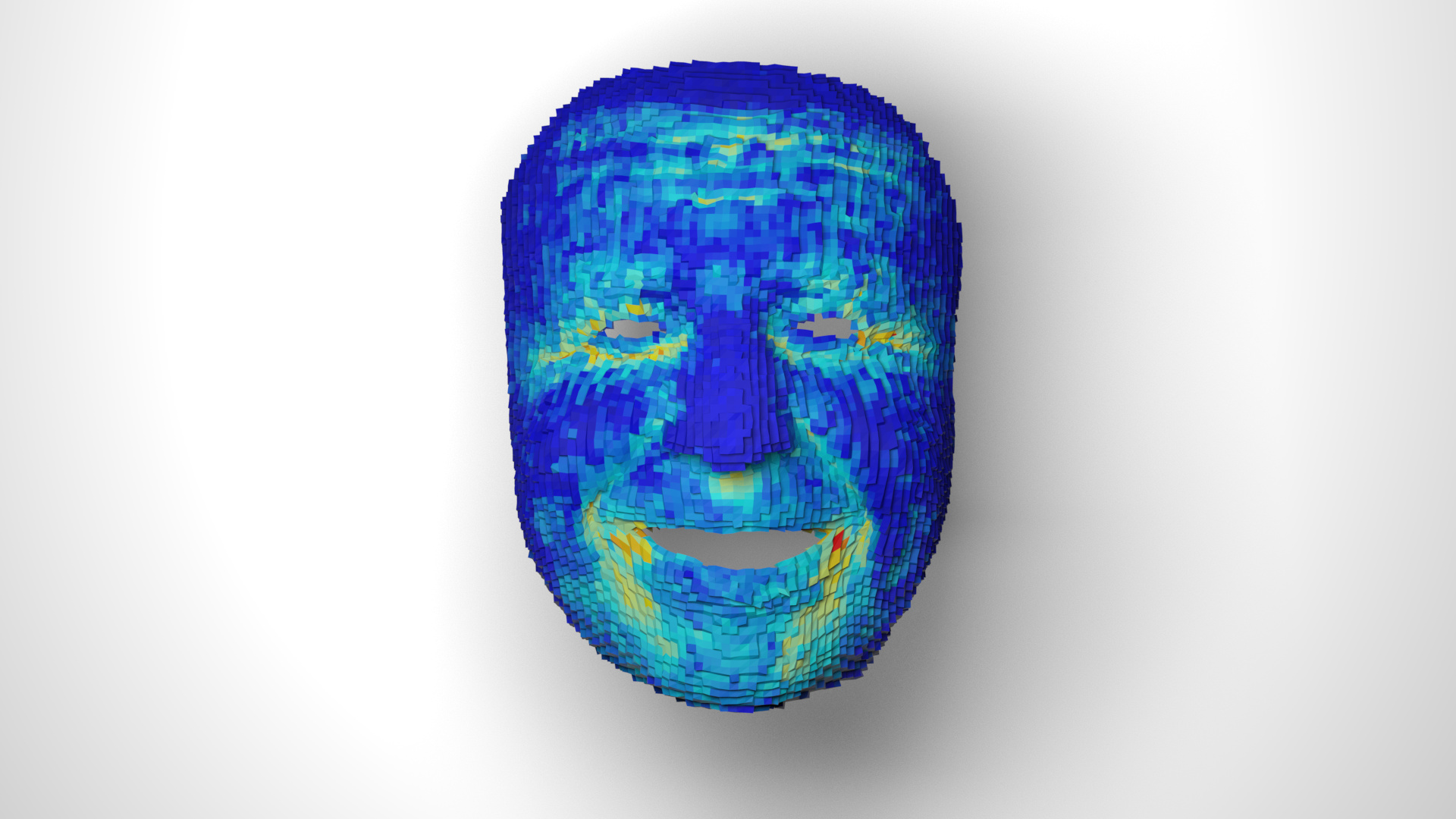}
\\
\vspace{3pt}\small{Target} & \vspace{3pt}\small{Ours} & \vspace{3pt}\small{Error} & \vspace{3pt}\small{Actuation}
\end{tabular}
\caption{Simulation of unseen target poses. From left to right: target pose, simulated result, reconstruction error, optimized actuation magnitudes.}
\label{fig:test}
\end{figure}
\begin{figure*}[ht!] %
\centering
\includegraphics[trim=480 0 480 0, clip, width=0.125\linewidth]{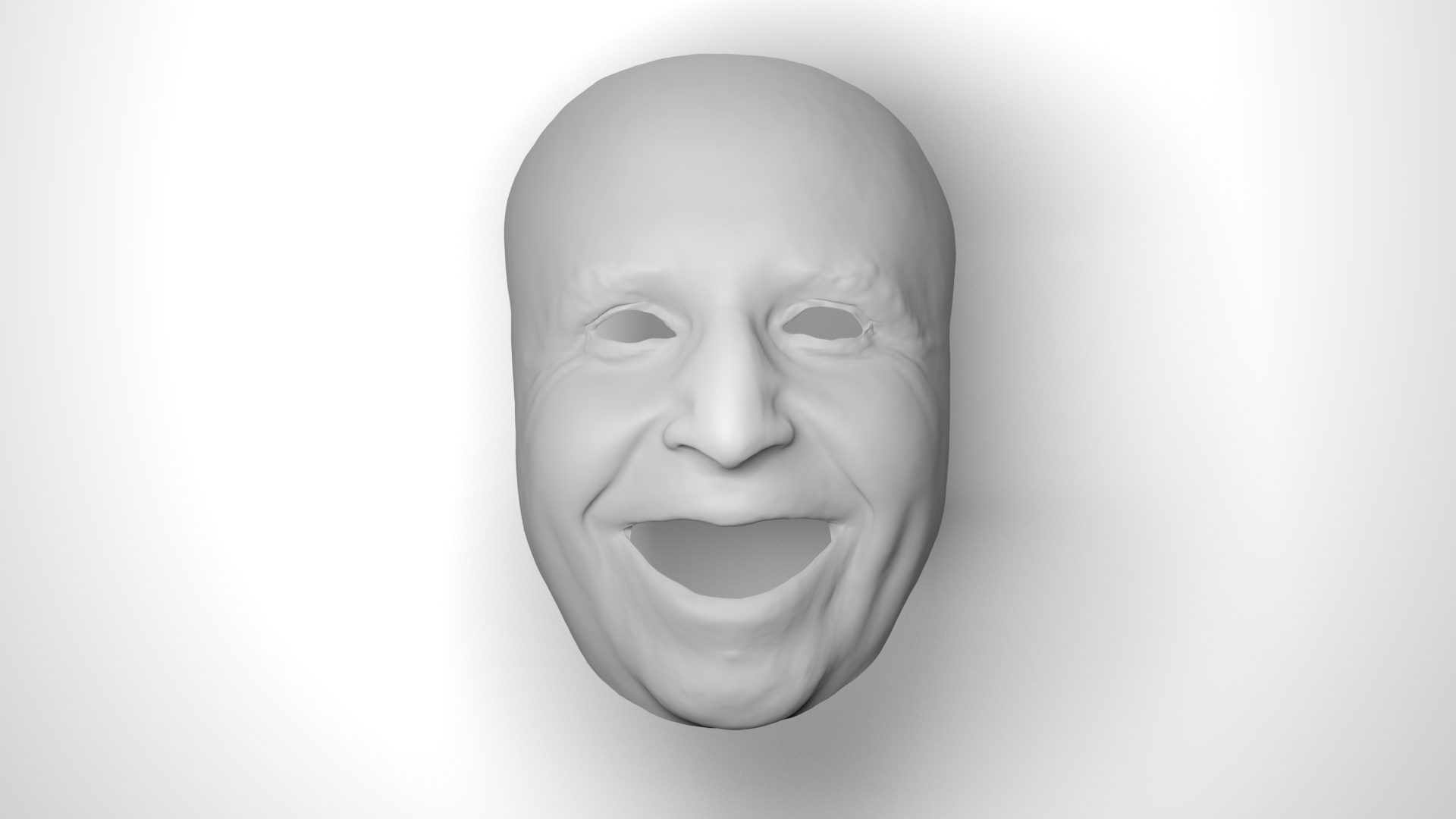}\hspace{-1pt}
\includegraphics[trim=480 0 480 0, clip, width=0.125\linewidth]{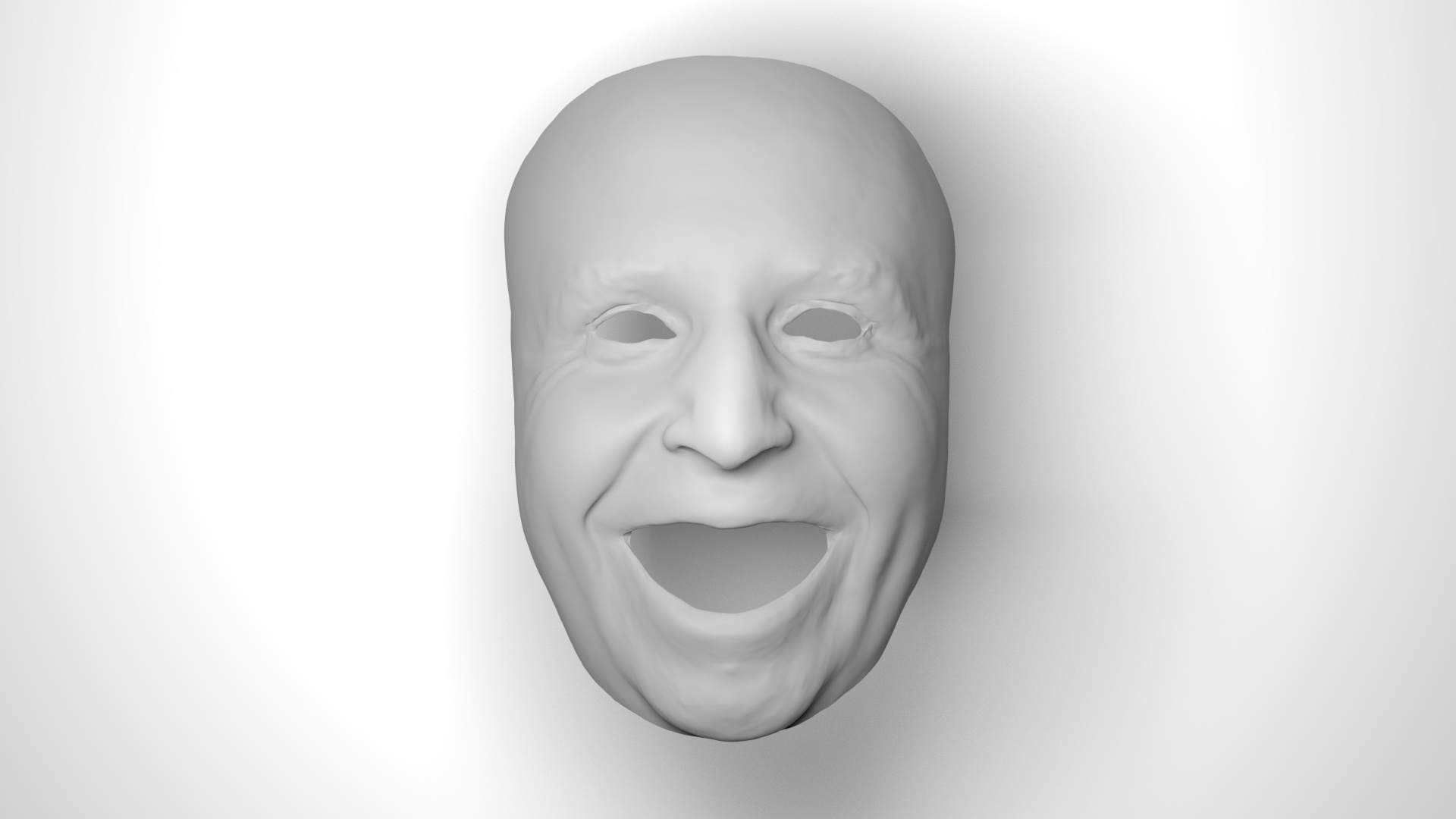}\hspace{-3pt}
\includegraphics[trim=480 0 480 0, clip, width=0.125\linewidth]{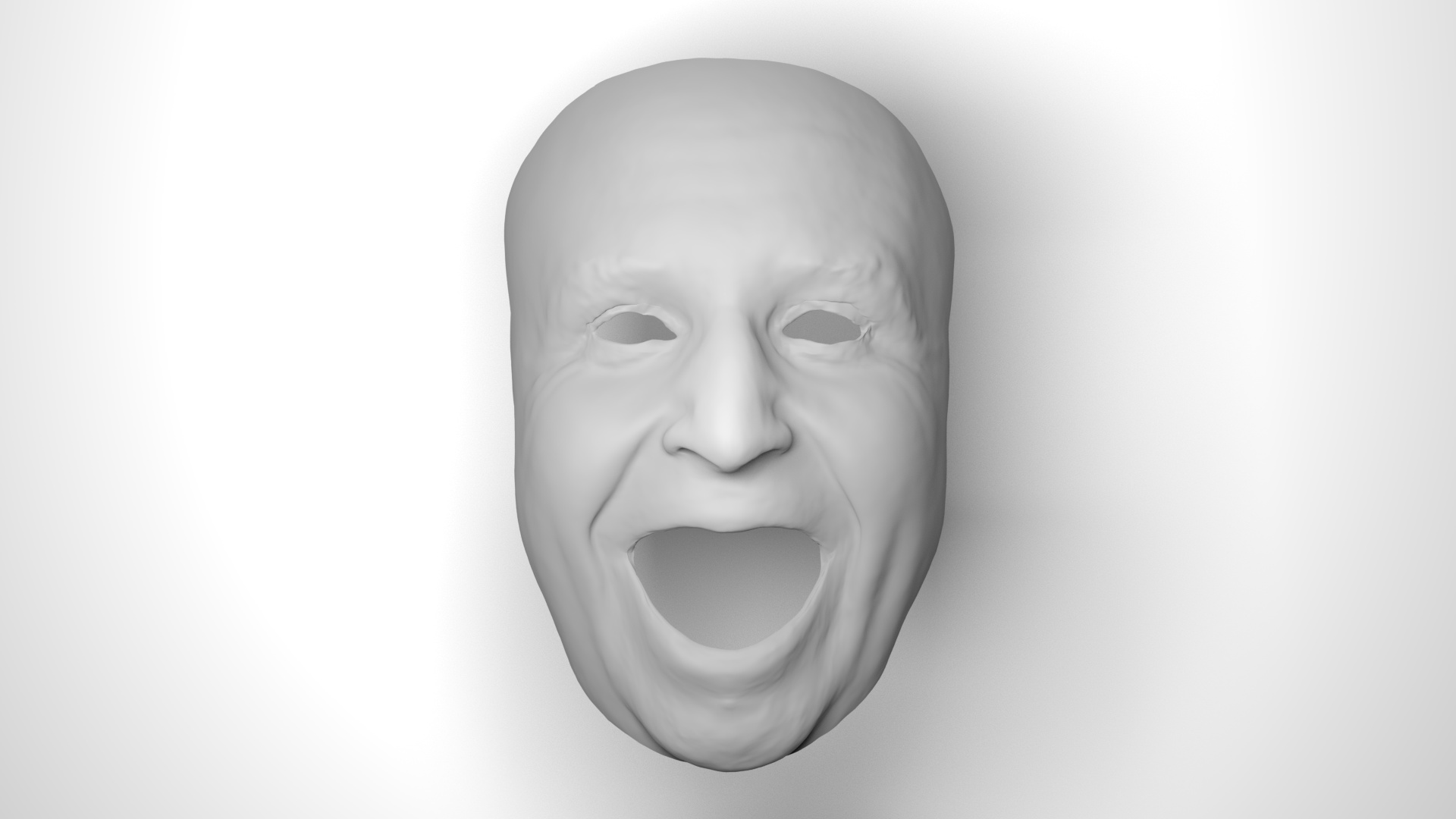}\hspace{-3pt}
\includegraphics[trim=480 0 480 0, clip, width=0.125\linewidth]{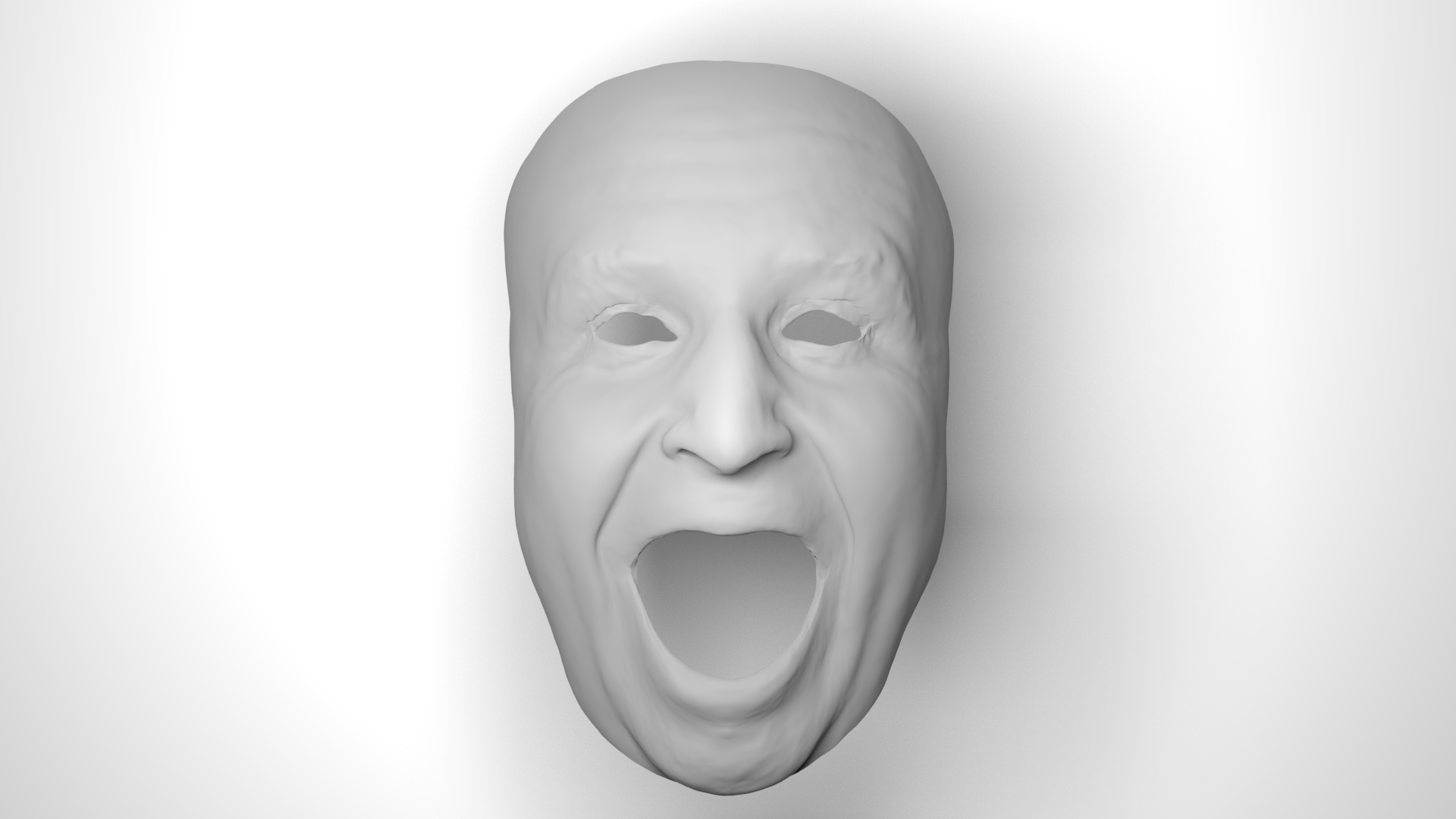}\hspace{-3pt}
\includegraphics[trim=480 0 480 0, clip, width=0.125\linewidth]{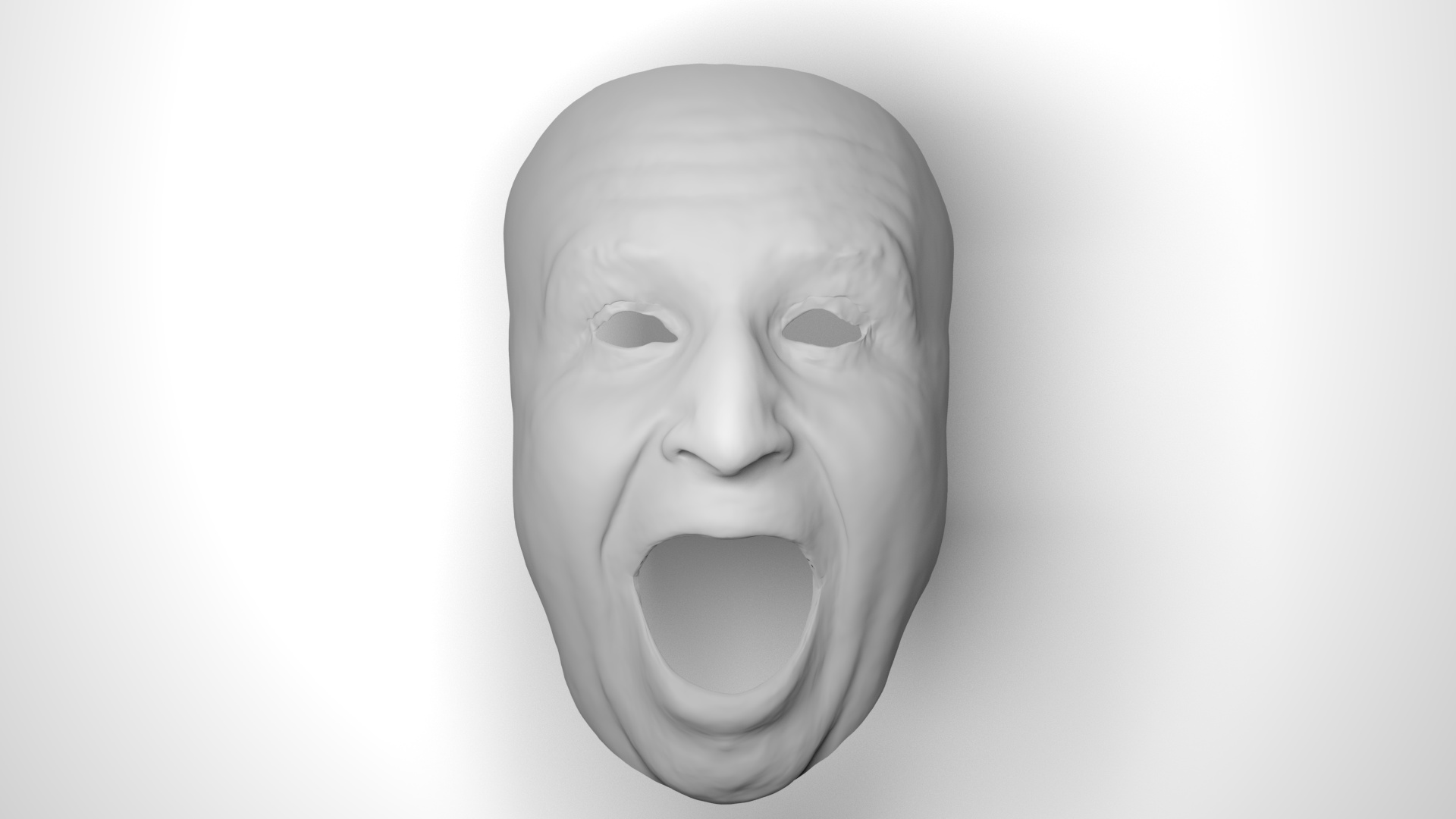}\hspace{-3pt}
\includegraphics[trim=480 0 480 0, clip, width=0.125\linewidth]{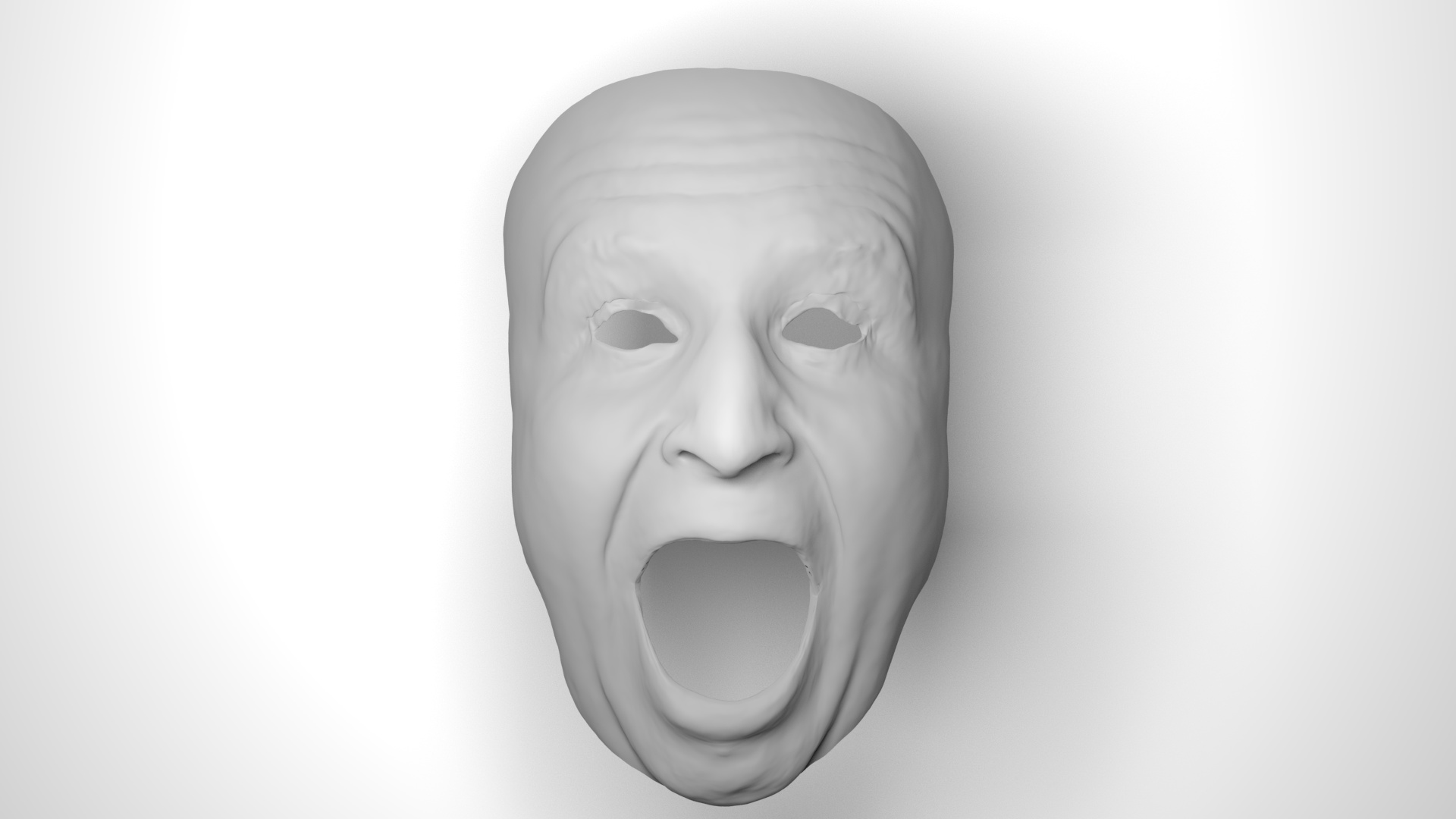}\hspace{-3pt}
\includegraphics[trim=480 0 480 0, clip, width=0.125\linewidth]{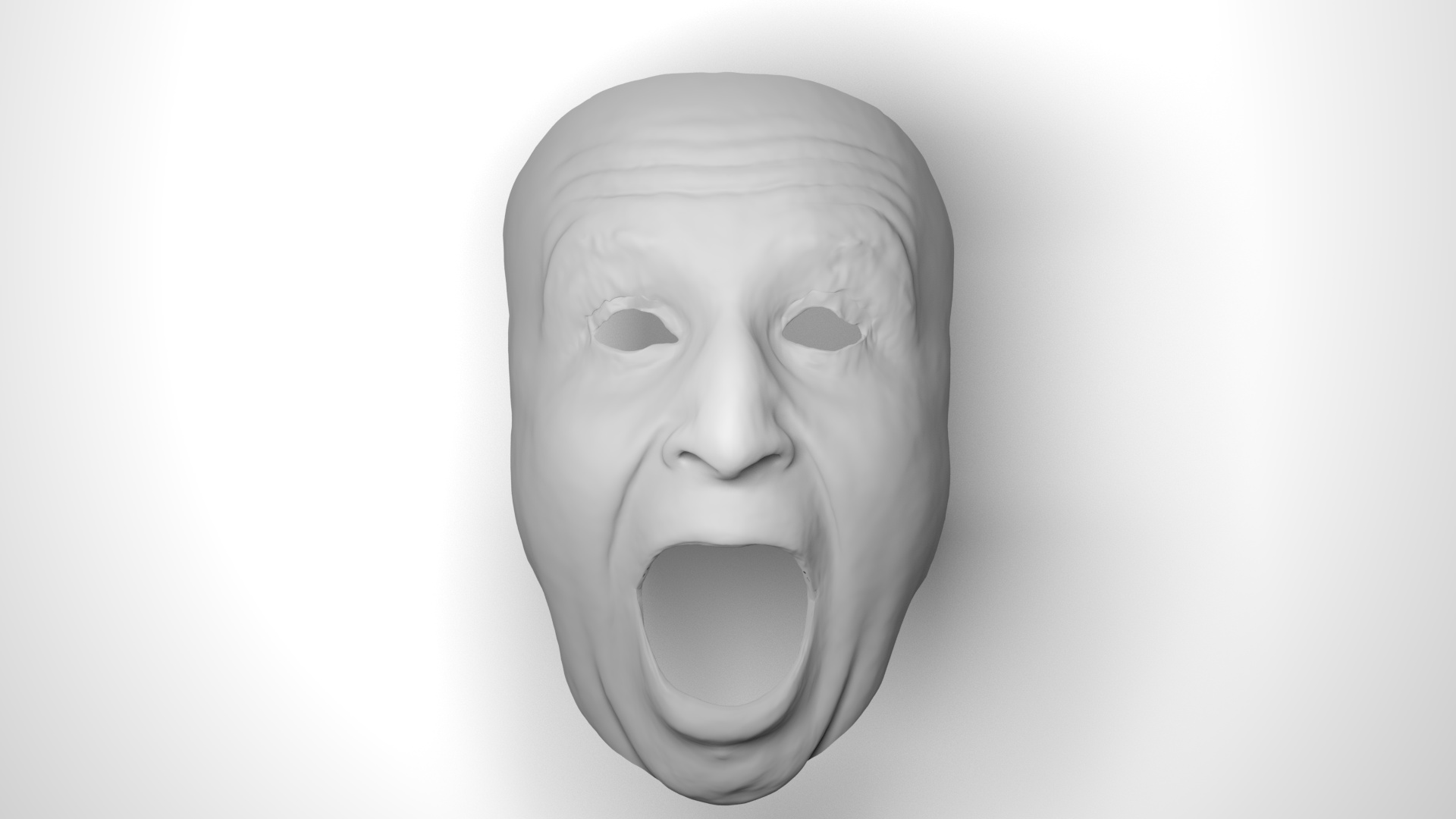}\hspace{-1pt}
\includegraphics[trim=480 0 480 0, clip, width=0.125\linewidth]{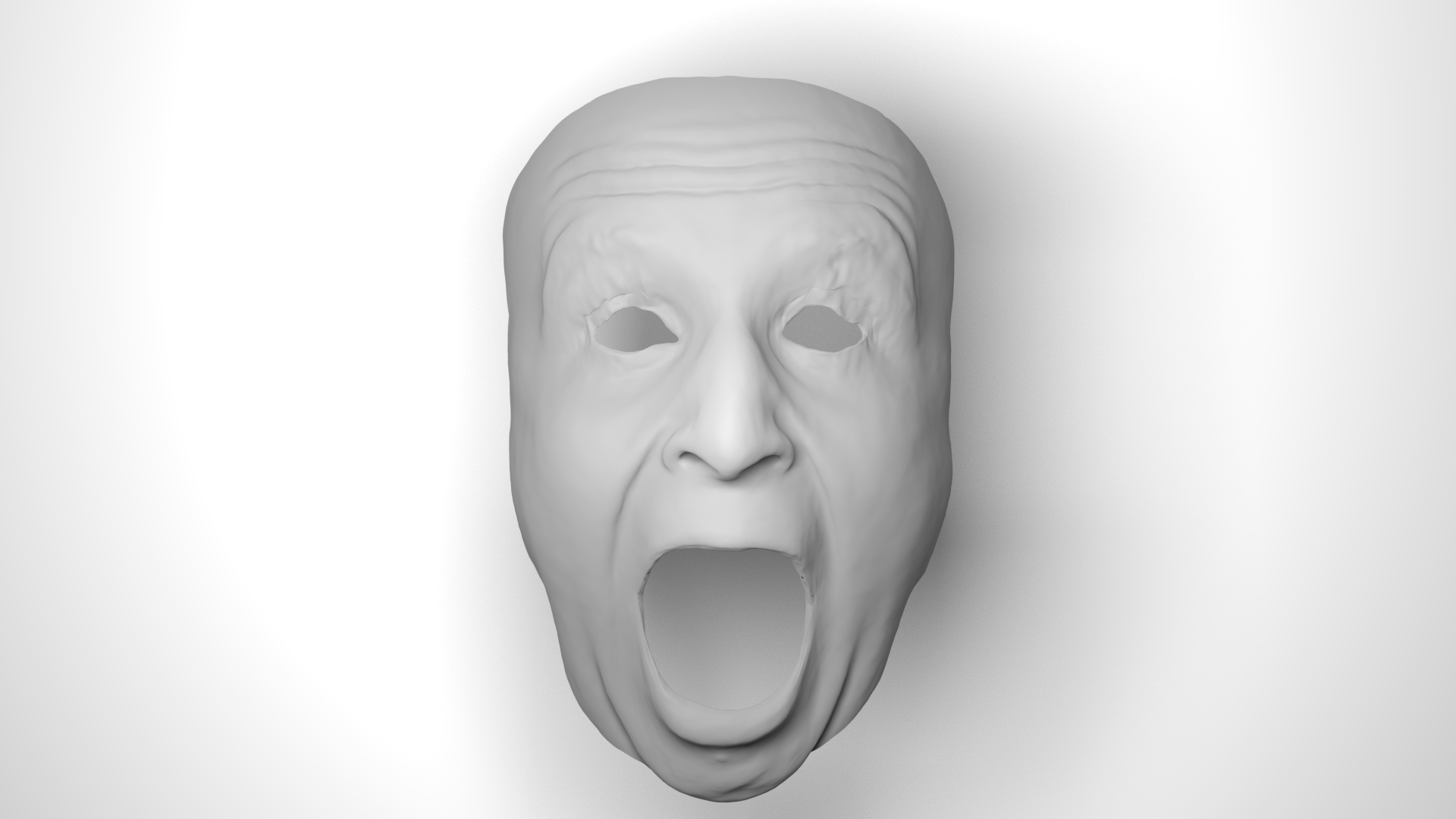}\\
\includegraphics[trim=480 0 480 0, clip, width=0.125\linewidth]{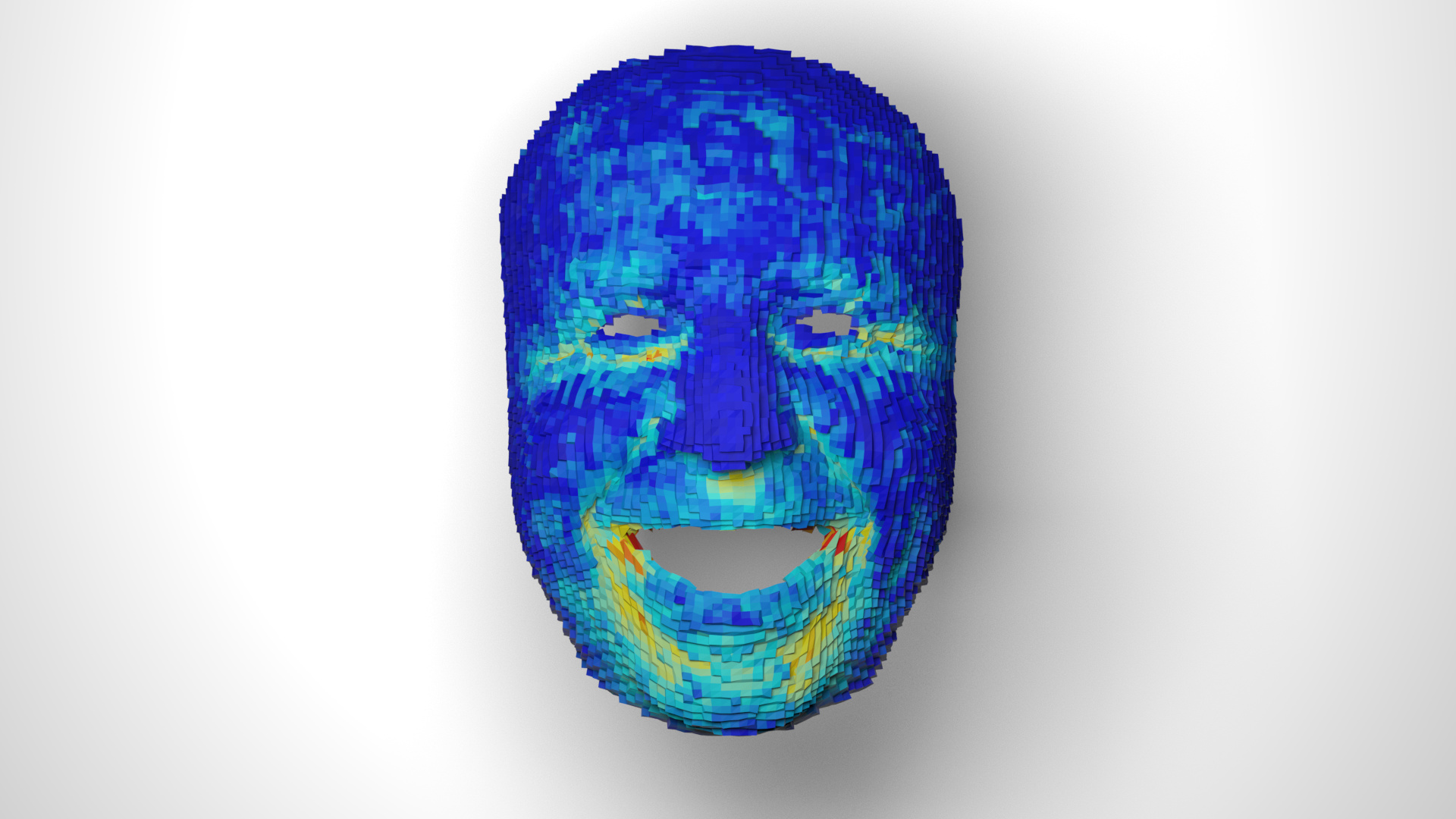}\hspace{-1pt}
\includegraphics[trim=480 0 480 0, clip, width=0.125\linewidth]{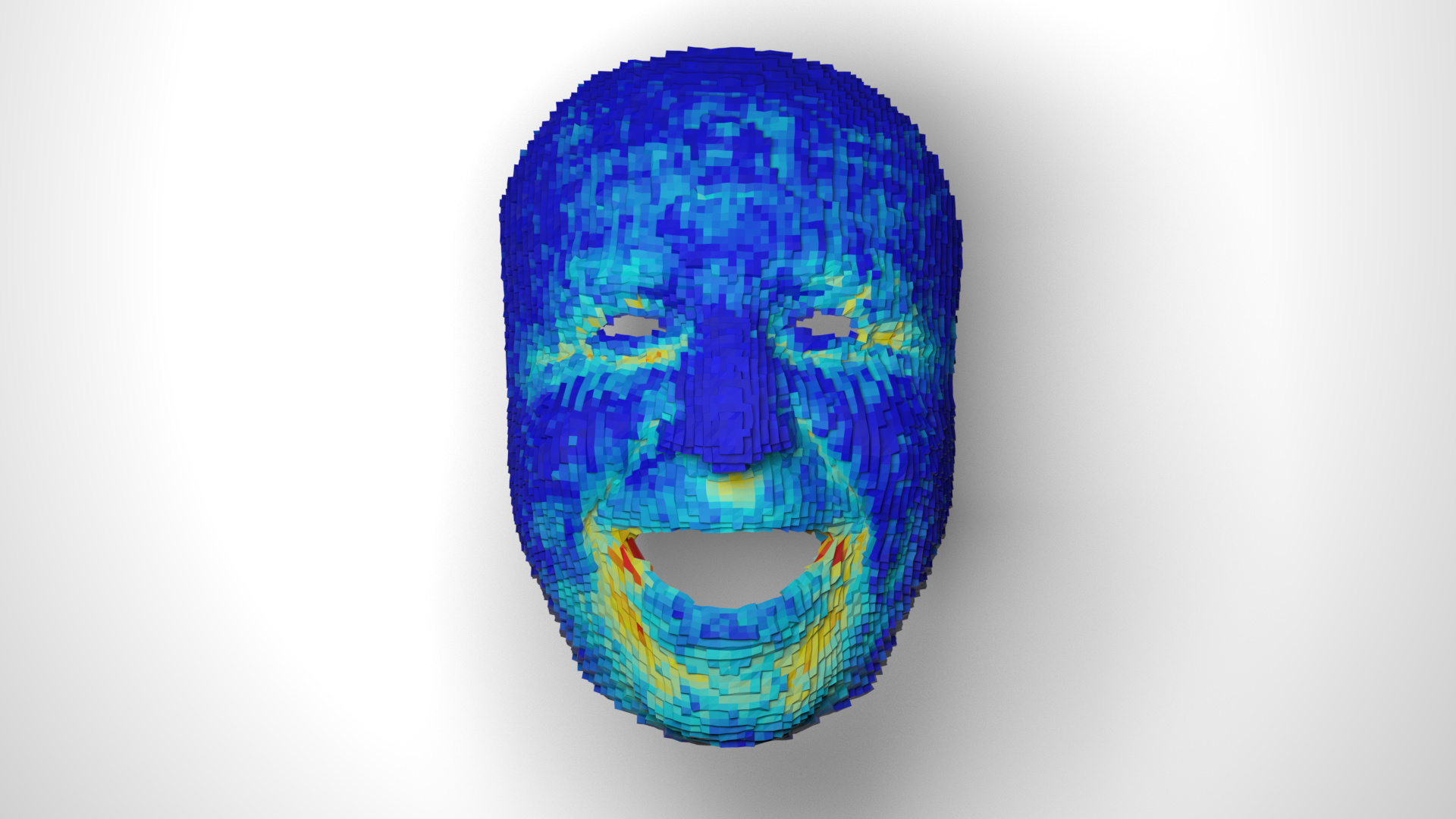}\hspace{-3pt}
\includegraphics[trim=480 0 480 0, clip, width=0.125\linewidth]{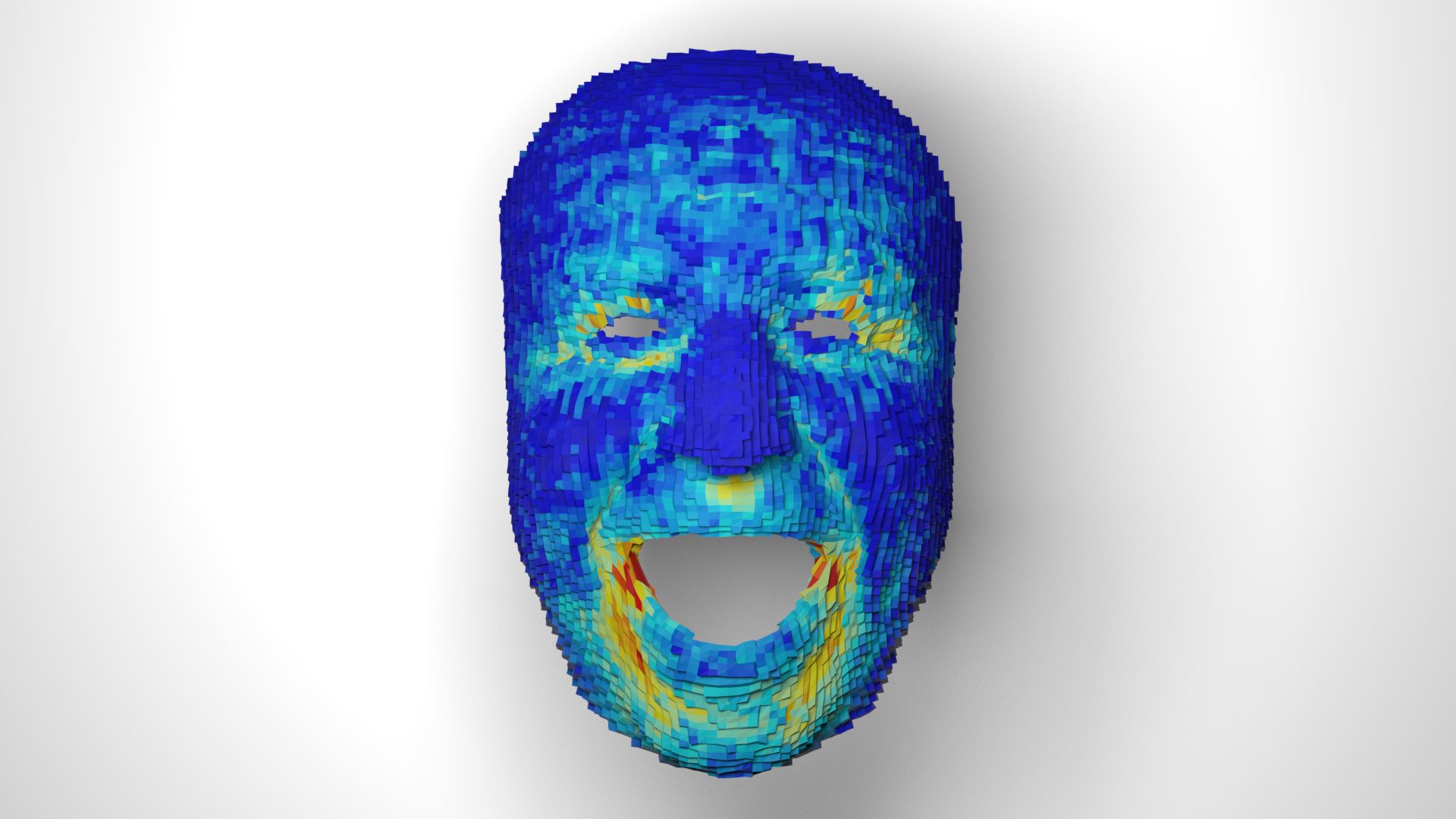}\hspace{-3pt}
\includegraphics[trim=480 0 480 0, clip, width=0.125\linewidth]{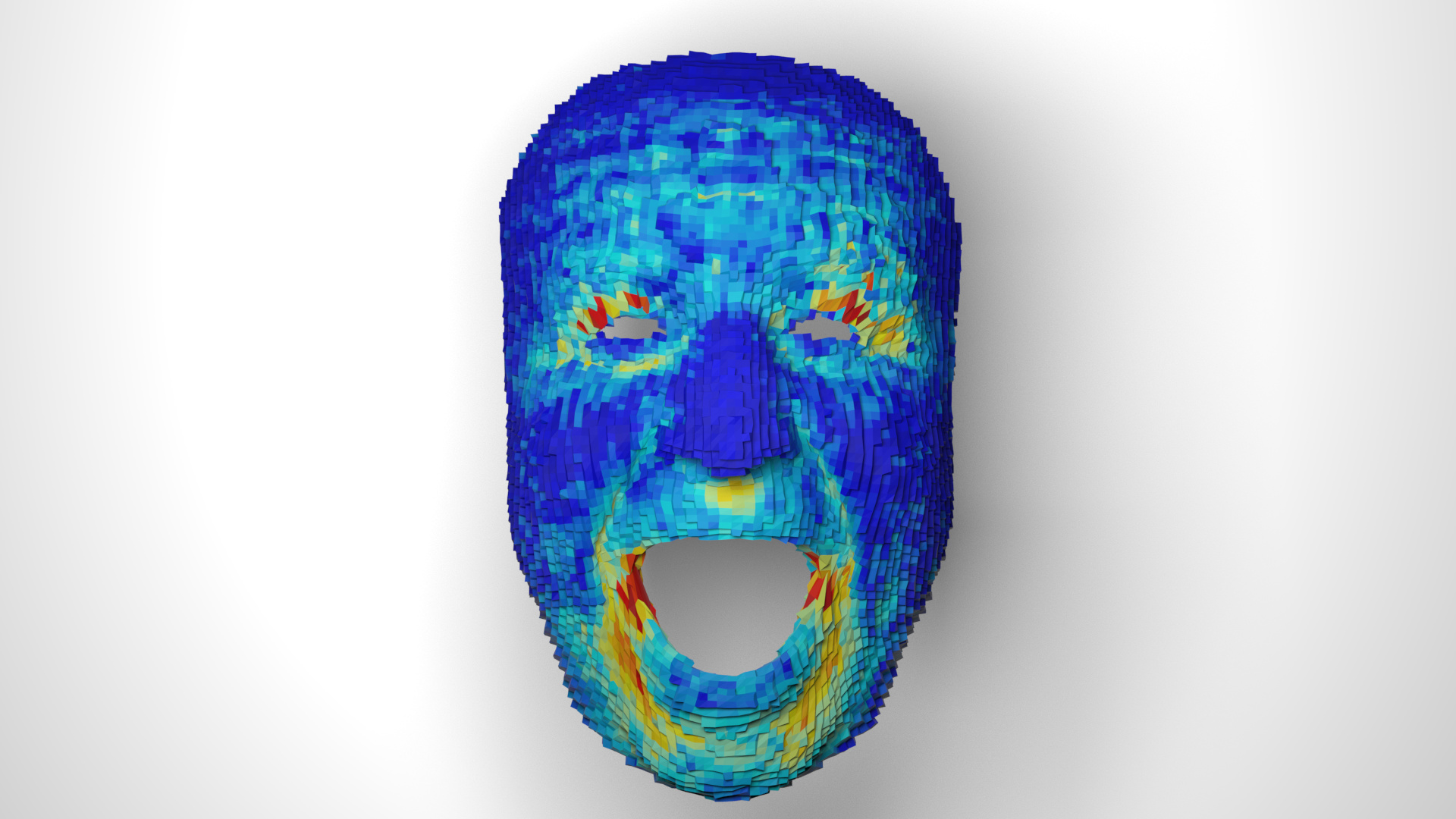}\hspace{-3pt}
\includegraphics[trim=480 0 480 0, clip, width=0.125\linewidth]{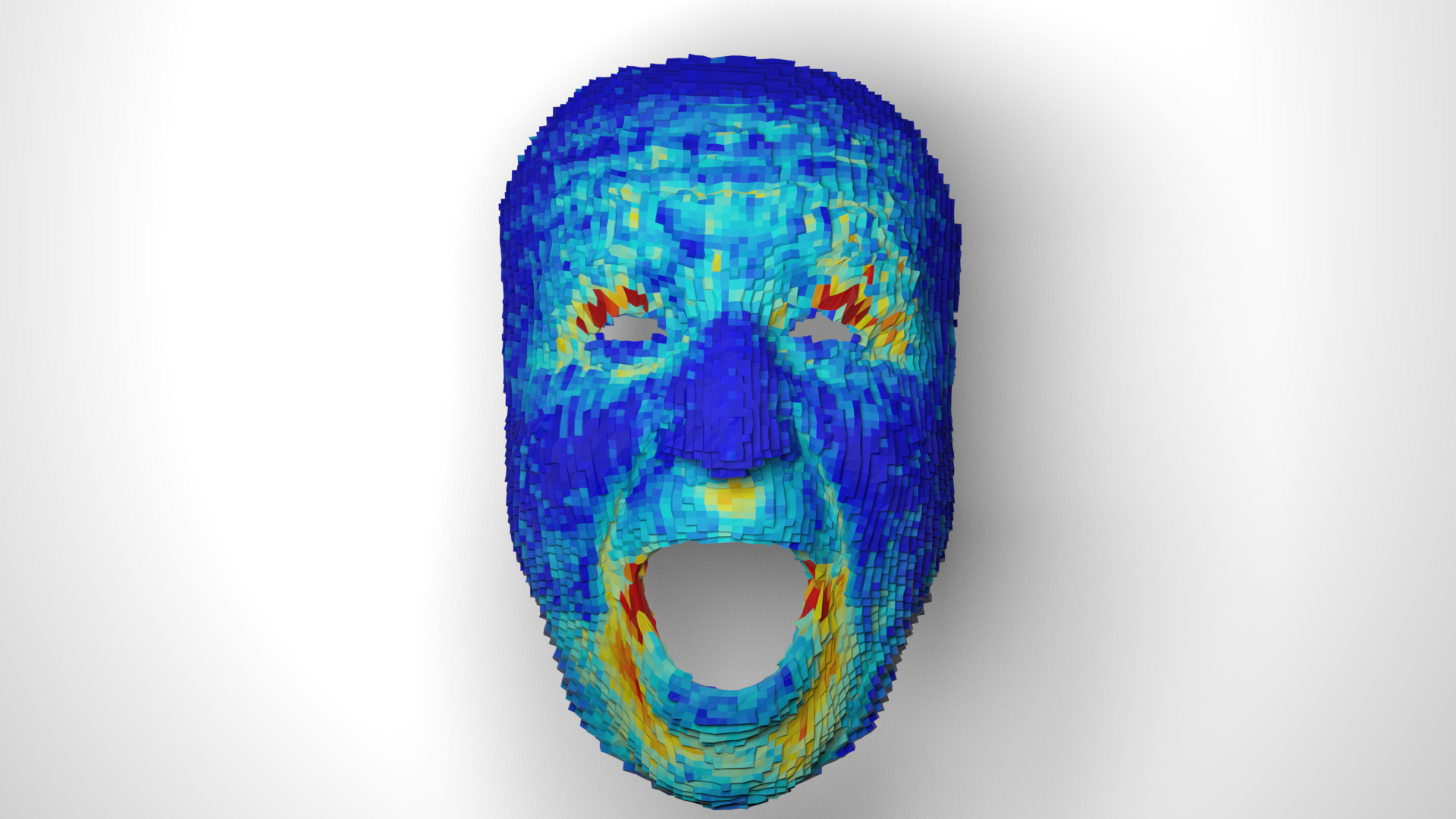}\hspace{-3pt}
\includegraphics[trim=480 0 480 0, clip, width=0.125\linewidth]{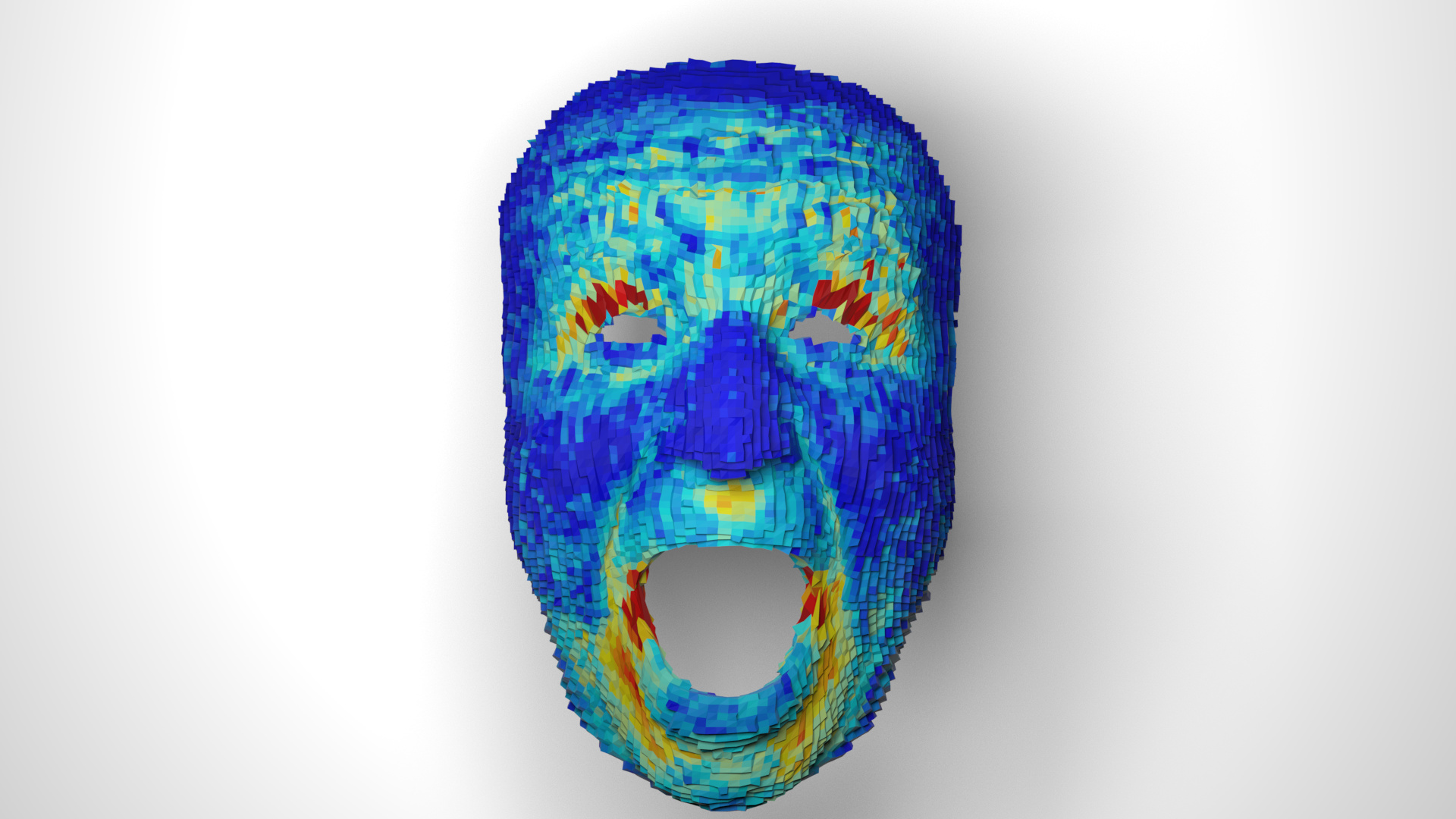}\hspace{-3pt}
\includegraphics[trim=480 0 480 0, clip, width=0.125\linewidth]{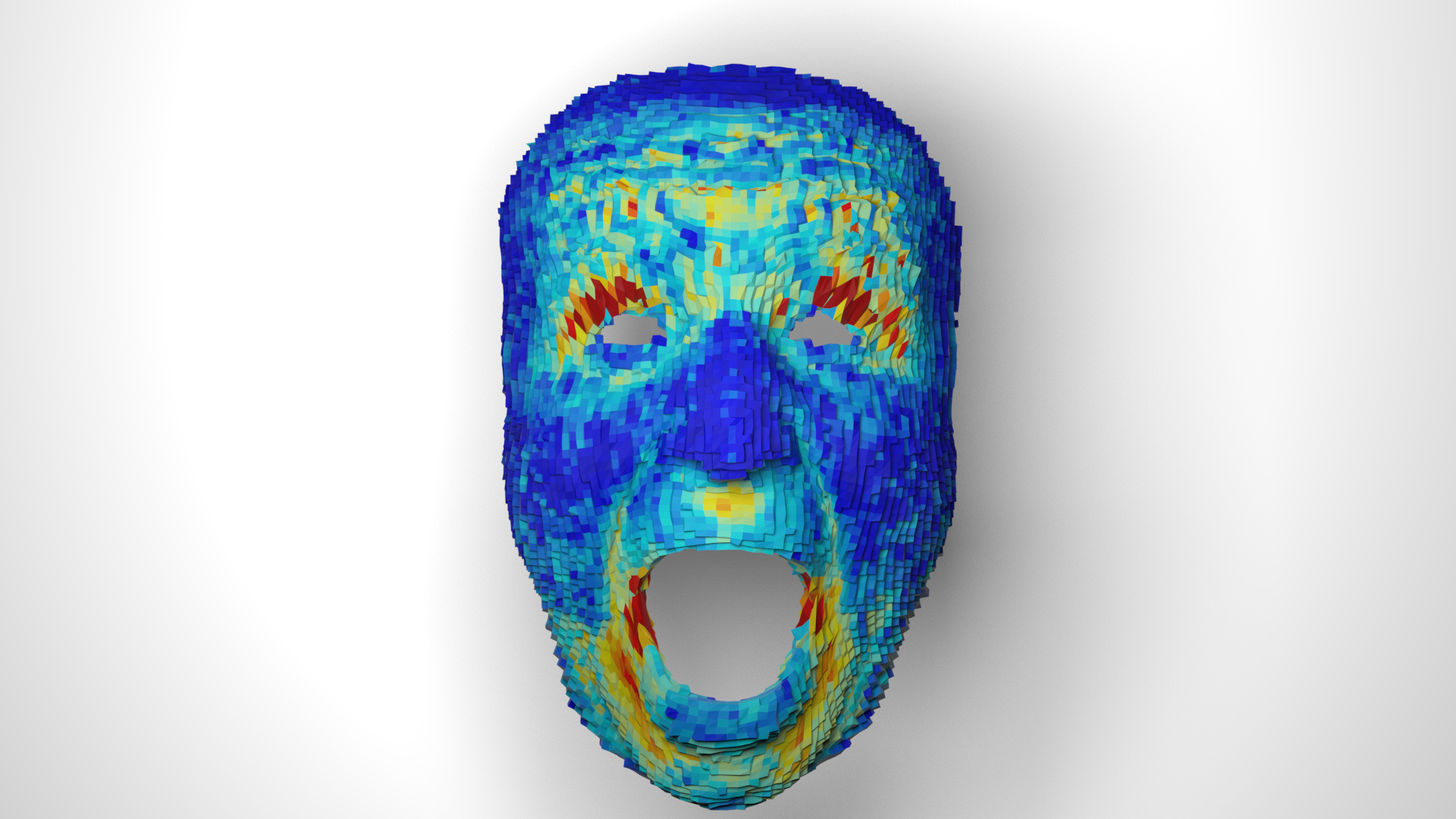}\hspace{-1pt}
\includegraphics[trim=480 0 480 0, clip, width=0.125\linewidth]{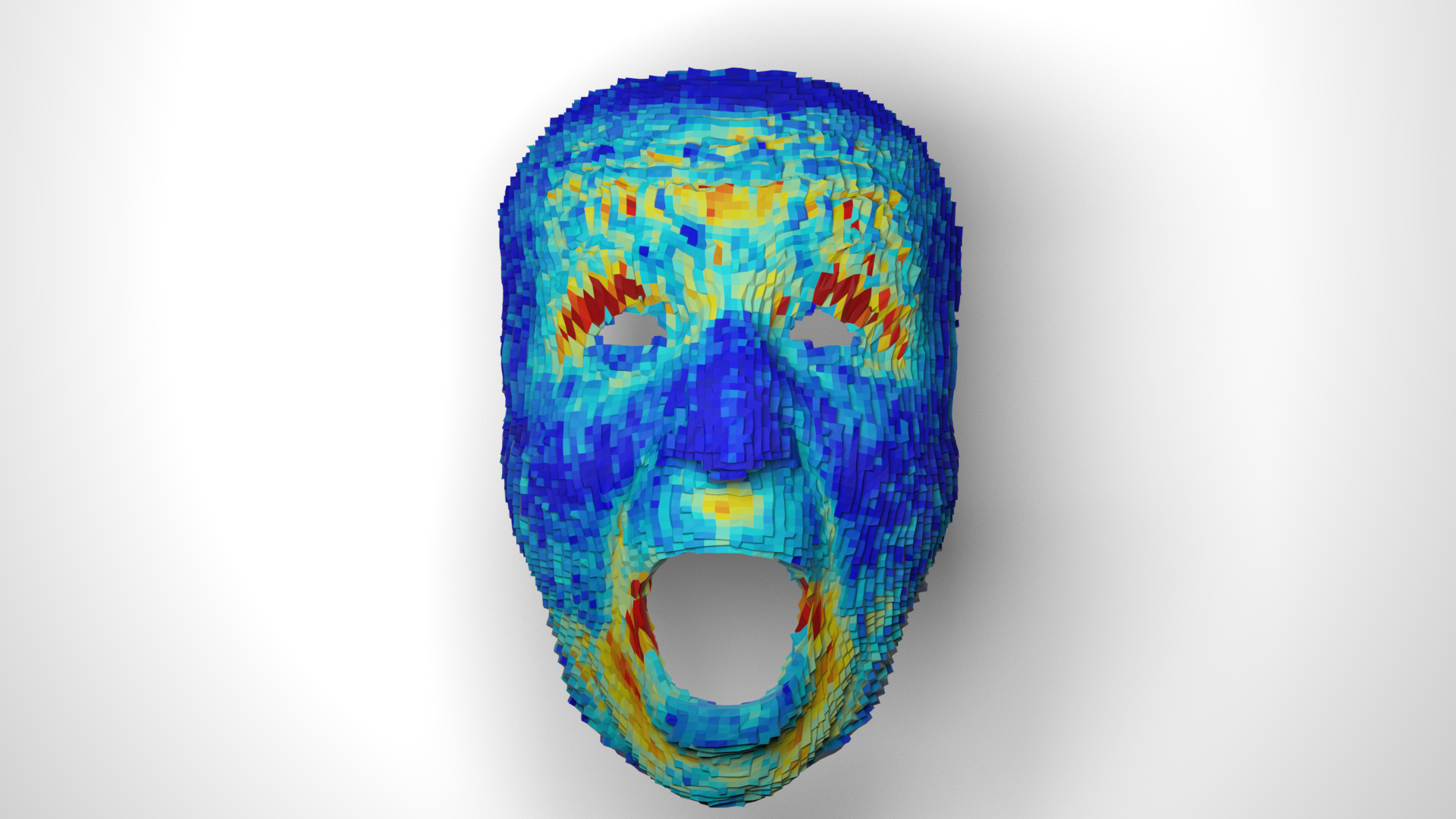}\\
\includegraphics[trim=480 0 480 0, clip, width=0.125\linewidth]{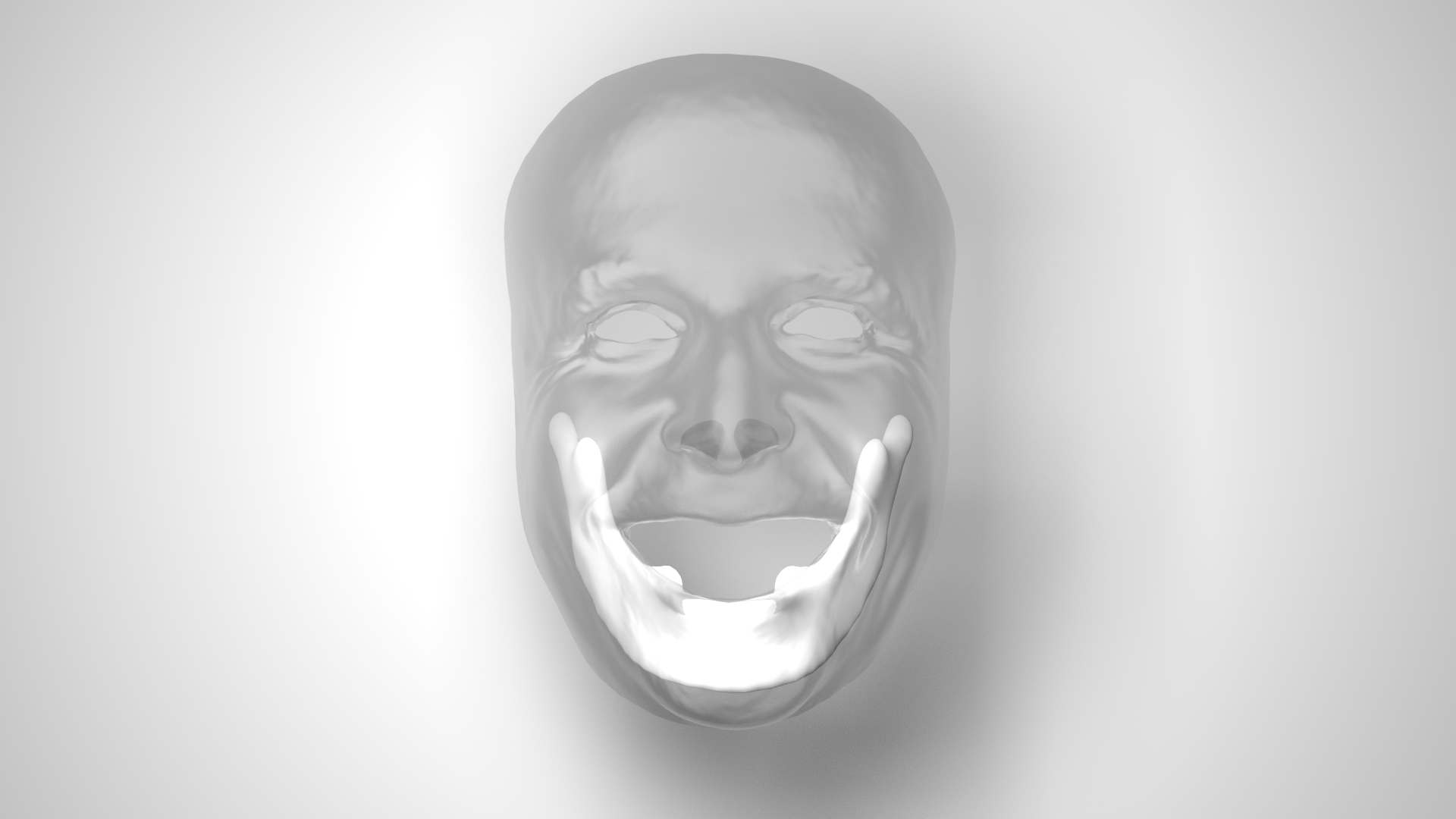}\hspace{-1pt}
\includegraphics[trim=480 0 480 0, clip, width=0.125\linewidth]{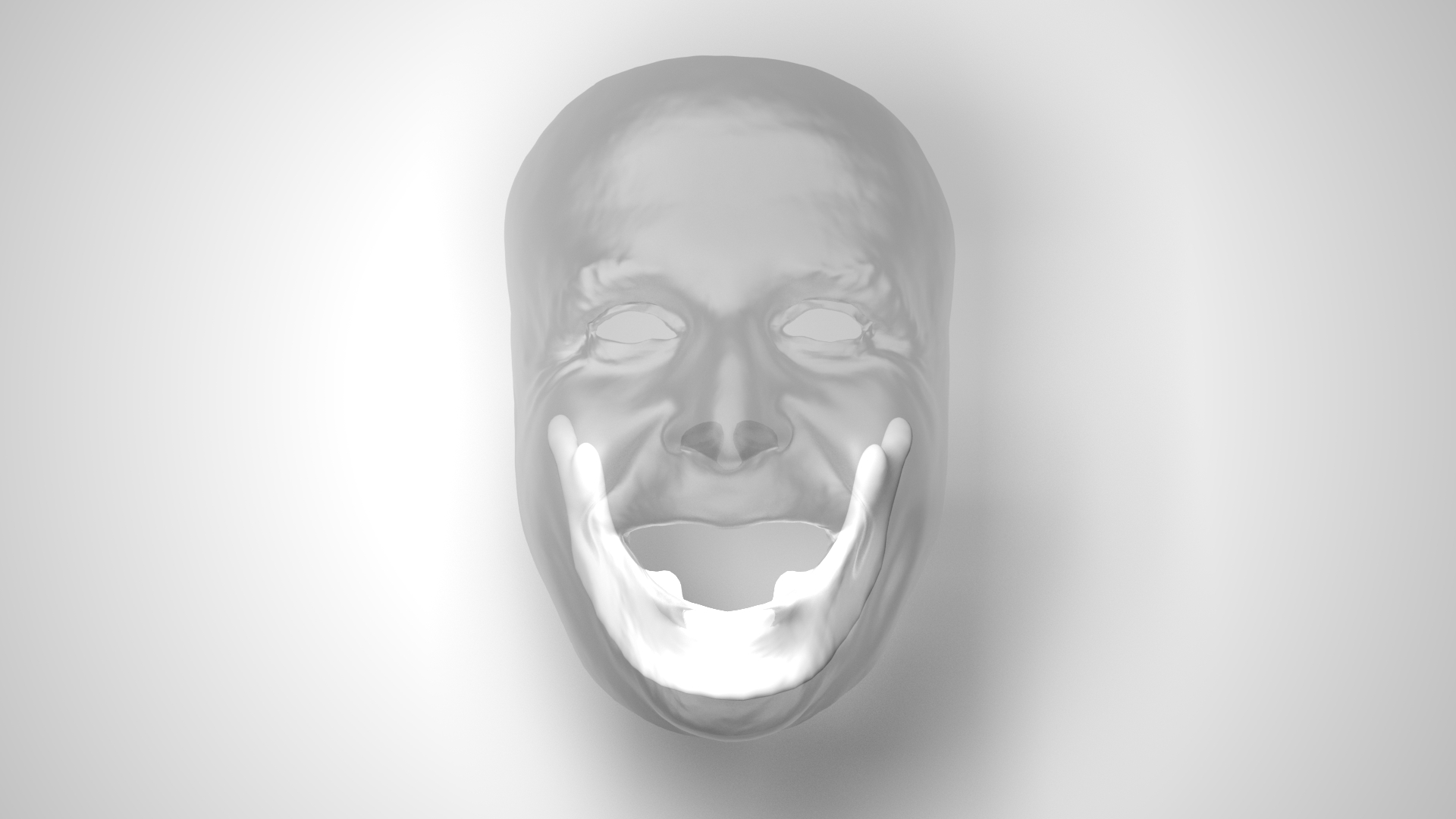}\hspace{-3pt}
\includegraphics[trim=480 0 480 0, clip, width=0.125\linewidth]{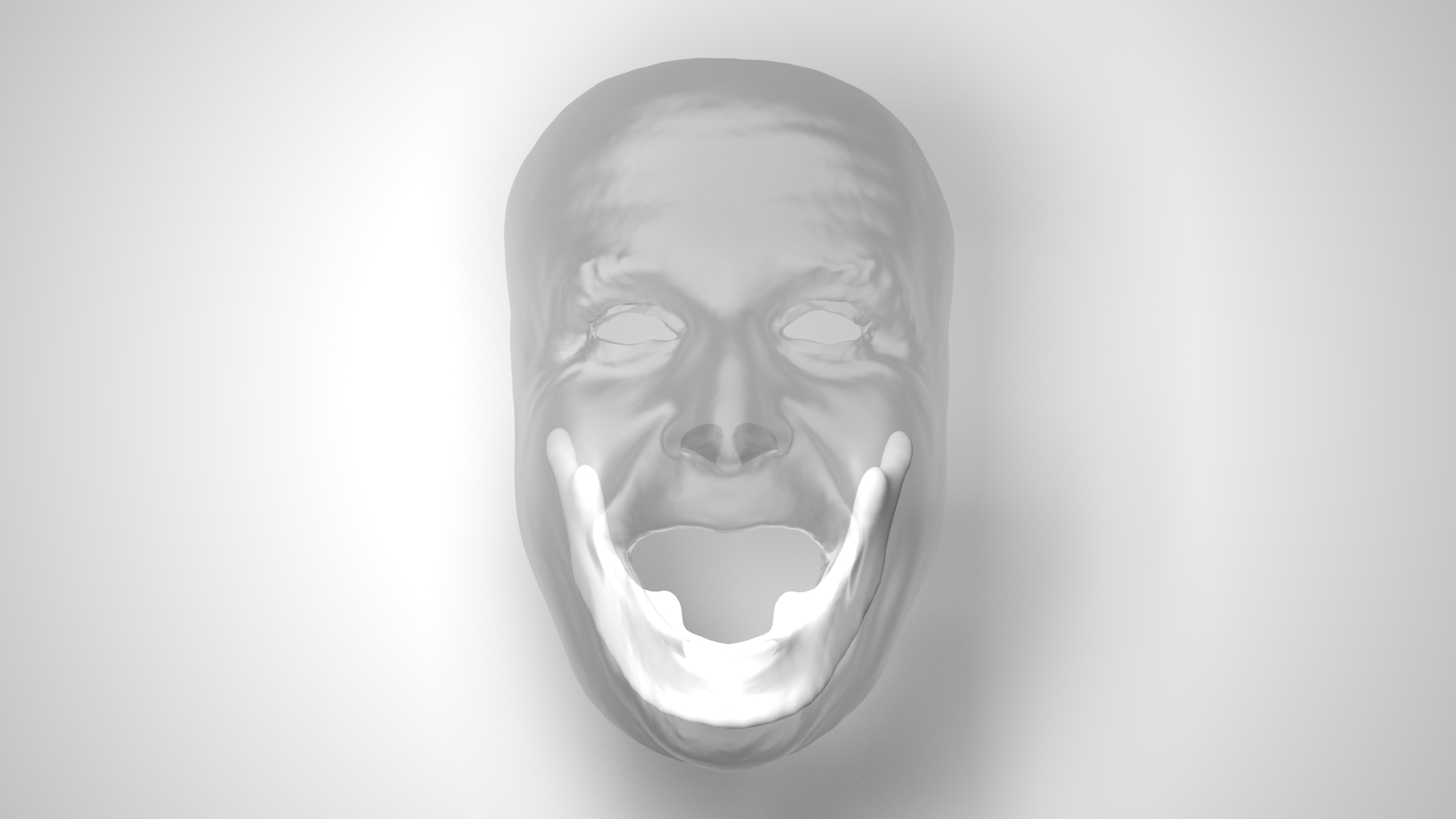}\hspace{-3pt}
\includegraphics[trim=480 0 480 0, clip, width=0.125\linewidth]{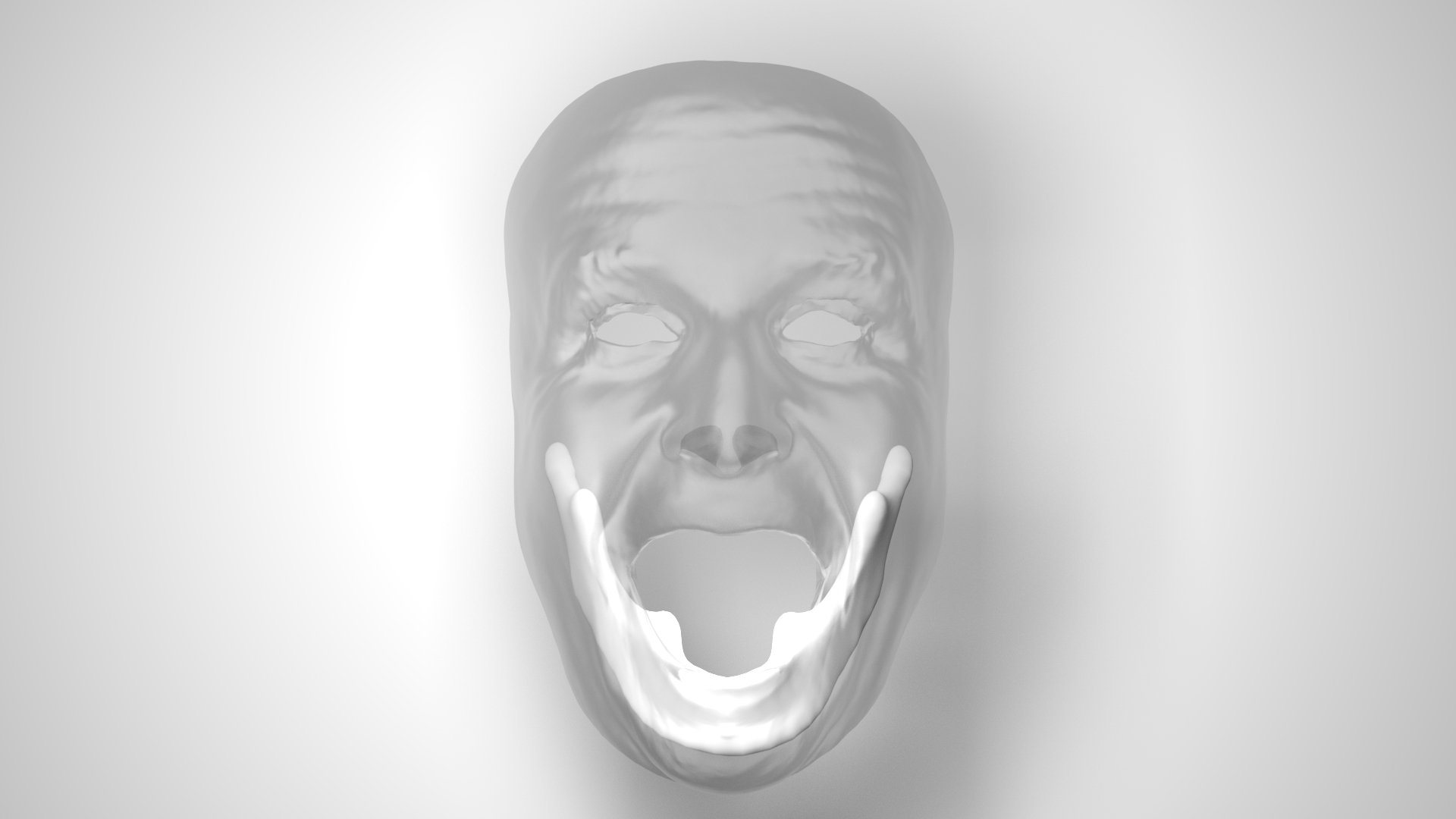}\hspace{-3pt}
\includegraphics[trim=480 0 480 0, clip, width=0.125\linewidth]{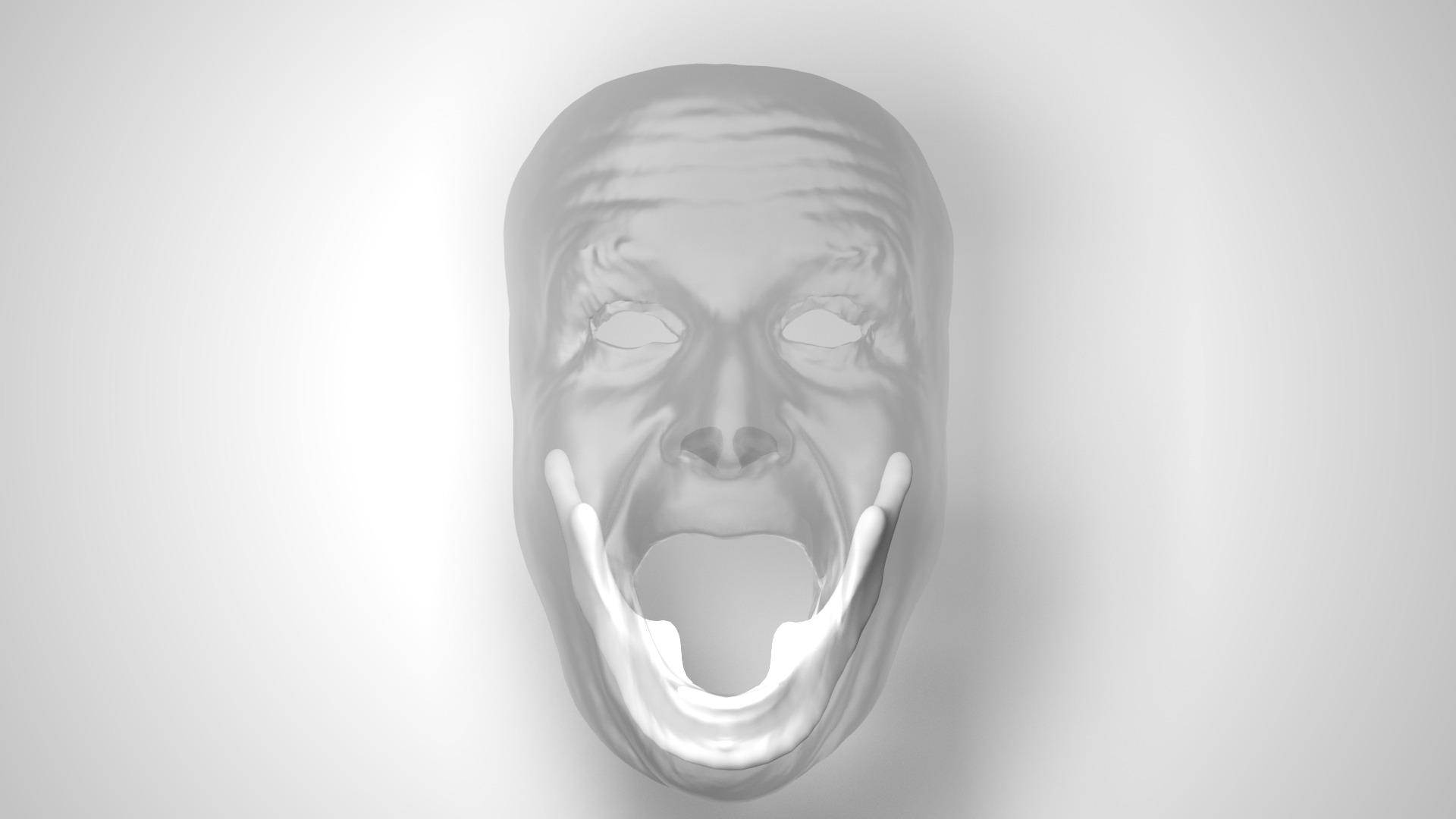}\hspace{-3pt}
\includegraphics[trim=480 0 480 0, clip, width=0.125\linewidth]{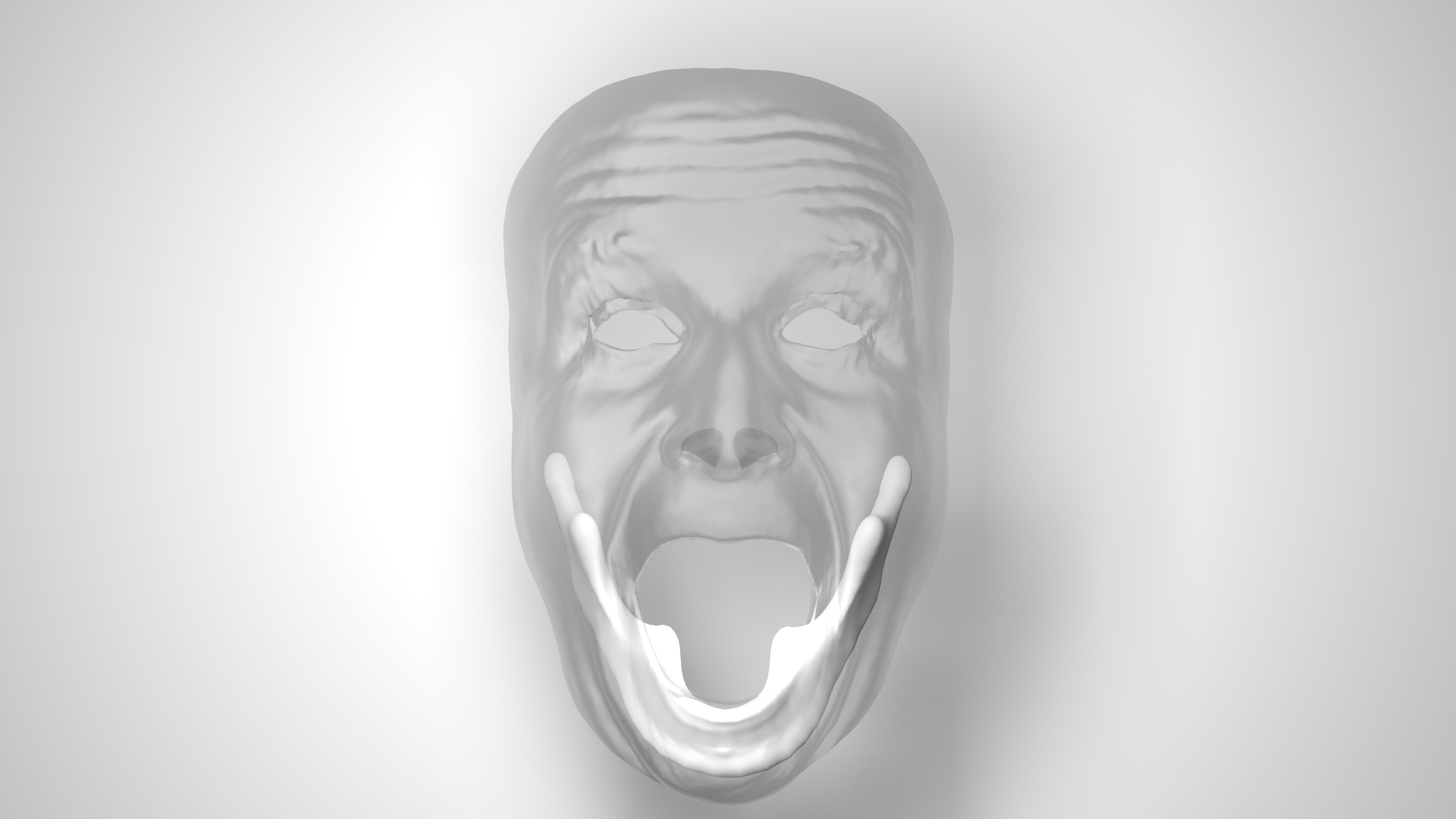}\hspace{-3pt}
\includegraphics[trim=480 0 480 0, clip, width=0.125\linewidth]{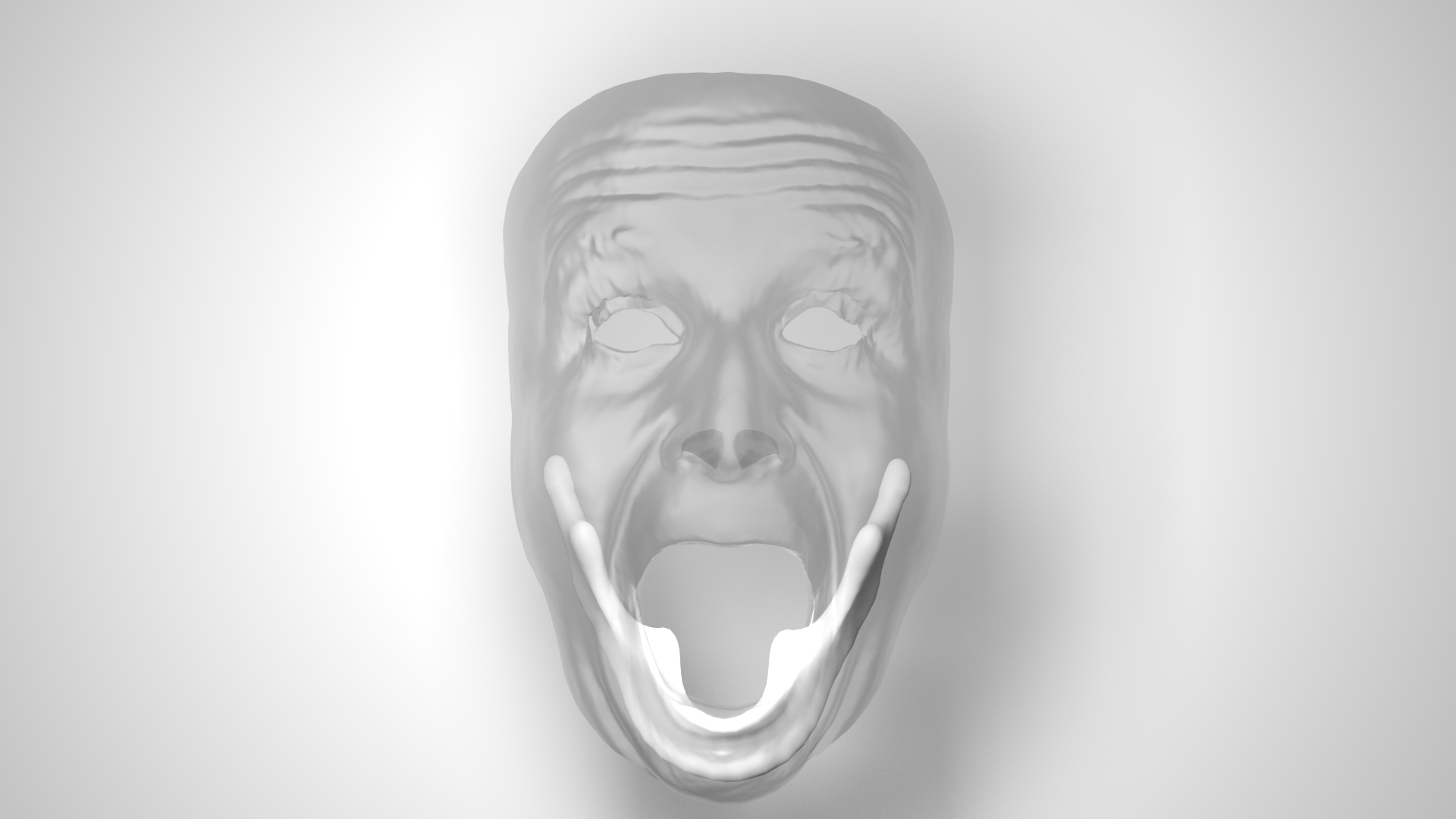}\hspace{-1pt}
\includegraphics[trim=480 0 480 0, clip, width=0.125\linewidth]{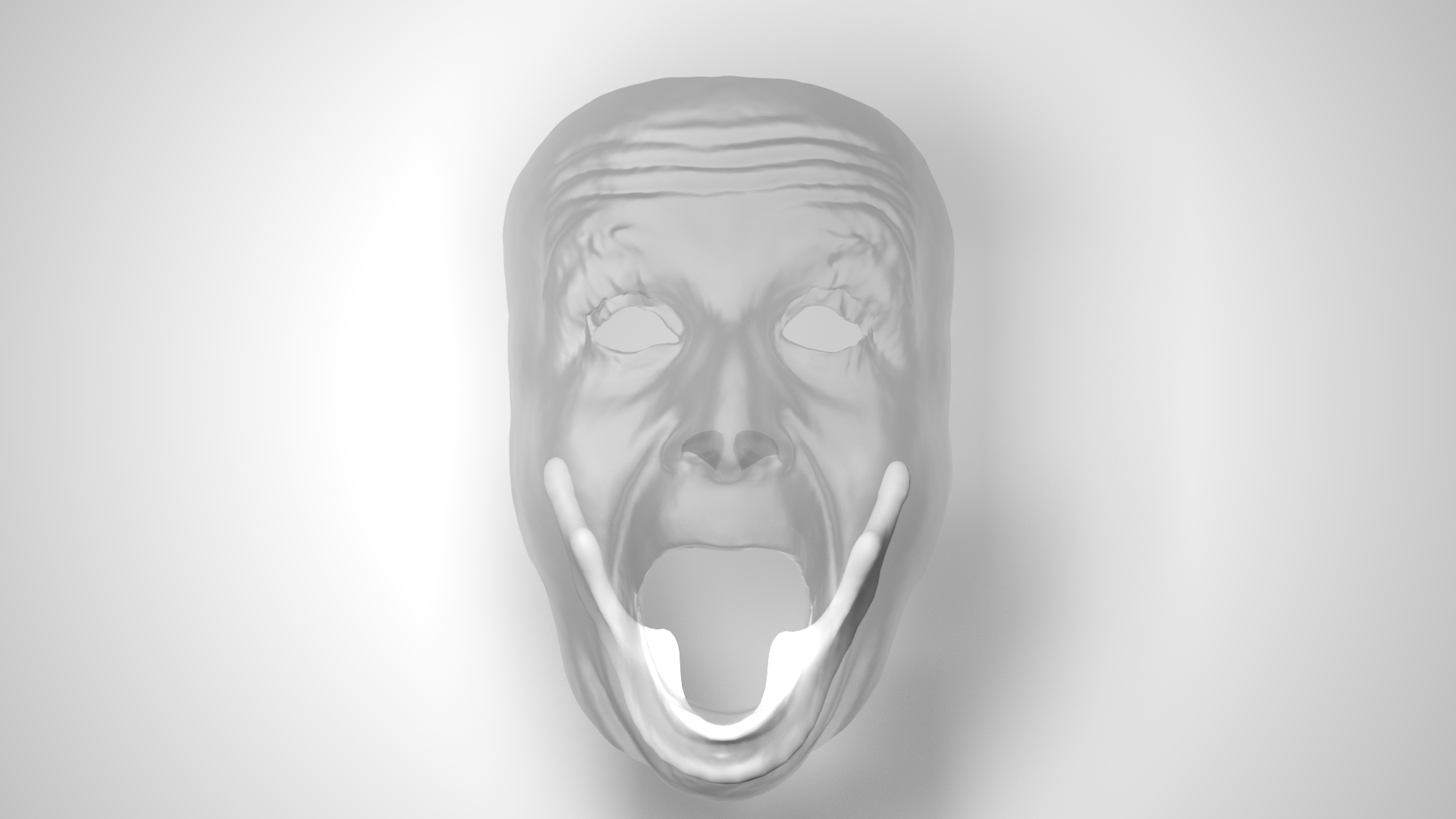}
\caption{Latent space interpolation between two selected expressions.}
\label{fig:latent_intp}
\end{figure*}
\subsection{Results}
\paragraph{Target shape matching}
\Fig{train} shows the training results for the three datasets. In all examples, our model can reliably reproduce the target shapes with the optimized parameters, as indicated by the low reconstruction error. The actuation signals are color-coded on the hexahedral mesh on the right, where blue and red indicate a low and high value, respectively. Note that the network contains only $0.3\text{M}$ parameters but has the representational power to reliably reproduce fine details such as wrinkles.
\review{In contrast, the number of parameters of explicit models depends on the simulation mesh (discretization) and is typically two orders of magnitude larger.
For example, Srinivasan et al. \shortcite{srinivasan2021learning} uses about 43M parameters in the first three layers of the decoder (not including the final and sparse layer with custom connectivity).
} 

For the face we optimize both actuation signals and mandible kinematics. 
Thanks to the jaw-generative network we are able to directly compute a jaw position that is compatible with the actuation (\Fig{train_jaw}). This was not possible in previous work~\cite{srinivasan2021learning}, where the boundary conditions associated with skeletal attachments are presumed known at all time instances. As expected, the differentiable connection between the jaw and simulator reduces the error \review{(see the \hyperref[sec:ablation]{Ablation Study} below)} and even successfully resolves collision artifacts (highlighted in the red box) of the given initialization~\cite{Zoss2019}.

\paragraph{New poses}
With our trained model we can fit new expressions by keeping the weights of $\N_{\A}$ and $\N_{\mathbf{B}}$ fixed 
and optimizing for the latent code that yields the best fit, as in~\cite{srinivasan2021learning}.
To do so, we first use our encoder to get an initialized latent code, 
then optimize it for 10 iterations.  
\Fig{test} illustrates the result for unseen target poses and the associated low errors.

In~\Fig{latent_intp} we interpolate between two selected expressions, \ie from smiling to angrily surprised, indicating smooth transitions despite the non-linearity of the deformation. The actuation and jaw position are also consistently interpolated.

\paragraph{Continuous resolution} 
\begin{figure}[ht!] %
\centering
\includegraphics[trim=480 0 480 0, clip, width=0.25\linewidth]{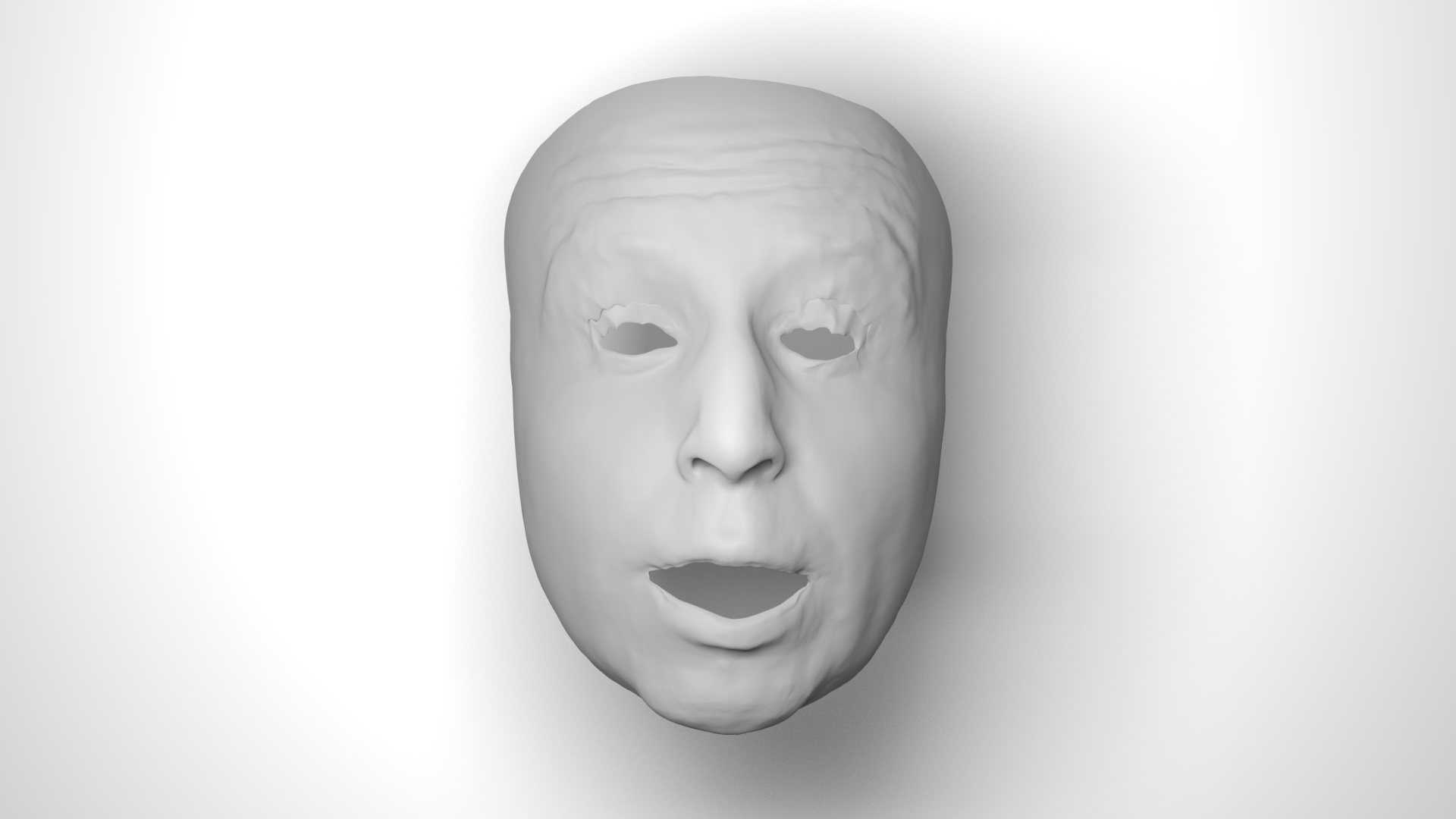}\hspace{-3pt}
\includegraphics[trim=480 0 480 0, clip, width=0.25\linewidth]{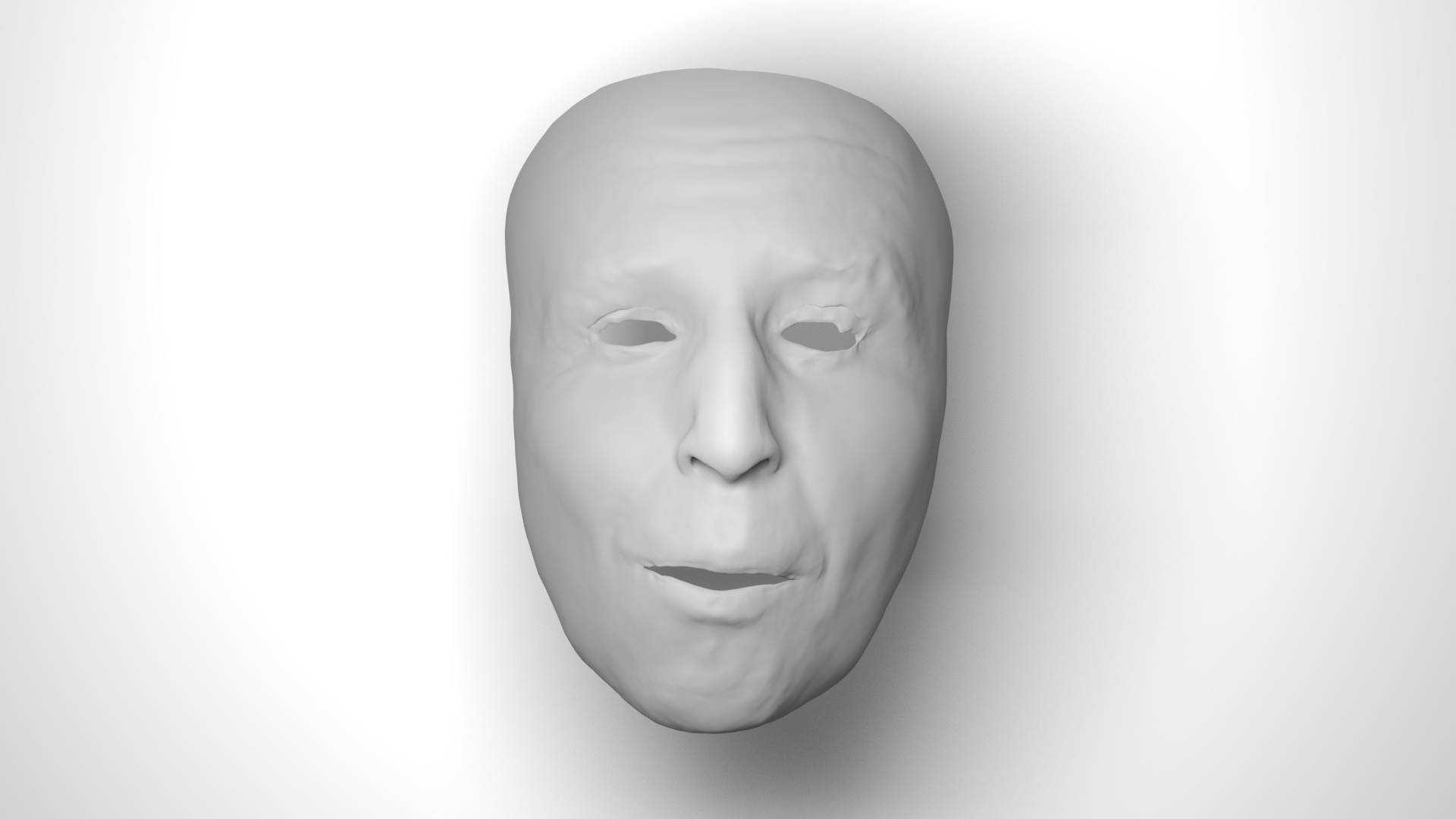}\hspace{-3pt}
\includegraphics[trim=480 0 480 0, clip, width=0.25\linewidth]{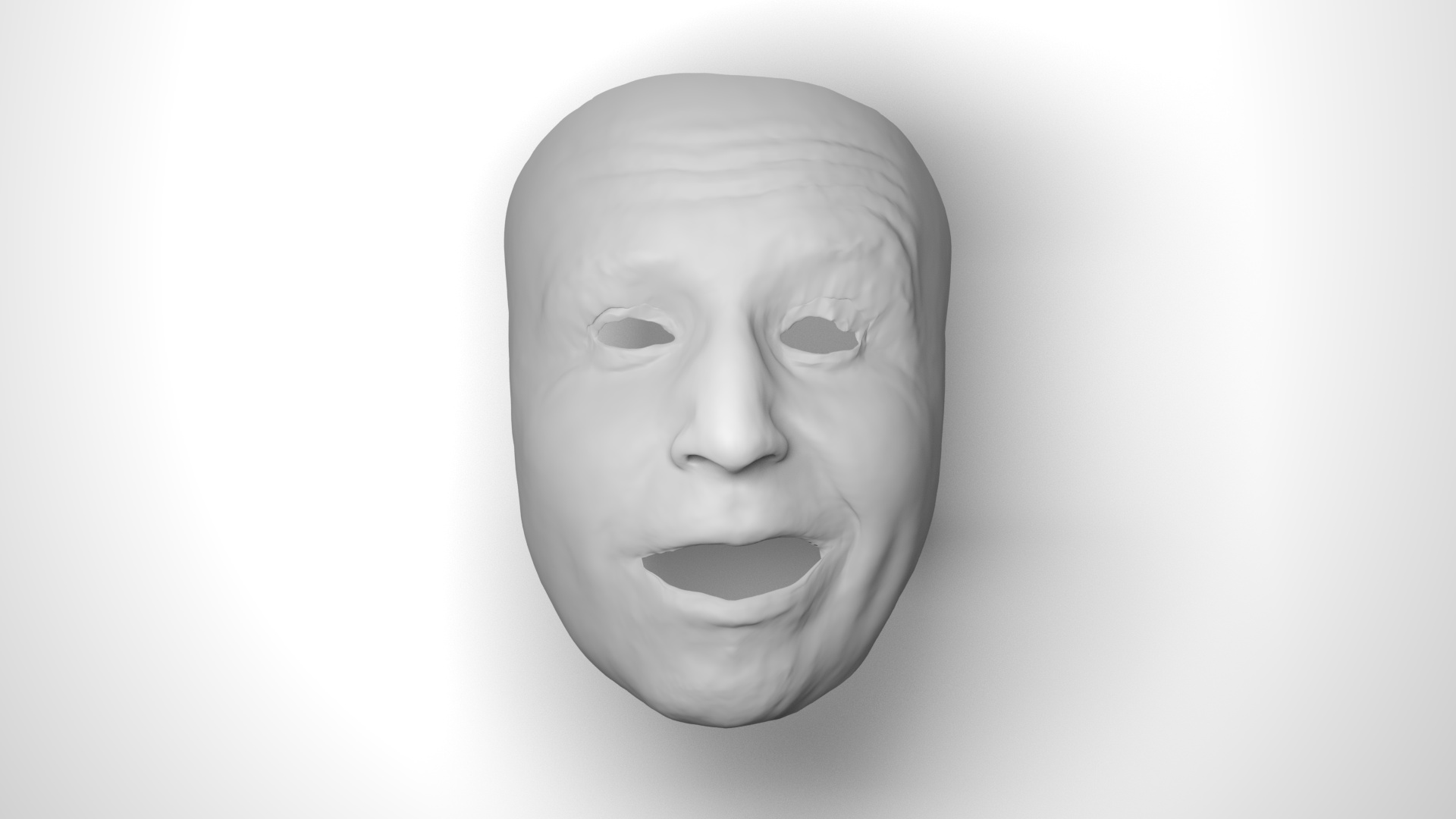}\hspace{-3pt}
\includegraphics[trim=480 0 480 0, clip, width=0.25\linewidth]{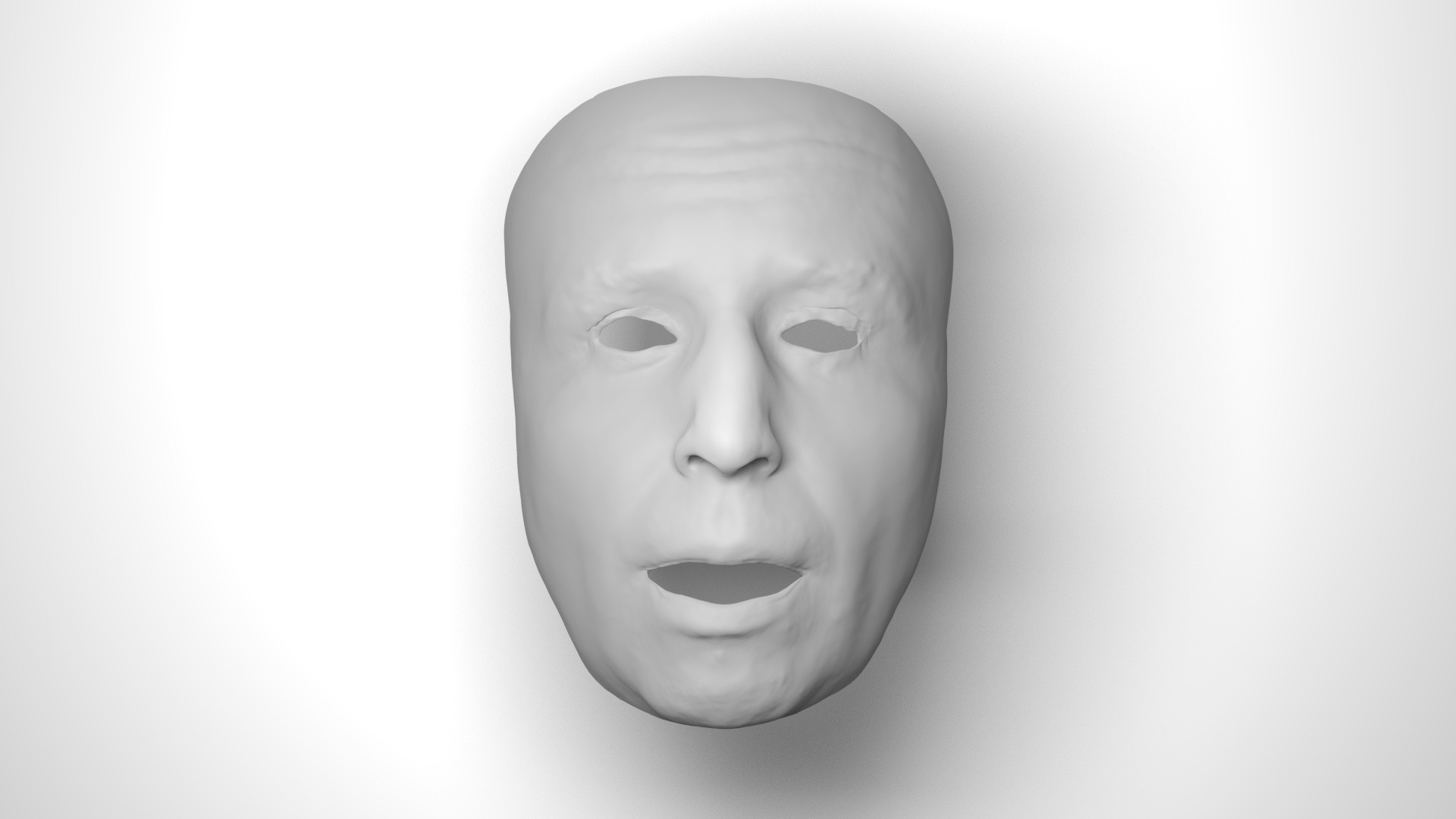}\\
\includegraphics[trim=480 0 480 0, clip, width=0.25\linewidth]{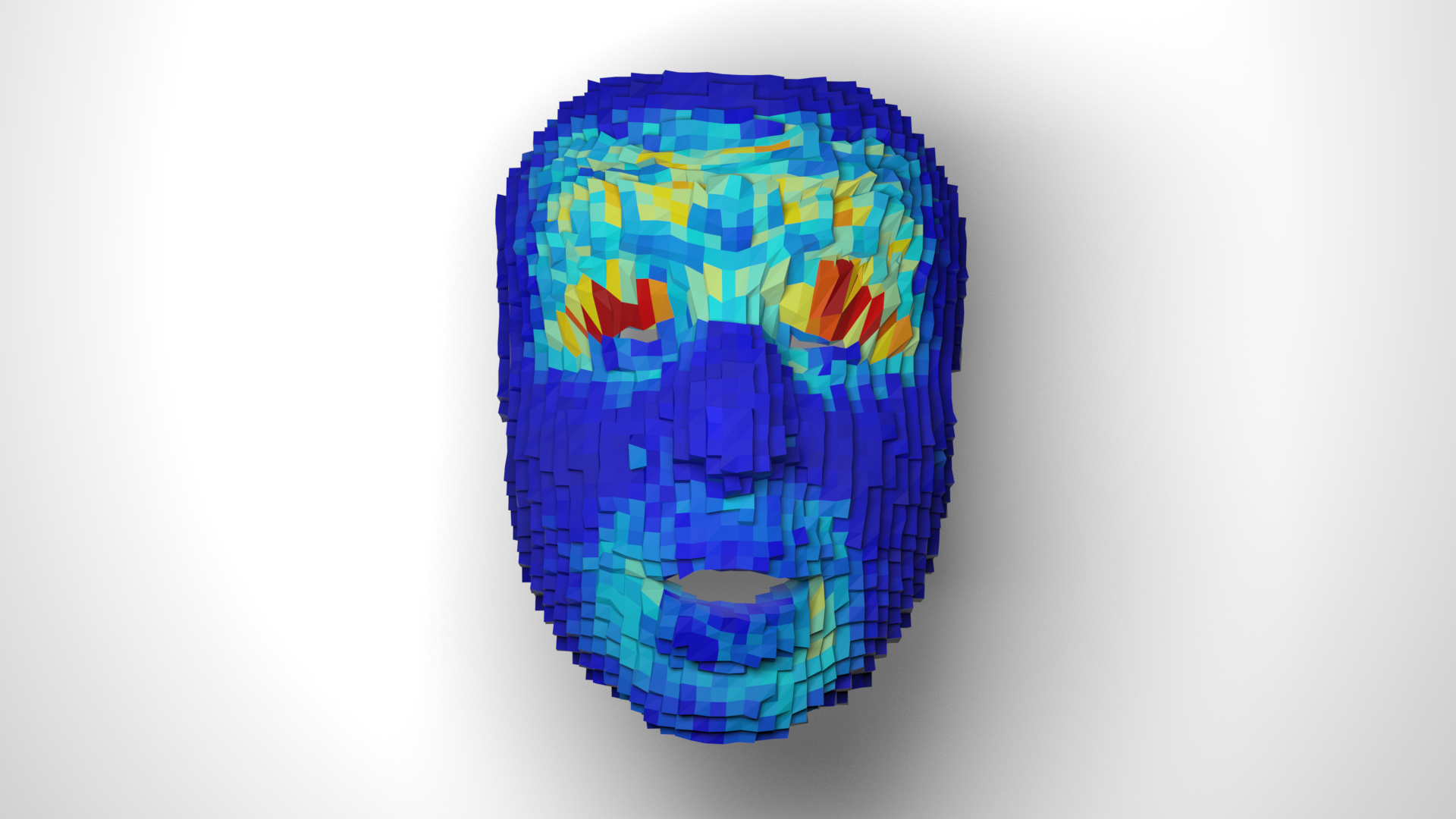}\hspace{-3pt}
\includegraphics[trim=480 0 480 0, clip, width=0.25\linewidth]{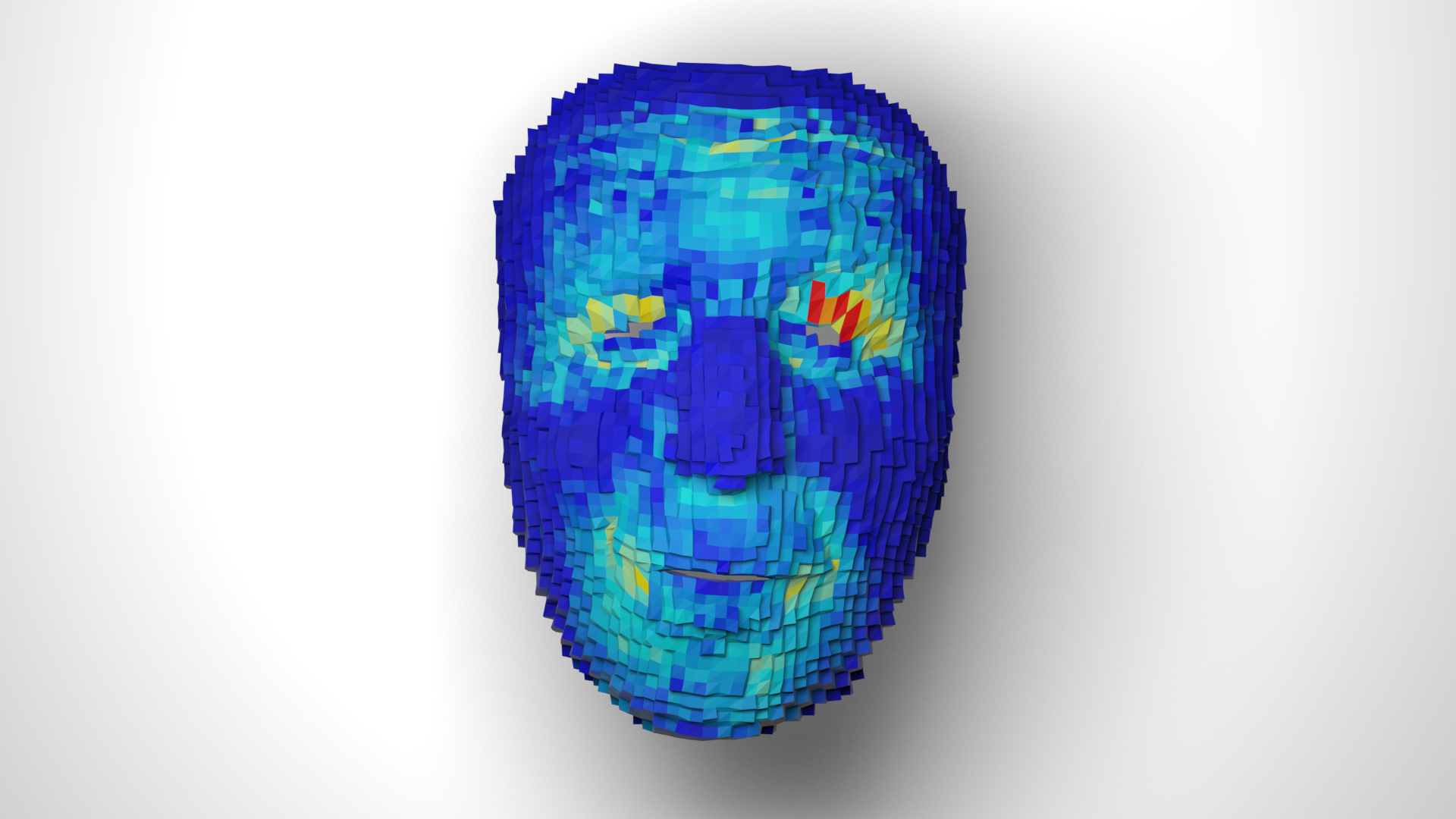}\hspace{-3pt}
\includegraphics[trim=480 0 480 0, clip, width=0.25\linewidth]{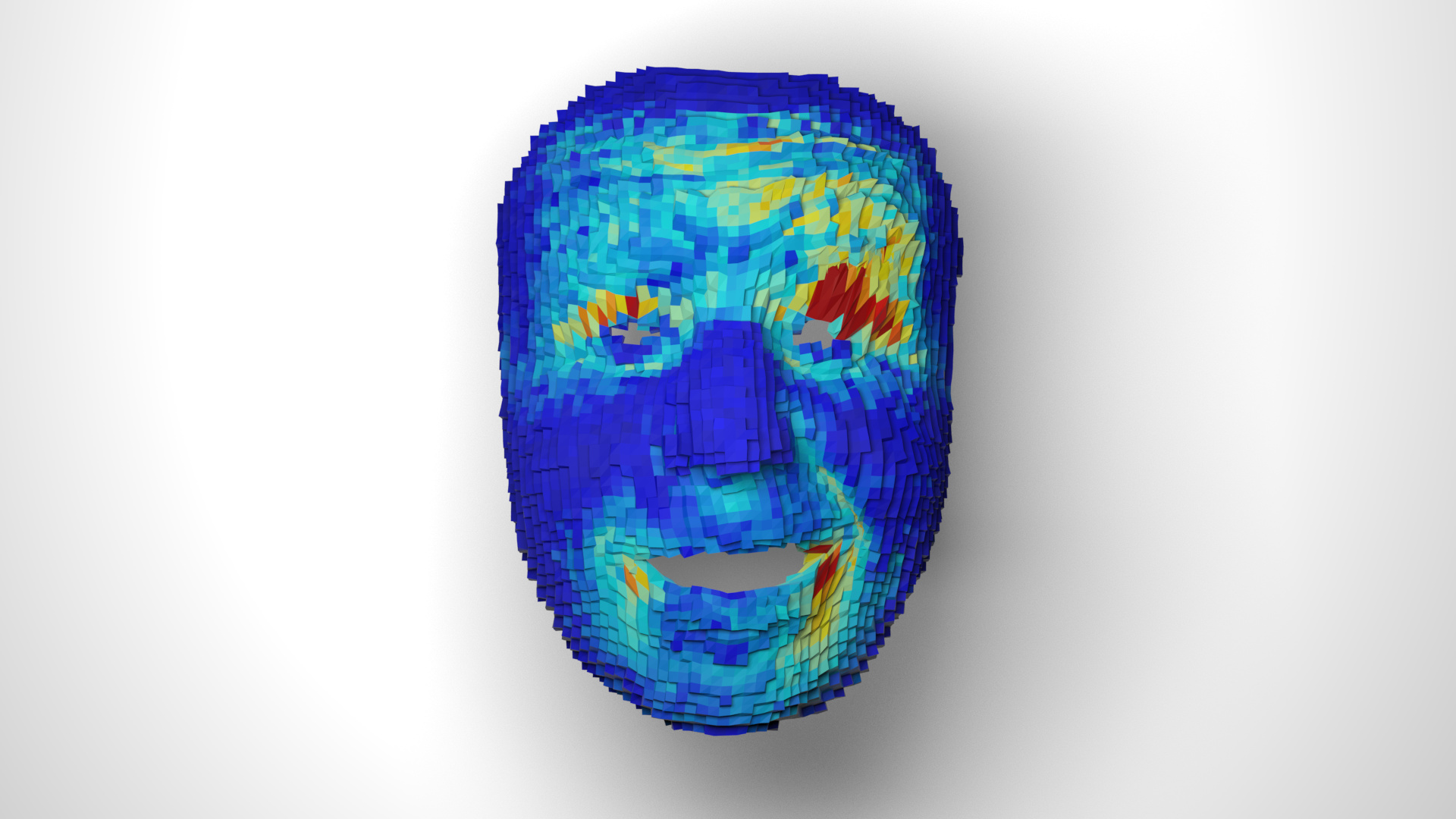}\hspace{-3pt}
\includegraphics[trim=480 0 480 0, clip, width=0.25\linewidth]{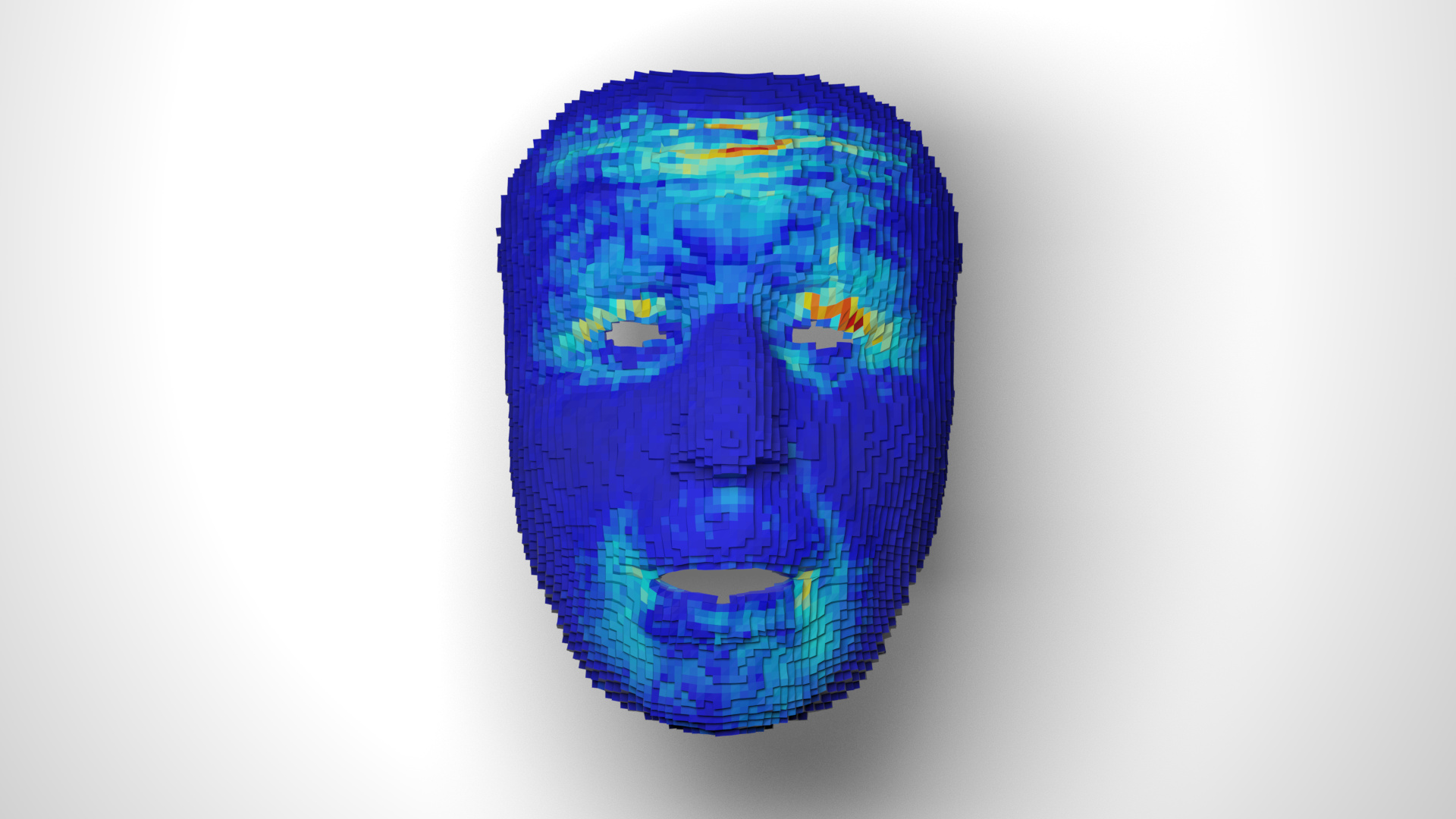}\\
\includegraphics[trim=480 0 480 0, clip, width=0.25\linewidth]{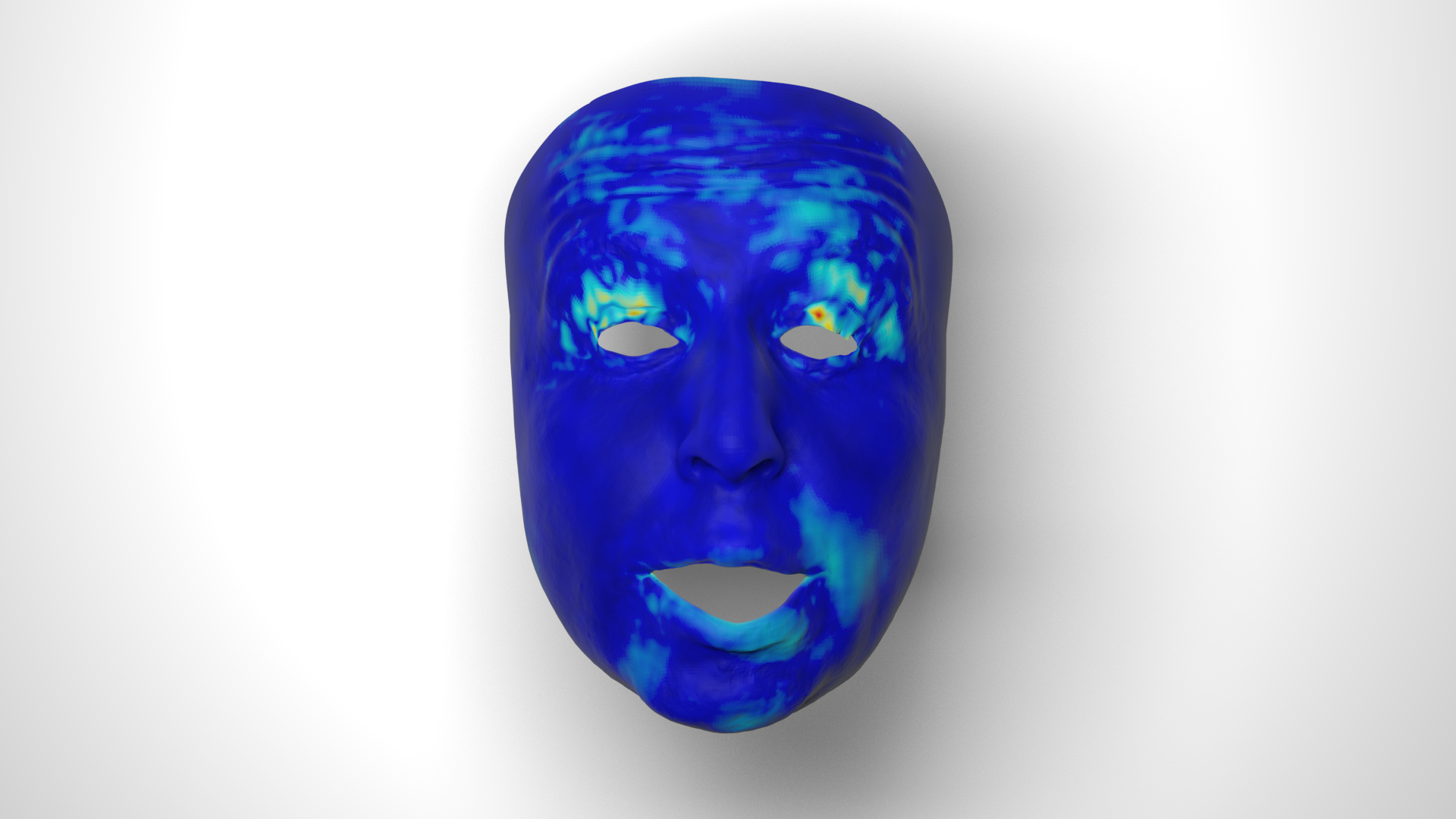}\hspace{-3pt}
\includegraphics[trim=480 0 480 0, clip, width=0.25\linewidth]{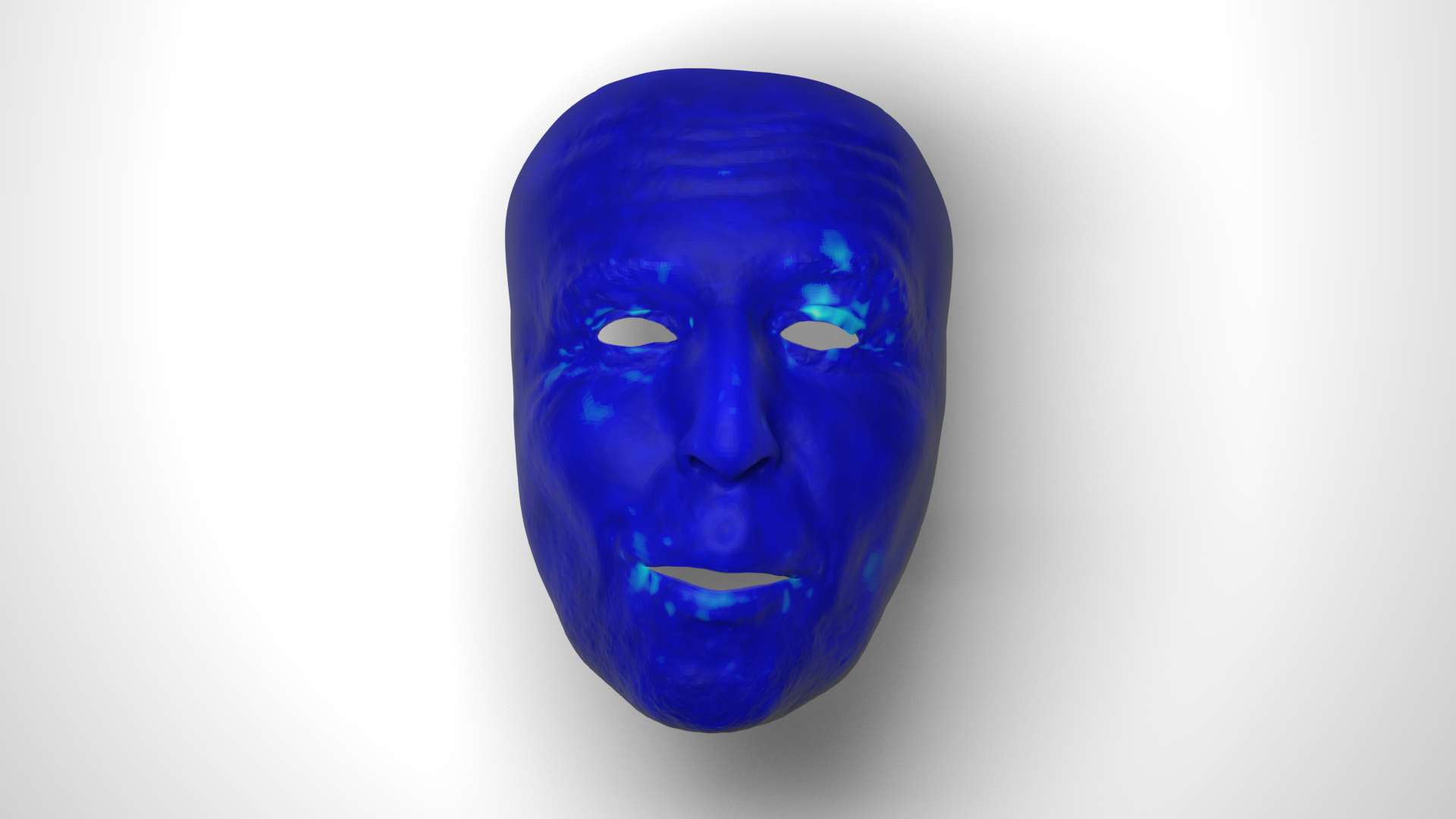}\hspace{-3pt}
\includegraphics[trim=480 0 480 0, clip, width=0.25\linewidth]{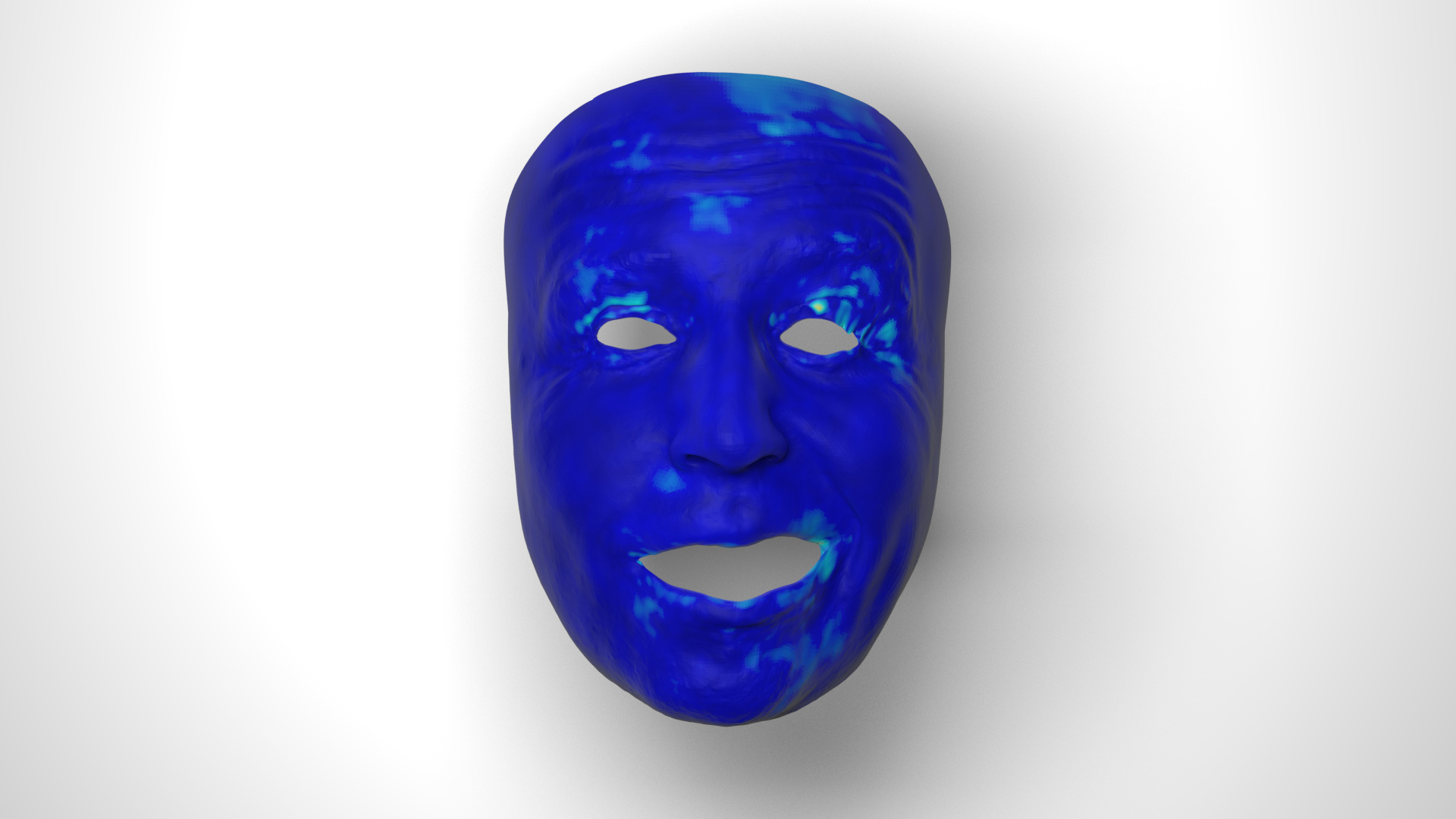}\hspace{-3pt}
\includegraphics[trim=480 0 480 0, clip, width=0.25\linewidth]{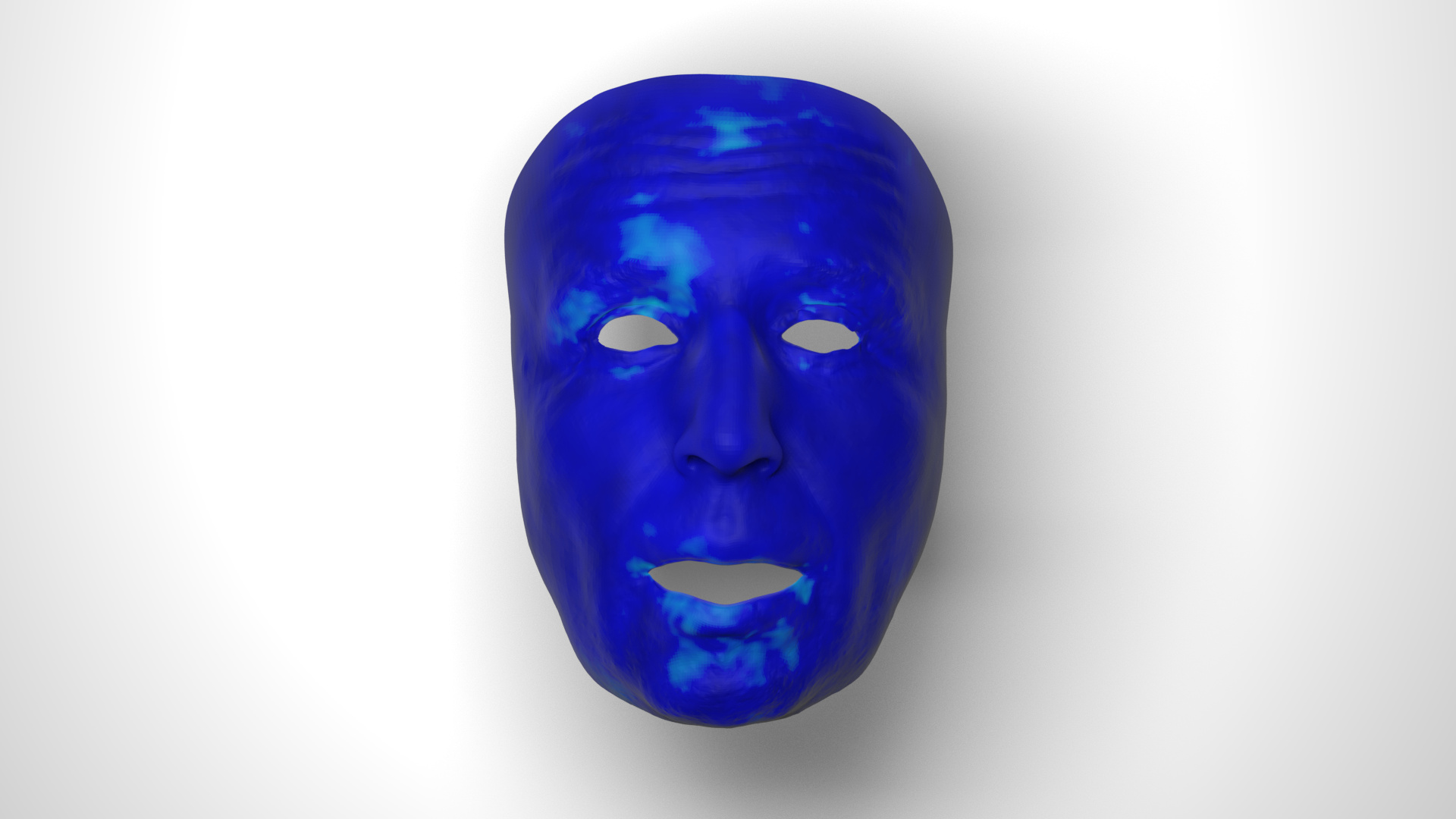}
\caption{Results with four unseen input resolutions used at test time (from left to right: 51K, 117K, 192K, 258K).}
\label{fig:lod}
\end{figure}
As demonstrated in~\Fig{lod}, our method successfully generalizes to continuous resolution input. We show results for four different, unseen input resolutions ranging from 51K to 258K. \review{More results are shown in the supplemental video.} The resolution invariance property paves the way for efficient practical use where the resolution can be chosen at test time depending on the requirements at hand. 

\begin{figure}[h!] %
\renewcommand{\arraystretch}{0}
\centering
\begin{tabular}{
P{0.25\linewidth}@{\hspace{0pt}}
P{0.25\linewidth}@{\hspace{0pt}}
P{0.25\linewidth}@{\hspace{0pt}}
P{0.25\linewidth}@{\hspace{0pt}}
}
\includegraphics[trim=480 0 480 0, clip, width=1\linewidth]{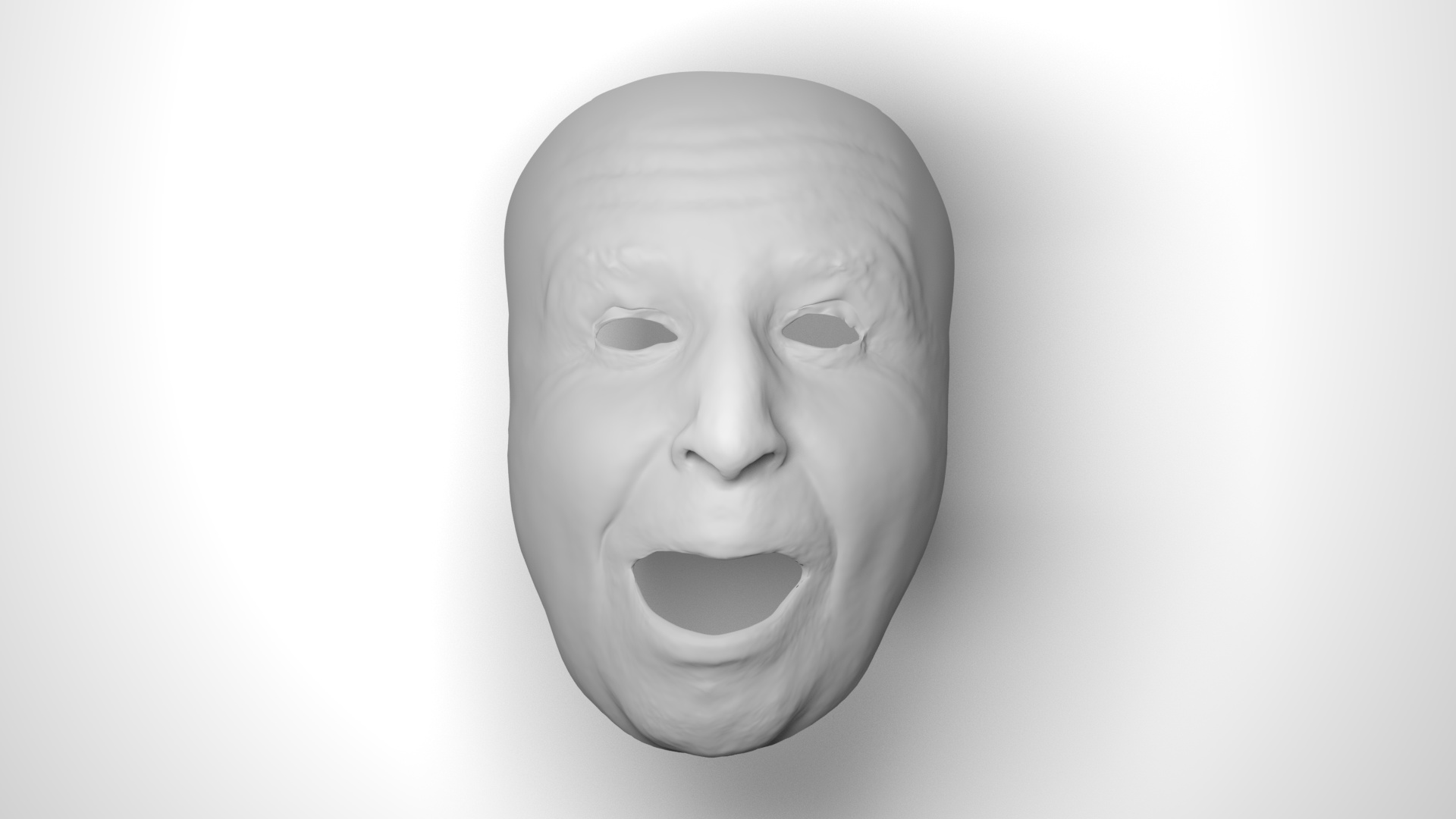} &
\includegraphics[trim=480 0 480 0, clip, width=1\linewidth]{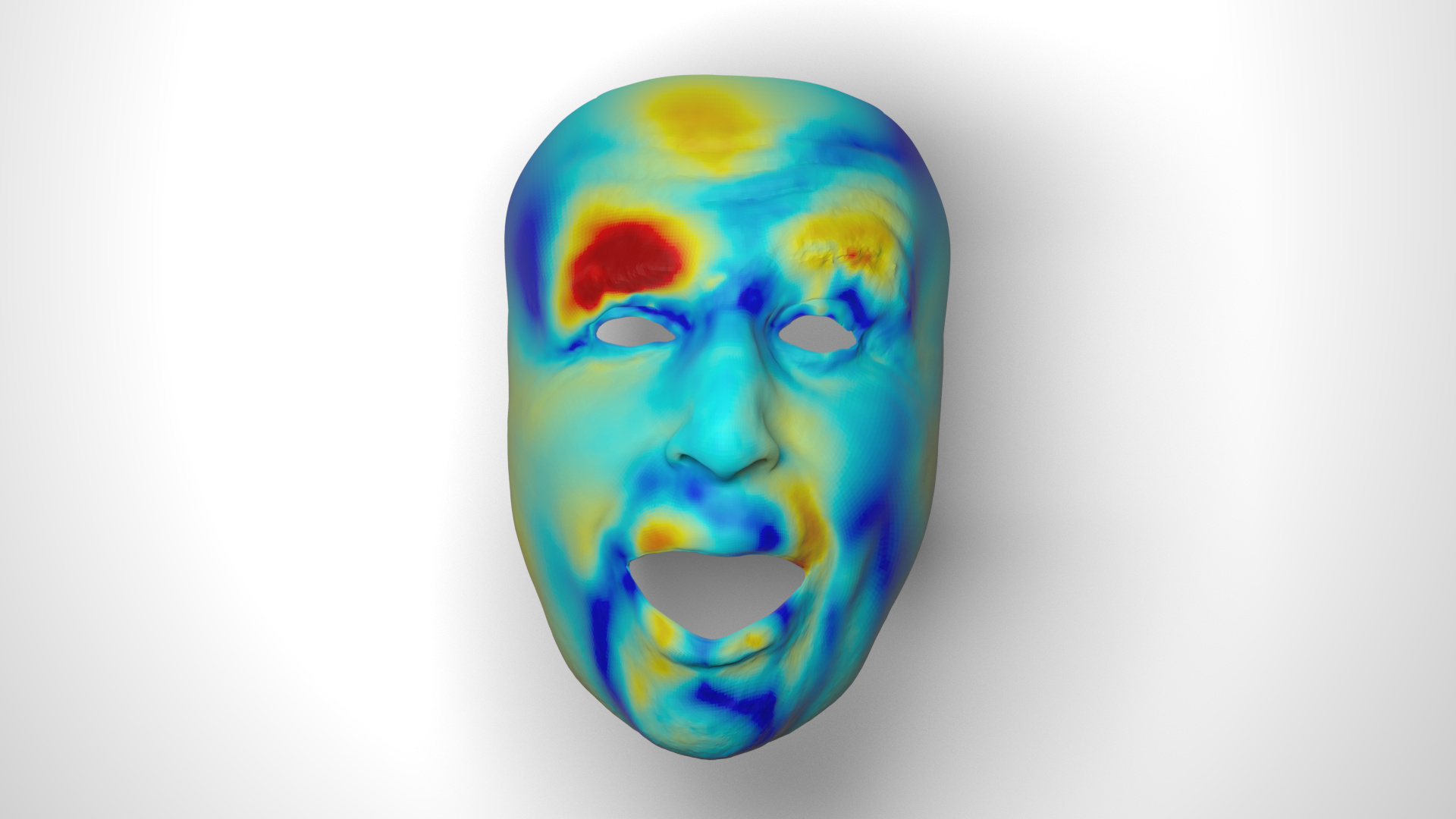} &
\includegraphics[trim=480 0 480 0, clip, width=1\linewidth]{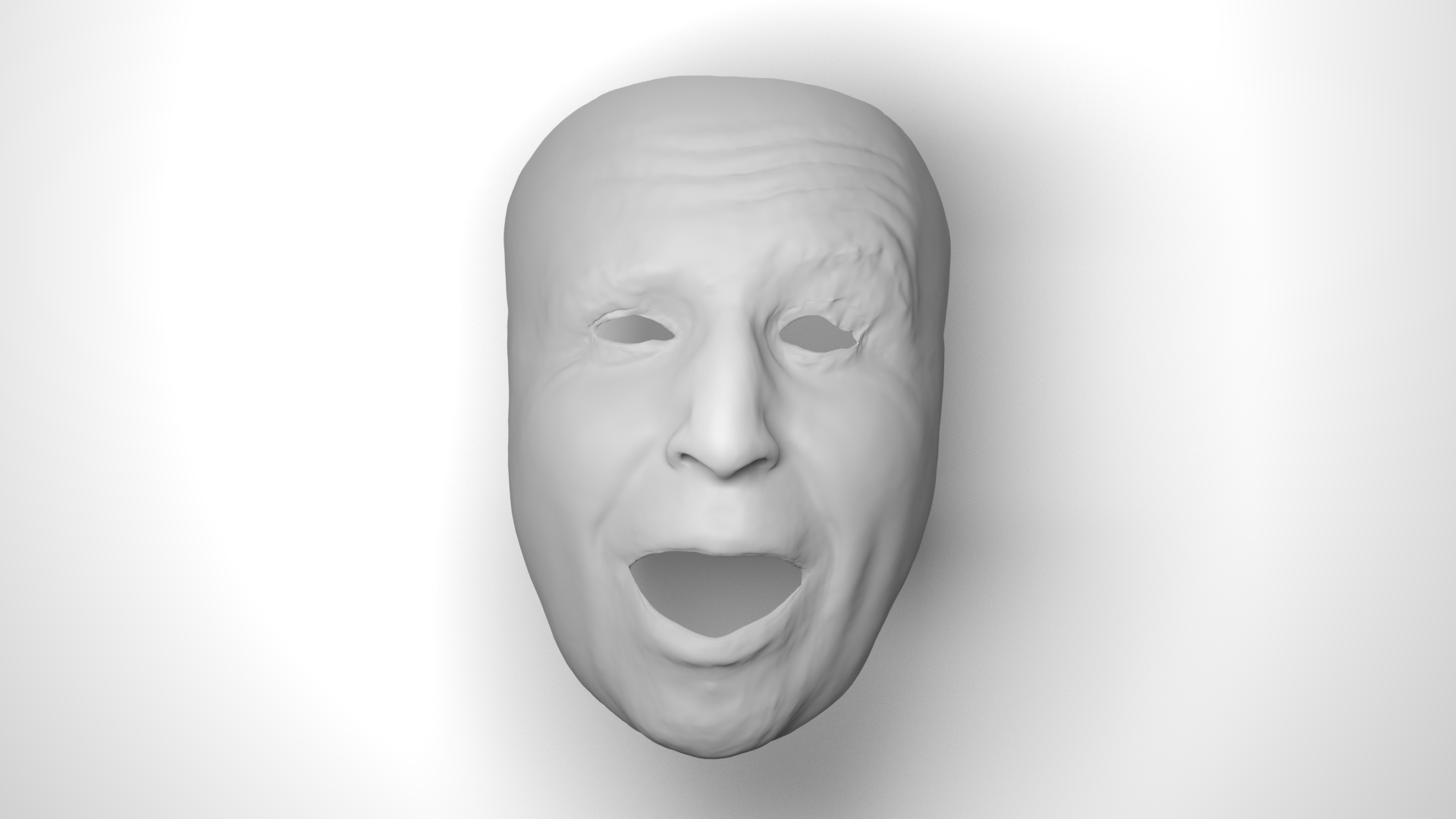} &
\includegraphics[trim=480 0 480 0, clip, width=1\linewidth]{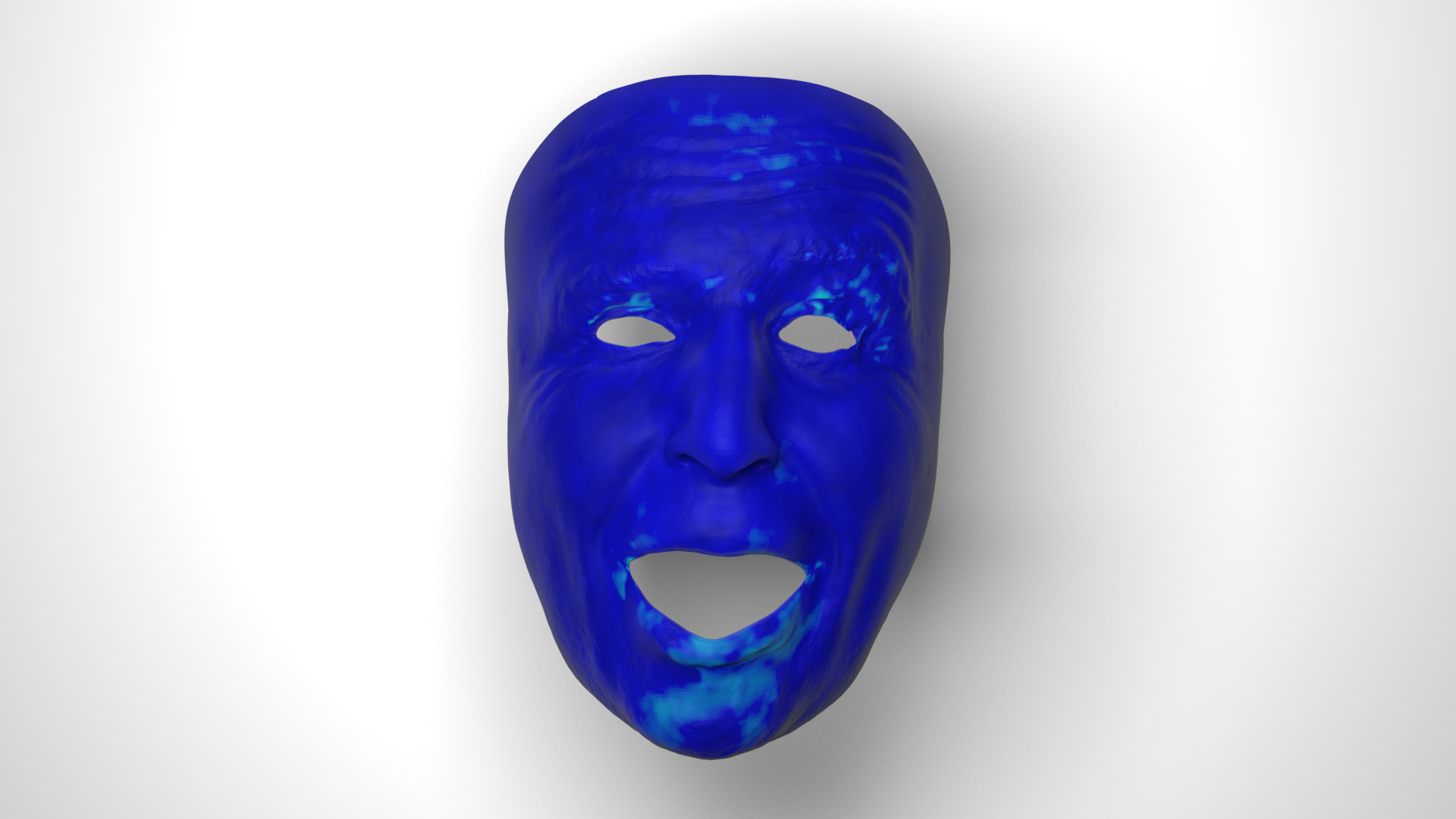}
\end{tabular}
\caption{Comparison of blendshapes (left) and our method (right).}
\label{fig:blendshape}
\end{figure}

\paragraph{Comparison with Blendshapes}
\Fig{blendshape} compares our method with blendshapes. The error maps show clearly the superiority of our method, which can be accounted to the non-linearity of our model. Visually, this property is important for reproducing geometric details and wrinkles, which in this example are conspicuous on the cheek. 

\paragraph{\review{Ablation Study}}
\label{sec:ablation}
We first show the necessity of our normal constraint in the loss function for the face example.
For this, we train a network without the normal constraint (Baseline-NNorm).
\Fig{ablation_norm} shows the difference between Baseline-NNorm and ours. 
The inclusion of the normal constraint positively affects 
the fidelity of the resulting wrinkles and facial details.

\begin{figure}[h!] %
    \centering
    \begin{tabular}{
    P{0.251\linewidth}@{\hspace{0pt}}
    P{0.251\linewidth}@{\hspace{0pt}}
    P{0.251\linewidth}@{\hspace{0pt}}
    P{0.251\linewidth}@{\hspace{0pt}}
    }
    \includegraphics[trim=480 0 480 0, clip, width=1\linewidth]{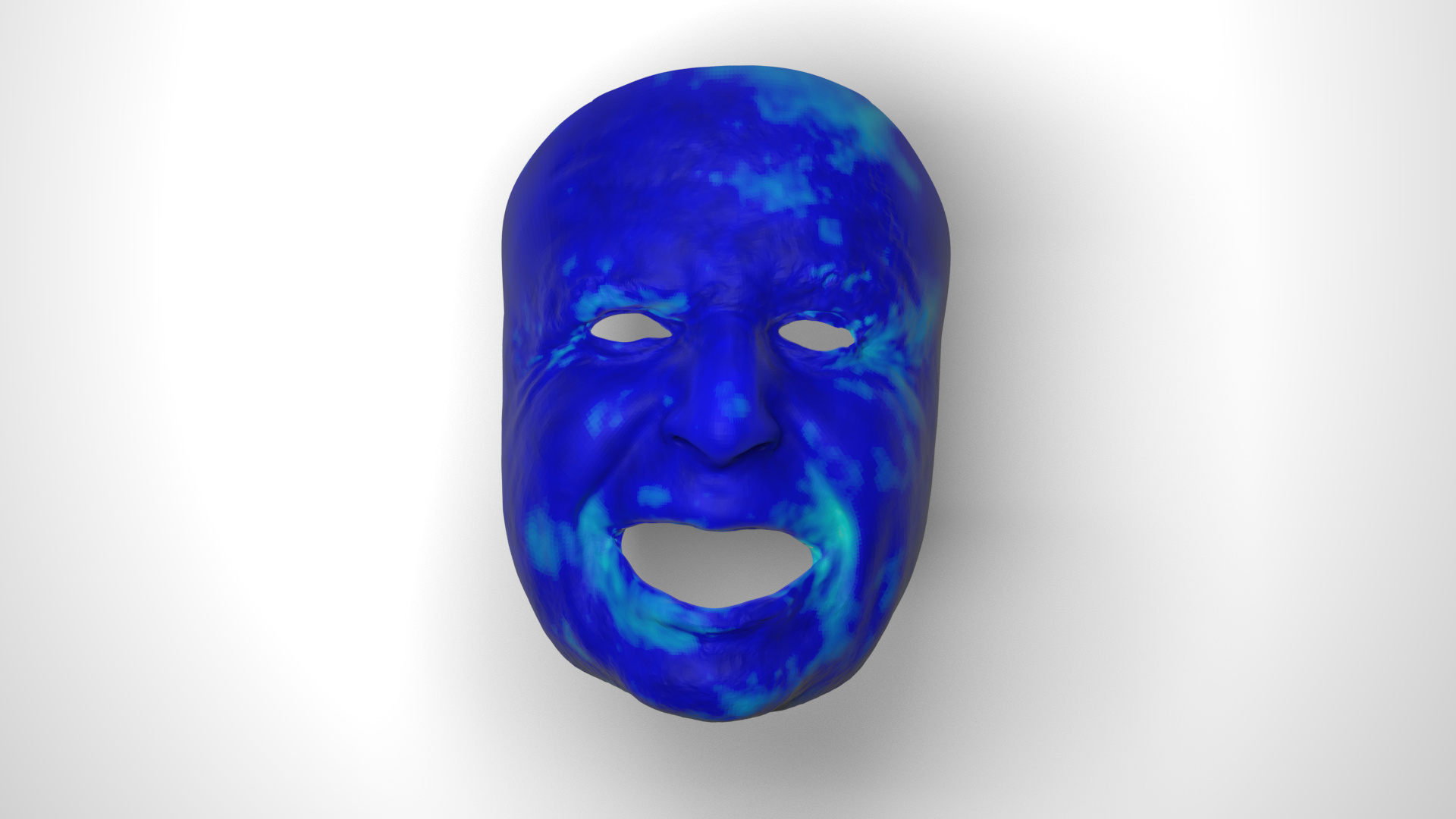} &
    \includegraphics[trim=480 0 480 0, clip, width=1\linewidth]{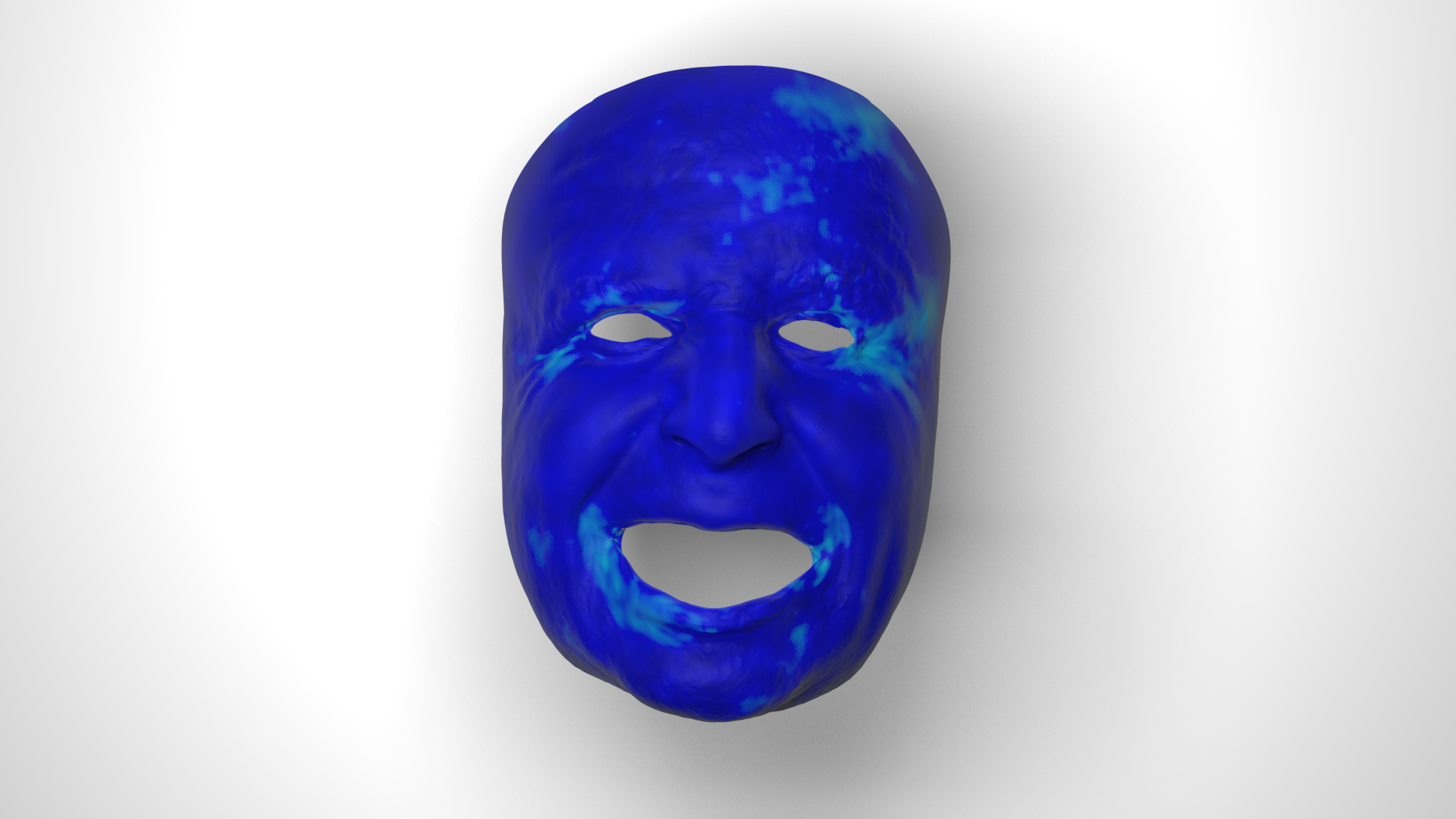} &
    \includegraphics[trim=480 0 480 0, clip, width=1\linewidth]{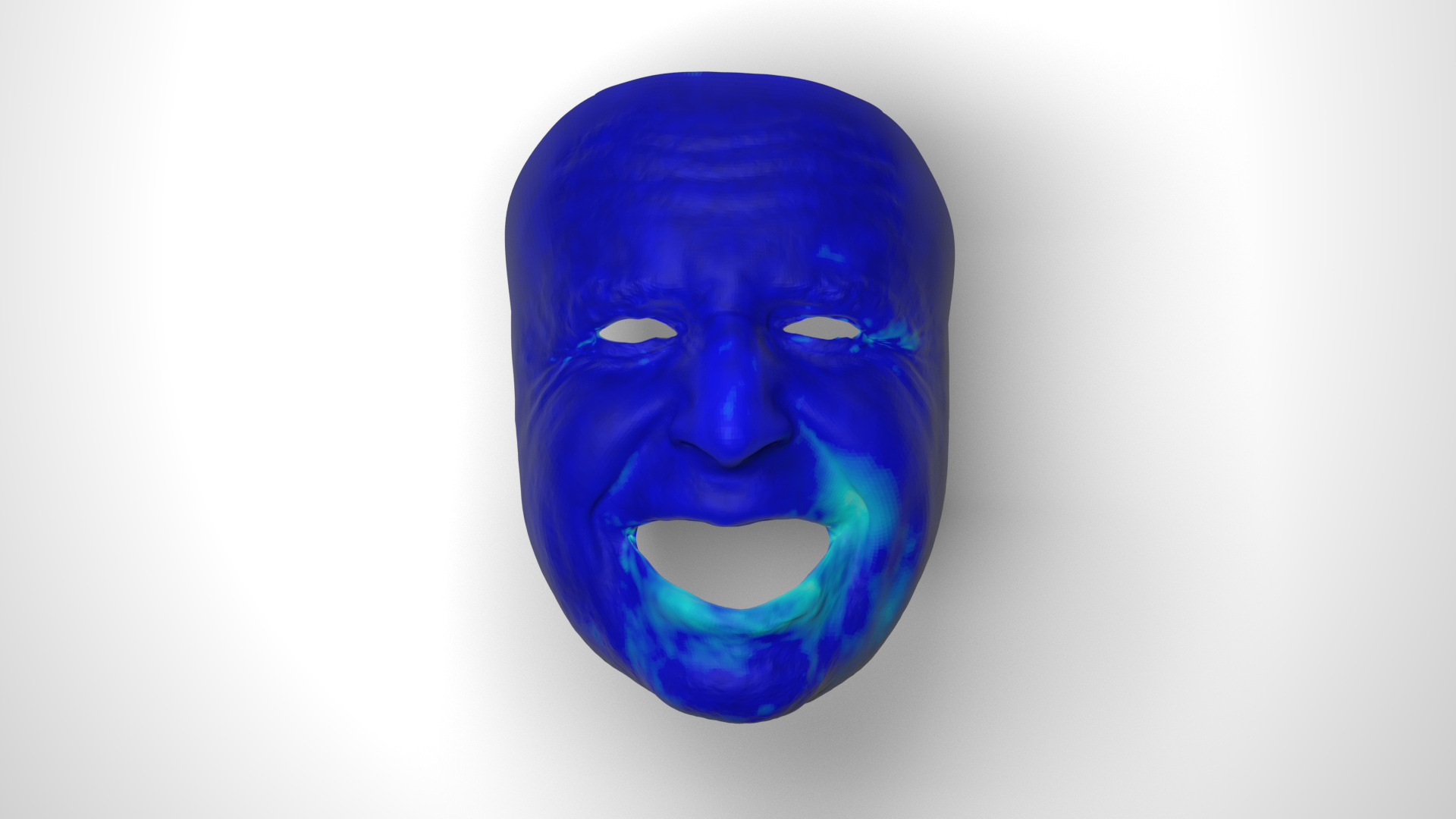} &
    \includegraphics[trim=480 0 480 0, clip, width=1\linewidth]{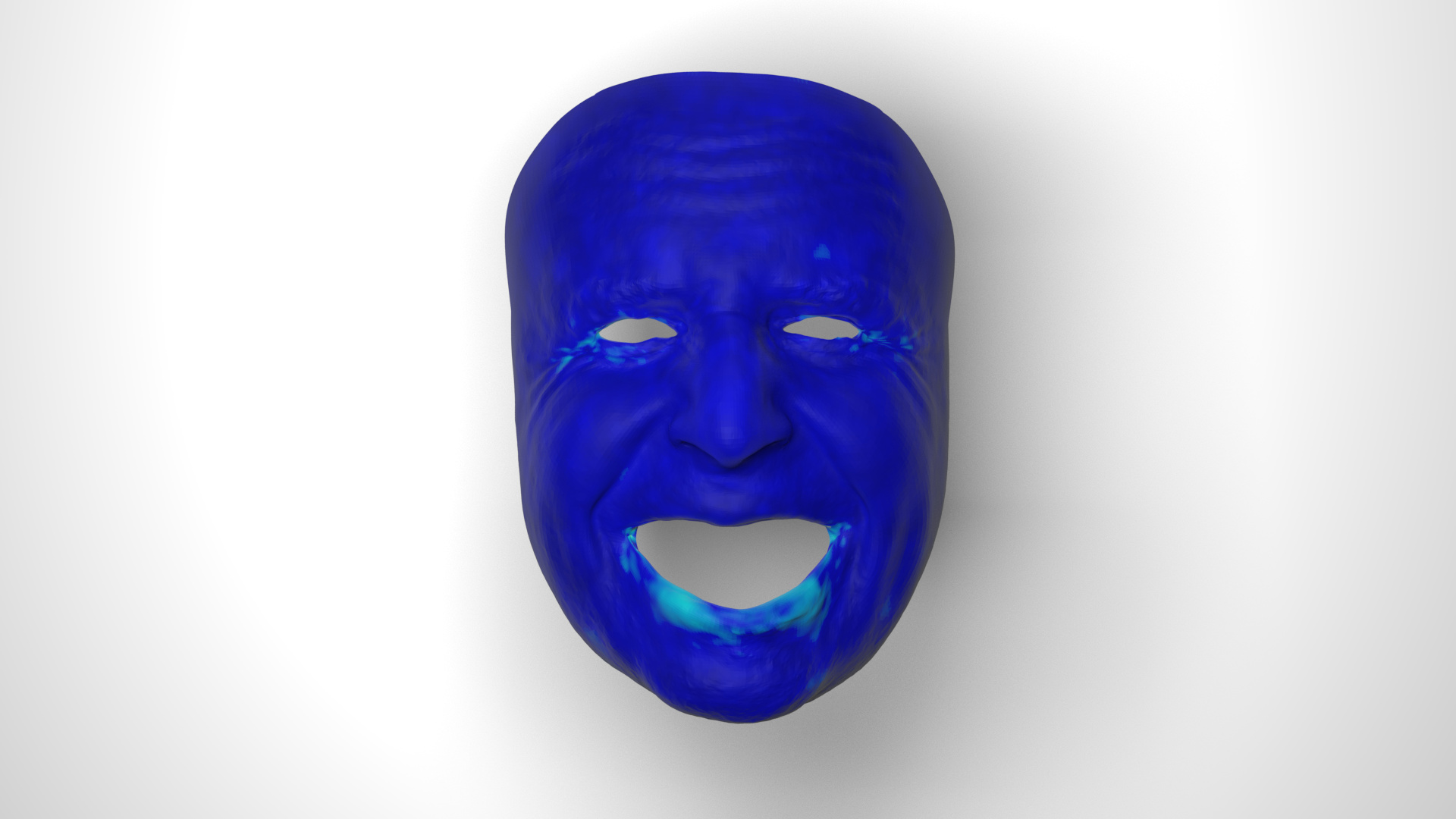} \\
    \small{Baseline-NDiffJaw} & \small{Ours} & \small{Baseline-NDiffJaw} & \small{Ours}
    \end{tabular}
    \caption{Comparison with Baseline-NDiffJaw. We show two different examples.}
    \label{fig:ablation_jaw}
\end{figure}

We examine the importance of our simulator-integrated pipeline using the face dataset.
To do this, we train two more networks with different simplifications:
one trained with only stage 1 (Baseline-Stage1) and another trained
without the differentiable connection between bone and simulator (Baseline-NDiffJaw).
Average vertex error is reported in \Tab{error}. 
It can be seen that our full model produces the best accuracy.
Note that the error difference between our full model and Baseline-NDiffJaw {is not trivial}, 
since we average the error over 44K vertices.
The differentiable connection between the bone and simulator mainly reduces the error around the mandible region, as shown in \Fig{ablation_jaw}.

\begin{table}
    \caption{Average vertex error (mm) for different methods}\vspace*{-3mm}
    \begin{tabular}{c|ccc}
        \toprule
        \textbf{Method} & \textbf{Baseline-Stage1} & \textbf{Baseline-NDiffJaw} & \textbf{Ours}  \\ \hline
        \textbf{Error}	          & 0.88 & 0.42 & 0.37	 \\
        \bottomrule
    \end{tabular}
    \label{tab:error}\vspace*{-3mm}
\end{table}

\input{fig/ablation_norm.tex} 

\section{Conclusion}

In this work, we have presented a physics- and data-driven method for computing control parameters of actuated soft bodies. We have demonstrated that the implicit representation offers the unique advantage of being shape invariant. This property enables the application of the method to different soft body input meshes, as demonstrated for three different datasets. Moreover, it renders the method agnostic to the discretization resolution used during training and testing, which is often a prerequisite in practical settings. For shapes that require an internal bone structure, such as the human face, we can additionally use an implicit formulation to parameterize the bone kinematics. 
Putting rigorously derived physics priors into the core of the learning facilitates the model to reproduce non-linear deformations and fine geometric details like wrinkles. 

While our implicit formulation is general across soft body models, it relies on some specific design choices.
First, the derived closed-form Hessian matrices are specific to the shape targeting constitutive model and thus not generally applicable to other models. Similarly, the actuation initialization would need to be changed with a different constitutive model.
Second, our implicit model of bone kinematics is targeted for the mandible and thus only supports a single rigid transformation. In the human body examples, we use the provided skeleton position to define the boundary conditions in the simulation but do not optimize for the kinematic chain. Our method could be extended to multiple bones and non-rigid transformations in future work.
Third, our method only considers actuation and bone kinematics. Additional mechanical parameters could be taken into account, such as in \citeN{Kadlecek2019BuildingData} where heterogeneous stiffness and prestrain are optimized for the face.

As discussed by \citeN{srinivasan2021learning}, another noticeable limitation of our model is that there are no explicit guarantees that the optimized mechanical parameters have an anatomical meaning. However, the chosen design alleviates the need for an anatomically correct geometric model and thus substantially reduces manual modeling labor, while still generating physically plausible deformations.
In accordance with related work, our method does not generalize well across different subjects as the implicit function encodes the actuation for a specific shape. 
While we can map the optimized actuation signals to another input shape, this will yield larger errors than if we were to retrain the model with the respective facial input scans or body poses.
Therefore, future work is needed to support important downstream applications such as expression retargeting.

\begin{acks}

We thank the anonymous reviewers for their constructive comments. We thank Daniel Dorda for his valuable feedback.
The work is supported by the 
{Swiss National Science Foundation} under Grant No.: {200021\_197136}.
\end{acks}

\bibliographystyle{ACM-Reference-Format}
\bibliography{references}

\clearpage
\section{Supplemental Material}
\subsection{Implicit Formulation for Actuation}
We adopt the energy density function $\Psi$ from shape targeting \cite{klar2020shape} for modeling the internal actuation mechanism:
\begin{equation}
    \Psi(\F, \A) = \underset{\R \in \SO(3)}{\operatorname{min}}||\F - \R\A||^2_F, \tag{\ref{eq:shapetarget}}
\end{equation}
where $\F$ is the local deformation gradient, 
$\A$ is a symmetrical $3 \times 3$ matrix (assuming 3D simulation), 
which could be represented as a vector $\mathbf{b} \in \mathbb{R}^6$ using the following convention:
\begin{equation}
    \label{eq:actuation_tensor}
\mathrm{A}=\left[\begin{array}{ccc}
1+\mathbf{b}_{1} & \mathbf{b}_{2} & \mathbf{b}_{3} \\
\mathbf{b}_{2} & 1+\mathbf{b}_{4} & \mathbf{b}_{5} \\
\mathbf{b}_{3} & \mathbf{b}_{5} & 1+\mathbf{b}_{6}
\end{array}\right].
\end{equation}
The rotation matrix $\R$ is used to make $\Psi$ rotationally invariant.
The optimal $\R^*$ is the polar decomposition of $\F\A$, as mentioned in \cite{klar2020shape}.
We assign an actuation matrix to every spatial point $\x$ in the object material space,
\ie $\A(\x) = \N_{\A}(\x)$. 

We introduce the following notations to simplify the derivation.
We define $\vec(\cdot)$ as row-wise flattening of a matrix into a vector:
$$
\operatorname{vec}(\mathbf{A})=\left[\begin{array}{c}
\A_{11} \\
\A_{12} \\
\A_{13} \\
\A_{21} \\
\A_{22} \\
\A_{23} \\
\A_{31} \\
\A_{32} \\
\A_{33}
\end{array}\right],
$$
where the subscript $ij$ indicates row $i$ and column $j$ of the original matrix $\A$.
We use the corresponding lower case letter with a symbol $\check{\cdot}$ 
to indicate that it is a vectorized matrix.
Similarly, we have $\f = \vec(\F)$, $\r = \vec(\R)$.
We define the expanded symmetric matrix $\hat{\A}$ as
\begin{equation}
    \hat{\A} = \left[
                \begin{array}{ccc} 
                    \A&0&0 \\
                    0&\A&0 \\
                    0&0&\A \\
                \end{array} 
            \right]. \tag{\ref{eq:actmat}}
\end{equation}
We add the symbol $\hat{\cdot}$ to indicate that it is an expanded matrix. 
In addition, $\hat{\F}$ and $\hat{\R}$ are similarly defined as
$$
    \hat{\F} = \left[
                \begin{array}{ccccccccc} 
                    \F_{11}&0&0&\F_{12}&0&0&\F_{13}&0&0 \\
                    0&\F_{11}&0&0&\F_{12}&0&0&\F_{13}&0 \\
                    0&0&\F_{11}&0&0&\F_{12}&0&0&\F_{13} \\
                    \F_{21}&0&0&\F_{22}&0&0&\F_{23}&0&0 \\
                    0&\F_{21}&0&0&\F_{22}&0&0&\F_{23}&0 \\
                    0&0&\F_{21}&0&0&\F_{22}&0&0&\F_{23} \\
                    \F_{31}&0&0&\F_{32}&0&0&\F_{33}&0&0 \\
                    0&\F_{31}&0&0&\F_{32}&0&0&\F_{33}&0 \\
                    0&0&\F_{31}&0&0&\F_{32}&0&0&\F_{33} 
                \end{array} 
            \right].
$$
With these notations, the matrix-matrix multiplication can be expressed as a matrix-vector multiplication, paving the way for deriving the hessian, 
\eg $\vec(\R\A) = \hat{\R}\a = \hat{\A}\r$, and $\vec(\F\A)=\hat{\F}\a$.

The continuous energy function $\tilde{E}$ for the simulation is defined as
\begin{equation}
    \tilde{E} = \int_{\mathcal{D}^{0}} \frac{1}{2}\left\|\mathbf{F(\x)}-\mathbf{R^{*}(\x) A(\x)}\right\|_{F}^{2} dV,
\end{equation}
where $\mathcal{D}^{0}$ denotes the material space (undeformed space) of the object.
Following the standard practices of finite element method, 
we discretize $\mathcal{D}^{0}$ using tiny elements connected by nodal vertices:
\begin{equation}
    \tilde{E} \approx \sum_{e}\int_{\mathcal{D}_{e}^{0}} \frac{1}{2}\left\|\mathbf{F}(\x)-\mathbf{R}^{*}(\x) \mathbf{A}(\x)\right\|_{F}^{2} dV,
\end{equation}
where $e$ denotes an element, $\mathcal{D}_{e}^{0}$ indicates the continuous region inside $e$ while $V_e$ is its volume.
We sample $N$ points inside each $\mathcal{D}_{e}^{0}$ to approximate the integral:
\begin{equation}
    \tilde{E} \approx \sum_{e}\frac{V_e}{N}\sum_{i}^{N} \frac{1}{2}\left\|\mathbf{F}(\x_{e, i})-\mathbf{R}^{*}(\x_{e, i}) \mathbf{A}(\x_{e, i})\right\|_{F}^{2}.
\end{equation}
The deformation gradient $\F$ at each point $\x$ can be approximated by the nodal vertices $\u$ around it 
through differentiating the interpolation weight $w$: 
\begin{align}
    \tilde{E}
         &\approx \sum_{e}\frac{V_e}{N}\sum_{i}^{N} \frac{1}{2}\left\|\frac{\partial\sum_j{w_j(\x_{e, i})\u_{j}}}{\partial \x_{e, i}}-\mathbf{R}^{*}(\x_{e, i}) \mathbf{A}(\x_{e, i})\right\|_{F}^{2} \\
         &= \sum_{e}\frac{V_e}{N}\sum_{i}^{N} 
         \frac{1}{2}\left\|\sum_j{\u_j \otimes \nabla w_j(\x_{e, i}) }-\mathbf{R}^{*}(\x_{e, i}) \mathbf{A}(\x_{e, i})\right\|_{F}^{2} 
\end{align}
We adopt hexahedral elements and a trilinear interpolation scheme.
Therefore, $\F$ at the location $\x_{e, i}$ is estimated only with 8 nodal vertices associated with the element $e$.
By using trilinear interpolation, we can apply a linear mapping matrix $\G \in \mathbb{R}^{9\times 24}$ 
to calculate the vectorized $\F$ as $\f = \G\u_e$, 
where $\u_e \in \mathbb{R}^{24}$ denotes the concatenated nodal vertices associated with element $e$.
Therefore, the discretized energy function $E$ is given as
\begin{equation}
    \label{eq:discretized}
    E(\u, \mathcal{A}) = \sum_{e}\frac{V_e}{N}\sum_{i}^{N} \underbrace{\frac{1}{2}\left\|\G(\x_{e, i})\u_e-\hat{\A}(\x_{e, i}){\r}^{*}(\x_{e, i}))\right\|_{2}^{2}}_{\Psi_{e,i}}, \tag{\ref{eq:energy3}}
\end{equation}
where $\mathcal{A}$ denotes all the sampled actuation matrices $\A$ 
from the network $\N_{\A}$.

\subsection{Hessians}
For deriving $\H_\u$ and $\H_\Omega$, {we use the fact that}
$\vec(\R\A) = \hat{\R}\a = \hat{\A}\r$, 
$\vec(\F\A)=\hat{\F}\a=\hat{\A}\f$,
and $\vec(\F) = \f = \G\u_e$.

$\Psi_{e, i}$ in \Eq{discretized} is the key for deriving $\H_\u$, 
since $\H_\u$ is the accumulation of all these tiny hessians $\nabla^2 \Psi_{e, i}$.
Now, we omit $(\x, e, i, *)$ except $\u_e$ for simplicity. 
Using projective dynamics, $\nabla \Psi = \G^{\top}(\G\u_e - \hat{\A}\r)$.
Taking the derivative of $\nabla \Psi$, we have
\begin{align}
    \frac{\partial \nabla \Psi}{\partial \u_e} 
    &= \G^{\top}\G - \G^{\top}\hat{\A} \frac{\partial \r}{\partial \hat{\A}\f}\frac{\partial \hat{\A}\f}{\partial \u_e}\nonumber\\
    &= \G^{\top}\G - \G^{\top}\hat{\A} \frac{\partial \r}{\partial \hat{\A}\f}\hat{\A}\frac{\partial \f}{\partial \u_e}\nonumber\\
    &= \G^{\top}\G - \G^{\top}\hat{\A} \frac{\partial \r}{\partial \hat{\A}\f}\hat{\A}\G\nonumber\\
    &= \G^{\top}\G - \G^{\top}\hat{\A} \H_\R \hat{\A}\G. \tag{\ref{eq:hessian1}}
\end{align}
Note that $\r$ comes from the polar decomposition of 
$\F\A$,
$\H_\R$ is thus the \emph{rotation gradient}.
The closed form for $\H_\R$ has already been {derived in \cite{kim2020dynamic}}, 
which can be constructed from its off-the-shelf three eigenvectors and eigenvalues, as
$$
\H_\R = \frac{\partial \r}{\partial \hat{\A}\f} = \sum_i^3 \lambda_i \operatorname{vec}\left(\mathbf{Q}_{i}\right) \vec\left(\mathbf{Q}_{i}\right)^{T}
$$
$$\lambda_{0}=\frac{2}{\sigma_{x}+\sigma_{y}} 
\quad 
\mathbf{Q}_{0}=\frac{1}{\sqrt{2}} \mathbf{U}\left[\begin{array}{ccc}0 & -1 & 0 \\ 1 & 0 & 0 \\ 0 & 0 & 0\end{array}\right] \mathbf{V}^{T}
$$
$$\lambda_{1}=\frac{2}{\sigma_{y}+\sigma_{z}}
\quad
\mathbf{Q}_{1}=\frac{1}{\sqrt{2}} \mathbf{U}\left[\begin{array}{ccc}0 & 0 & 0 \\ 0 & 0 & 1 \\ 0 & -1 & 0\end{array}\right] \mathbf{V}^{T}
$$
$$\lambda_{2}=\frac{2}{\sigma_{x}+\sigma_{z}}
\quad
\mathbf{Q}_{2}=\frac{1}{\sqrt{2}} \mathbf{U}\left[\begin{array}{ccc}0 & 0 & 1 \\ 0 & 0 & 0 \\ -1 & 0 & 0\end{array}\right] \mathbf{V}^{T}
$$
where $\mathbf{U}\mathbf{\Sigma}\mathbf{V}^{\top}$ is the singular value decomposition of $\F\A$, and
$\sigma_x$, $\sigma_y$, $\sigma_z$ are the three diagonal entries in $\Sigma$.
Similarly, we have $\partial \nabla\Psi / \partial \a$ given as follows:
\begin{align}
    \frac{\partial \nabla \Psi}{\partial \a} 
    &= - \G^{\top} \frac{\partial \hat{\R}\a}{\partial \a} - \G^{\top}\hat{\A}\frac{\partial \r}{\partial \hat{\F}\a}\frac{\partial \hat{\F}\a}{\partial \a}\nonumber\\
    &= - \G^{\top}\hat{\R} - \G^{\top}\hat{\A}\H_\R\hat{\F}, \tag{\ref{eq:hessian2}}
\end{align}
which can be used for constructing $\H_{\Omega}$.

\subsection{Network}
Even though our network is primarily designed to animate the human face, it is also applicable to other soft bodies.
In our settings, we only consider the the relative movement between skull and mandible as bone kinematics,
since that is enough to articulate diverse expressions. Thus, the skull is fixed all the times.

\begin{figure} %
    \centering
    \includegraphics[trim=0px 0px 0px 0px, clip, width=0.45\textwidth]{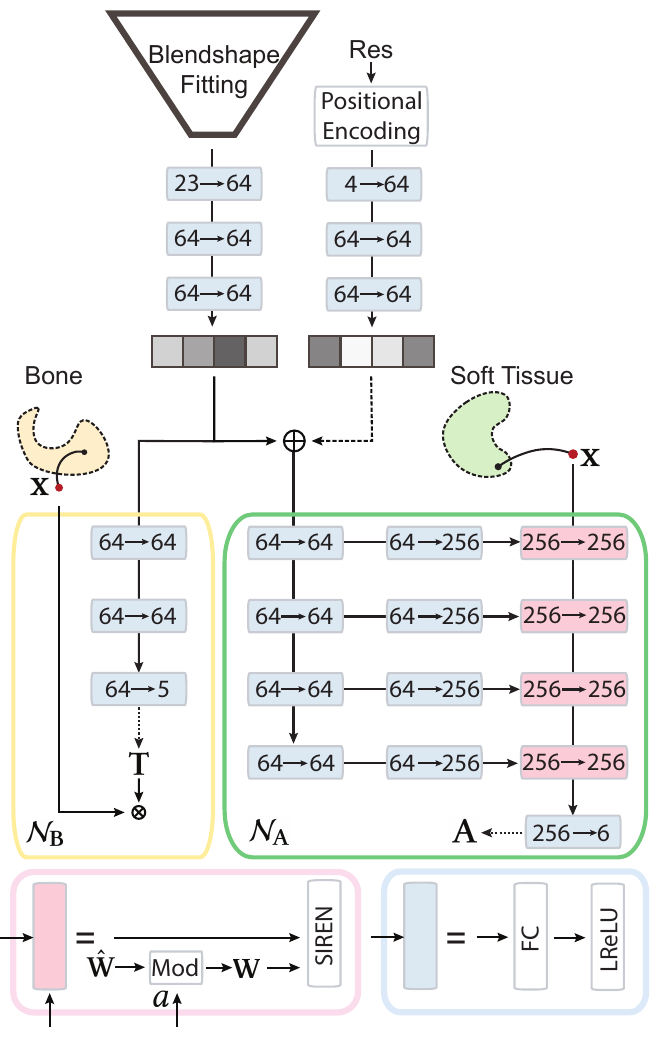}
    \caption{Architecture of our network. 
    Blocks with the same color share the same function.
    The text $n_i \rightarrow n_o$ in each colored block 
    means that the dimentions of the input and the output feature vectors are $n_i$ and $n_o$ respectively.
    The text FC means the fully connected layer, LReLU indicates the LeakyReLU nonlinearity, 
    and SIREN represents a fully connected layer followed by sine activation function, whose weights are $\W$ conditional on $a$.
    For continuous resolution conditioning, we add another branch on top of $\N_\A$, as indicated by the dashed arrow.}
    \label{fig:network_res}
\end{figure}

The entire architecture consists of three parts, an encoder, 
an actuation-generative coordinate-based network $\N_{\A}$,
and a bone-generative network $\N_{\mathbf{B}}$.
The encoder outputs a latent code $\z$ representing an input shape.
$\N_{\A}$ is conditioned on $\z$ to generate an actuation matrix $\A$ for each input spatial point in the soft tissue domain.
$\N_{\mathbf{B}}$ is dependent upon $\z$ to produce the transformed mandible position 
for each input spatial point in the bone domain (mandible domain for the face).
In practice, we parameterize $\A$ with a vector $\b \in \mathbb{R}^6$, as in \Eq{actuation_tensor}, which is the direct output of $\N_{\A}$.
In addition, the mandible motion is linked to the skull (fixed in our settings) via a pivot point, represented as a
joint with two degrees of freedom for rotation and three for translation, as $\Theta = \{\theta_x, \theta_y, t_x, t_y, t_z\} \in \mathbb{R}^5$, 
which is the direct output of $\N_{\mathbf{B}}$.
$\Theta$ is subsequently converted into a transformation matrix $\mathbf{T}$ subsequently.
The encoder is a global shape descriptor, 
namely the blendweights fitted from 23 blendshapes followed by 3 fully connected layers.
We use our proposed modulated SIREN layer as the backbone layer
for $\N_{\A}$, whose modulating coefficients are mapped from the latent code $\z$ with a tiny MLP.
We use 4 such layers in total. The hyperparamter $\omega_0$ \cite{siren2020} for SIREN is set to 30.
For $\N_{\mathbf{B}}$, we simply use 3 fully connected layers with LeakyReLU nonlinearity (0.01 negative slope).
\Fig{network_res} shows the detailed architecture. 
We have chosen $\alpha=0$ for the normal weight in the loss function for both the starfish and human body examples, 
and $\alpha=1$ for the face examples, as we found that the inclusion of the normal constraint positively affects 
the fidelity of the resulting wrinkles and facial details.
We use the ADAM optimizer~\cite{kingma2015adam} to jointly train our networks.
We run 1700 epochs for training stage 1 with a batch size of $4$ and an initial learning rate of $0.0002$.
We run 30 epochs for training stage 2 with a batch size of $1$ and an initial learning rate of $0.0001$.
The learning rates for both stages are decayed to $0$ gradually.
 
\input{fig/transfer.tex}

For continuous resolution conditioning, we add another branch on top of $\N_\A$, as indicated by the dashed arrow in \Fig{network_res}, which starts with a positional encoding layer to convert the 
$1$ dimensional scalar input into a $4$ dimensional feature vector, 
followed by 3 fully connected layers. 
We use our network pretrained without this branch on one resolution ($268$K sampled points) to 
execute the transfer learning for continuous resolution conditioning.
For training, we uniformly sample 20 resolutions with the number of sampled points ranging from $42$K to $268$K, 
and use our simulator-integrated pipeline.
We run a total of 30 epoches with a batch size of $1$ and a learning rate of $0.0001$ (decayed to $0$ gradually).
For testing, we uniformly sample 25 resolutions (different from training).

\subsection{Experiments}
In this section we discuss additional experiments and results.
\paragraph{Transfer Learning and Resampling Results}
Transfer learning on a single resolution can be used as an alternative strategy to the proposed continuous resolution approach, but requires 1 epoch to make the network consistent with the new discretization.
We show results on different resolutions in~\Fig{transfer}.
At the top, we trained the model on the resolution that entails sampling 268K points (left) 
and applied it at test time to 42K (right).
At the bottom, we trained the model on a resolution of 89K (left) and applied it to 502K at test time (right). 
{Our results indicate that the trained model can accurately represent the dominant frequencies of the actuation signal and reproduce them at test time.} 
The middle columns show the results with standard up- and downsampling of the actuation values. While the error magnitudes are comparable to our result, artifacts are clearly visible on the surface, for example near the eyebrow and forehead.

\end{document}